\documentclass[acmtog,nonacm]{acmart}

\setcopyright{rightsretained}
\usepackage{color}
\usepackage{soul}
\usepackage{booktabs}
\usepackage{colortbl}
\usepackage{tabularx}
\usepackage{multirow}
\usepackage{rotating}
\usepackage{subcaption}
\usepackage{pgfplots}

\definecolor{niceblue}{rgb}{0, 0.635, 0.929}

\DeclareRobustCommand{\hlcol}[2]{{\sethlcolor{#1}\hl{#2}}}
\newcommand{\sota}[1]{\cellcolor{green!25}\textbf{#1}}
\newcommand{\dsota}[1]{\cellcolor{yellow!25}\underline{#1}}

\newcolumntype{Y}{>{\centering\arraybackslash}X}
\newcommand{\SOMA}{SOMA}
\newcommand{\SMPLH}{{SMPL-H}}
\newcommand{\SynthBody}{\emph{SynthBody}}
\newcommand{\SynthHand}{\emph{SynthHand}}
\newcommand{\SynthFace}{\emph{SynthFace}}

\newcommand{\copylabel}[2]{\textit{\footnotesize #1 $\copyright$ #2}}

\begin{document}

\title{Look Ma, no markers: holistic performance capture without the hassle}

\author{Charlie Hewitt}
\email{chewitt@microsoft.com}
% \affiliation{\institution{Microsoft}\city{Cambridge}\country{United Kingdom}}
\orcid{0000-0003-3943-6015}
\author{Fatemeh Saleh}
\email{fatemehsaleh@microsoft.com}
% \affiliation{\institution{Microsoft}\city{Cambridge}\country{United Kingdom}}
\orcid{0000-0002-3695-9876}
\author{Sadegh Aliakbarian}
\email{saliakbarian@microsoft.com}
% \affiliation{\institution{Microsoft}\city{Cambridge}\country{United Kingdom}}
\orcid{0000-0003-3948-6418}
\author{Lohit Petikam}
\email{lohitpetikam@microsoft.com}
% \affiliation{\institution{Microsoft}\city{Cambridge}\country{United Kingdom}}
\orcid{0000-0001-6629-7490}
\author{Shideh Rezaeifar}
\email{srezaeifar@microsoft.com}
% \affiliation{\institution{Microsoft}\city{Zurich}\country{Switzerland}}
\orcid{0000-0002-8103-3722}
\author{Louis Florentin}
\email{lflorentin@microsoft.com}
% \affiliation{\institution{Microsoft}\city{Paris}\country{France}}
\orcid{0009-0000-1825-1180}
\author{Zafiirah Hosenie}
\email{zhosenie@microsoft.com}
% \affiliation{\institution{Microsoft}\city{Cambridge}\country{United Kingdom}}
\orcid{0000-0001-6534-593X}
\author{Thomas J. Cashman}
\email{tcashman@microsoft.com}
% \affiliation{\institution{Microsoft}\city{Cambridge}\country{United Kingdom}}
\orcid{0000-0001-7975-8567}
\author{Julien Valentin}
\email{juvalen@microsoft.com}
% \affiliation{\institution{Microsoft}\city{Zurich}\country{Switzerland}}
\orcid{0000-0001-8503-6325}
\author{Darren Cosker}
\email{coskerdarren@microsoft.com}
% \affiliation{\institution{Microsoft}\city{Cambridge}\country{United Kingdom}}
\additionalaffiliation{\institution{University of Bath}\city{Bath}\country{United Kingdom}}
\orcid{0000-0001-5177-4741}
\author{Tadas Baltru\v{s}aitis}
\email{tabaltru@microsoft.com}
% \affiliation{\institution{Microsoft}\city{Cambridge}\country{United Kingdom}}
\orcid{0000-0001-7923-8780}
\affiliation{\institution{Microsoft}\country{United Kingdom, Switzerland and France}}

\renewcommand{\shortauthors}{Hewitt et al.}

\begin{abstract}
We tackle the problem of highly-accurate, holistic performance capture for the face, body and hands simultaneously. 
Motion-capture technologies used in film and game production typically focus only on face, body or hand capture independently, involve complex and expensive hardware and a high degree of manual intervention from skilled operators. 
While machine-learning-based approaches exist to overcome these problems, they usually only support a single camera, often operate on a single part of the body, do not produce precise world-space results, and rarely generalize outside specific contexts.
In this work, we introduce the first technique for marker-free, high-quality reconstruction of the complete human body, including eyes and tongue, without requiring any calibration, manual intervention or custom hardware. 
Our approach produces stable world-space results from arbitrary camera rigs as well as supporting varied capture environments and clothing. 
We achieve this through a hybrid approach that leverages machine learning models trained exclusively on synthetic data and powerful parametric models of human shape and motion. 
We evaluate our method on a number of body, face and hand reconstruction benchmarks and demonstrate state-of-the-art results that generalize on diverse datasets. 
% We also release the synthetic datasets of the body, face and hands which we use to train our models.
\end{abstract}




\begin{teaserfigure}
  \centering
    \includegraphics[height=0.33\linewidth]{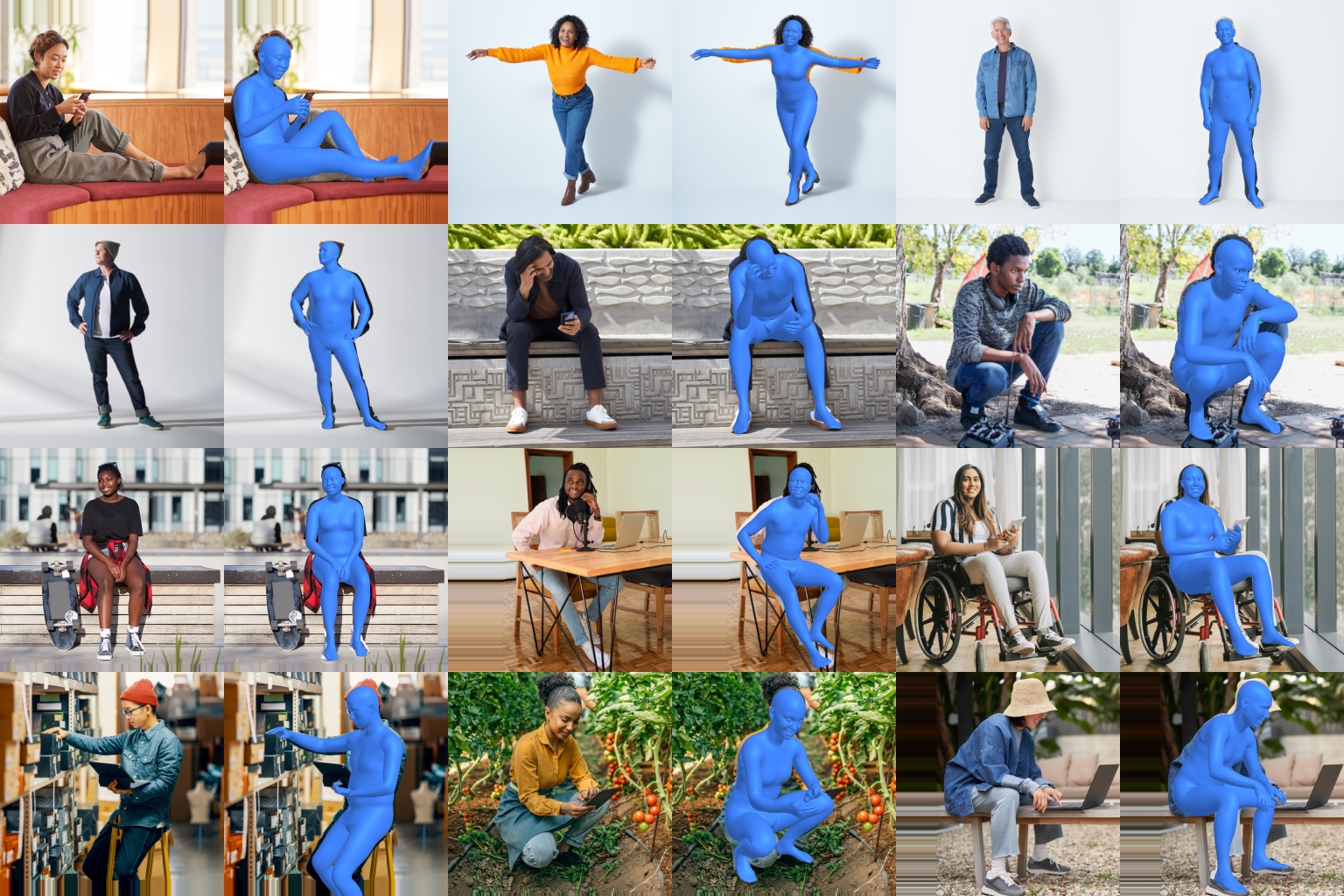}
    \hfill
    \includegraphics[height=0.33\linewidth]{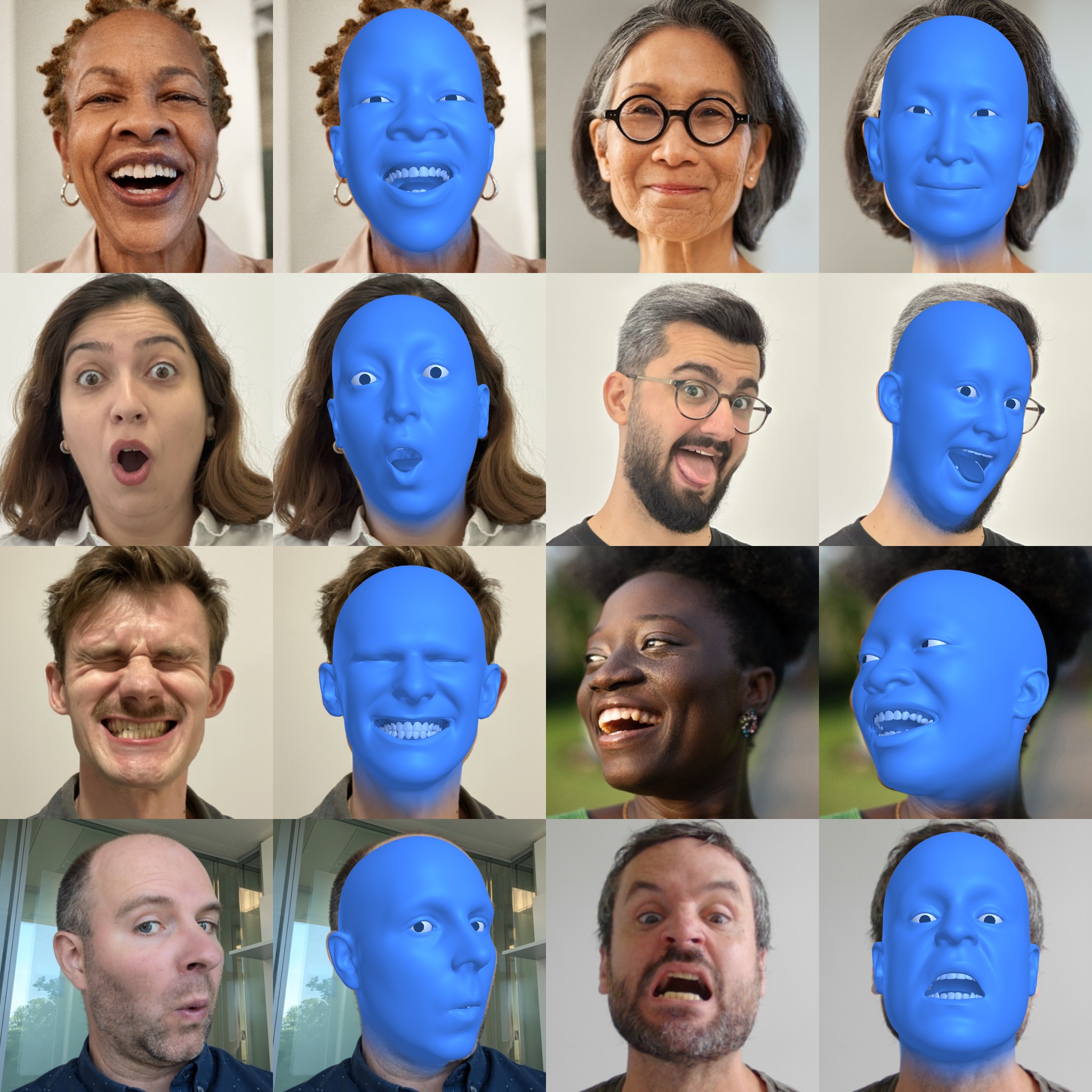}
    \hfill
    \includegraphics[height=0.33\linewidth]{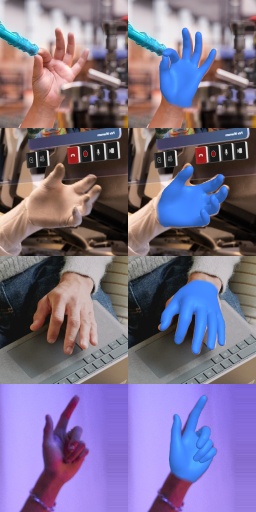}\\
  \caption{Results of our holistic performance capture pipeline for the body, face, and hands. We fit our parametric model to image data using neural networks for dense landmark, pose and shape prediction followed by optimization, providing robust \emph{and} accurate reconstructions. Our approach supports single- or multiple-camera setups without requiring calibration or specialist hardware and tracks the whole body including the eyes and tongue.}
  \label{fig:teaser}
\end{teaserfigure}

\maketitle

\section{Introduction}

High-quality performance capture is a challenging problem, which traditional approaches simplify by introducing additional constraints.
These take the form of expensive specialist hardware such as optical markers, precisely calibrated multi-camera rigs, or specific lighting setups. 
Additionally, the scope of capture is often limited; just the face or hands are targeted, further simplifying the problem. 
These methods achieve high-quality results, but only under certain circumstances and at significant cost. 
They usually require experts to configure and operate them, and the results need artist work to clean up or combine feeds from different capture technologies. 
In short, the traditional approach has exchanged rigidity for quality, and, in the process, has limited high-quality performance capture to the worlds of film, television and game production.

In this work, we tackle the problem of accurate, robust and adaptable performance capture of the whole human. 
Our approach eliminates the need for specialist hardware, supports any number of cameras without needing calibration, can achieve state-of-the-art results fully automatically, and significantly lowers the amount of expert manual labor required to reach production-grade performance capture results. 
% That is, we show how to do high-quality performance capture at scale. 
We demonstrate holistic performance capture of the face, hands and body simultaneously in \autoref{fig:teaser}.

Recent academic work relies heavily on ML-based approaches and \citet{moveai} introduced an ML capture solution for body pose aimed at production settings.
These are less dependent on specialist hardware, and function from a single image taken with any camera~\cite{zhang2023pymaf,goel2023humans,wood20223d}.
Some also provide holistic capture of the body, hands and face~\cite{zhang2023pymaf,pavlakos2019expressive}.
ML approaches which directly regress body pose and shape can be highly robust, as a result of training on large and diverse datasets~\cite{goel2023humans,zhang2023pymaf}.
However, they are less precise than traditional approaches (e.g., markers or photogrammetry) and deep neural networks (DNNs) which predict image-space features such as 2D landmarks.
They are also not easily adapted to arbitrary multi-camera rigs and are often unable to produce world-space results.
Our method exploits the benefits of ML to provide robustness, but improves accuracy and adaptability by incorporating model fitting.

Many ML approaches share a challenge in ensuring the privacy and diversity of their training data.
Human-annotated datasets are also limited in the fidelity of annotations that can be achieved and often include errors and noise.
We choose instead to use \emph{only} synthetic data to train the ML components of our method.
This allows us to track very dense 2D landmarks, as well as predict precise model parameters without label noise.
It also eliminates privacy concerns and gives us explicit control over the data in order to mitigate fairness disparities.

In summary, we combine the benefits of model fitting and ML-based methods to achieve:
\emph{(1) Robustness} of DNNs which directly regress body pose and shape.
\emph{(2) Accuracy} of DNNs predicting image-space features.
\emph{(3) Adaptability} of model fitting to support arbitrary cameras.
A comparison of the capabilities of our method with other performance capture technologies is shown in \autoref{tab:method_comparison}.
Specifically, in this work we present the following:

\begin{enumerate}
    \item A system providing state-of-the-art holistic 3D human reconstruction including body shape and pose, face shape and expression, hand and tongue articulation and eye gaze prediction from monocular images.
    \item Extension of this system to the multi-view scenario, providing improved accuracy and world-space 3D results. The system is adaptable to arbitrary camera rigs without explicit calibration and does not require any specialized hardware or manual intervention.
    \item A synthetic data pipeline for generating realistic images of humans with rich ground-truth annotations using a parametric model of the whole human with diverse face shape and tongue motion. We release the datasets of the body, face and hands which we use for training our method.
\end{enumerate}

Our method shows that it is possible to achieve high-fidelity, holistic performance capture fully automatically from image inputs alone.
% Our approach does not require precisely calibrated multi-camera rigs or other proprietary hardware, and can achieve state-of-the-art results for human reconstruction.\TCnote{This feels repetitive with the contributions above; why repeat here?}
This demonstrates that wider access to very high-quality performance capture is possible at lower cost.
It can also accelerate the workflow for production performance capture significantly, as calibration is not required and manual work is reduced. % given the holistic capture.
More information and instructions for accessing the datasets can be found at {\color{niceblue}{\url{https://aka.ms/SynthMoCap}}}.

\newcommand{\YES}{\cellcolor{green!25}\checkmark}
\newcommand{\NO}{\cellcolor{red!25}$\times$}
\newcommand{\MAYBE}[1]{\cellcolor{green!25}\checkmark\tiny{#1}}

\begin{table}
\caption{
    Comparison of existing performance capture technologies with our proposed method.
    \label{tab:method_comparison}
}
\footnotesize
\begin{tabularx}{\linewidth}{@{}lYYYYYYYYY}
\toprule
System & \begin{sideways}Marker-free\end{sideways} & \begin{sideways}Body Pose\end{sideways} & \begin{sideways}Hand Pose\end{sideways} & \begin{sideways}Facial Expression\end{sideways} & \begin{sideways}Shape\end{sideways} & \begin{sideways}Calibration Free\end{sideways} & \begin{sideways}Commodity Hardware\end{sideways} & \begin{sideways}Arbitrary Rig\end{sideways} & \begin{sideways}World-space\end{sideways} \\
\midrule
Optical Markers \tiny{(e.g., Vicon, Optitrack)} & \NO & \YES & \MAYBE{*} & \MAYBE{\dag} & \NO & \NO & \NO & \YES & \YES \\
\citet{moveai} & \YES & \YES & \YES & \NO & \NO & \NO & \YES & \YES & \YES \\
\citet{faceware} & \YES  & \NO & \NO & \YES & \NO & \NO & \NO & \NO & \NO \\
Smart Suits \tiny{(e.g., XSens, Manus)} & \YES & \YES & \MAYBE{\S} & \NO & \NO & \NO & \NO & - & \NO \\
SMPLify-X~\cite{pavlakos2019expressive} & \YES & \YES & \YES & \YES & \YES & \YES & \YES & \NO & \NO \\
\citet{wood20223d} & \YES & \NO & \NO & \YES & \YES & \YES  & \YES & \YES & \YES \\
BEDLAM~\cite{black2023bedlam} & \YES & \YES & \NO & \NO & \YES & \YES & \YES & \NO & \NO \\
HMR2~\cite{goel2023humans} & \YES & \YES & \NO & \NO & \NO & \YES & \YES & \NO & \NO \\
PyMAF-X~\cite{zhang2023pymaf} & \YES & \YES & \YES & \YES & \YES & \YES & \YES & \NO & \NO \\
TokenFace~\cite{zhang2023accurate} & \YES & \NO & \NO & \YES & \YES & \YES & \YES & \NO & \NO \\
TEMPO~\cite{choudhury2023tempo} & \YES & \YES & \NO & \NO & \NO & \NO & \YES & \YES & \YES \\
WHAM~\cite{shin2024wham} & \YES & \YES & \NO & \NO & \NO & \YES & \YES & \NO & \YES \\
Ours & \YES & \YES & \YES & \YES & \YES & \YES & \YES & \YES & \YES \\
\bottomrule
\end{tabularx}\\[3pt]
\parbox{\columnwidth}{* often requires extensive manual cleanup,
\dag{} difficult to capture in the same shoot as body/hands,
\S{} typically requires manual alignment between body and hands.}
\end{table}

\section{Related Work}

% Here we provide a brief summary of related work.
% A tabular comparison to similar performance capture technologies and their capabilities is given in \autoref{tab:method_comparison}.

\subsection{Traditional Performance Capture}

Most performance capture technologies used in production settings are able to produce high quality results due to additional constraints imposed on the problem; limiting to a single body part, using optical markers, requiring precise calibration or enrollment data.
This usually increases cost and the amount of manual intervention required.

For example, \citet{faceware} consider only the face from a head-mounted camera and track a small number of landmarks, usually relying on manual annotation of a number of `training frames' and actor-specific enrollment sequences.
Optical marker systems (\citet{vicon}, Optitrack~\cite{optitrack}) are popular for body-pose capture.
These use specialist camera rigs that are expensive and require expert operators. 
The data often requires manual clean-up to resolve noisy, missing or swapped markers and to facilitate fitting a parametric model.
% These systems can be used for face and hand capture, but usually not holistically, and hands can require prohibitive levels of clean-up due to high self-occlusion~\cite{}.\TCnote{Do we have a citation to support?}
XSens~\cite{xsens}, \citet{stretchsense} and \citet{manus} use wearable `smart suits' to determine motion.
These rely heavily on enrollment sequences, can be expensive, and manual effort is often needed to combine the signals from different body parts.

Our system exploits advances in ML to move away from specialist hardware and allow anyone with a camera to achieve state-of-the-art performance capture.
Moreover, our unified model allows for holistic capture of facial expression, body pose and hand pose in one, without manual work to combine separate streams. 
% \TBnote{might want to talk about need to do marker mapping from landamrks to model being fit, which we do not need to do as we use synthetic data and same model, also can talk about avoiding sliding issue}

\subsection{Machine Learning Approaches}

Recent academic work focuses on the application of ML techniques to solve the problem of performance capture.
\citet{pavlakos2019expressive} use a DNN to regress skeletal 2D landmarks in an image and optimization to fit their parametric model to these observations.
The sparse nature of the landmarks mean they are unable to recover accurate body shape and facial expression.
\citet{wood20223d} employ dense 2D landmarks to achieve significantly higher-fidelity shape reconstruction and expression tracking, but only support the face and without tongue articulation.
\citet{zhang2023pymaf} exploit image-space information to produce holistic reconstructions at high fidelity from an image, but do not support video or arbitrary camera rigs.
Other ML approaches directly regress pose from monocular images, providing robust but less-precise results~\cite{goel2023humans,black2023bedlam} or only targeting specific body parts~\cite{zhang2023accurate}.
\citet{shin2024wham} also predict global trajectory and \citet{choudhury2023tempo} support arbitrary viewpoints, both providing world-space results but only reconstructing skeletal body pose.
\citet{moveai} introduced an ML-based production performance capture system supporting arbitrary cameras.
It uses skeletal landmark tracking, so is unable to recover facial expression or detailed body shape, and requires a calibration step.
We find the combination of image-space features (dense landmarks) and direct pose regression lets us achieve accurate \emph{and} robust results, while the combination of DNNs with conventional model-fitting %optimization
makes our system more adaptable than pure ML approaches.

\subsection{Models and Data}

\paragraph{Parametric models}
Many works consider different body parts in isolation, such as the face~\cite{li2017learning,wood2021fake}, hand~\cite{khamis2015learning,romero2017embodied} and body~\cite{allen2003space,anguelov2005scape,loper2015smpl}.
Some have combined hands and body into a single model, such as SMPL-H~\cite{romero2017embodied} and FrankMocap~\cite{rong2020frankmocap}, with Adam~\cite{joo2018total} and SMPL-X~\cite{pavlakos2019expressive} also including the face.
These are good statistical models of humans, but most do not lend themselves well to computer graphics applications for generating synthetic data.
For example, FLAME~\cite{li2017learning}, used in SMPL-X~\cite{pavlakos2019expressive}, has no teeth or tongue and the mesh does not subdivide well for close-up renders of the face.
% The wide-ranging adoption of many of these models for computer vision applications demonstrates the value of expressive parametric human models and the downstream tasks that they can enable.
We build on the work of \citet{romero2017embodied} (SMPL-H) and \citet{wood2021fake} to produce a parametric model of the full human including hand articulation and facial expression that is well suited to synthetic data generation.
We also improve the shape model for the face and add tongue articulation.

\paragraph{Synthetic Data}
Recent research has shown that synthetic data can be an effective means of achieving high-quality, robust reconstruction of humans. 
\citet{wood2021fake} demonstrate the possibility of performing face-related computer vision in the wild using synthetic data alone.
Similarly, \citet{black2023bedlam} and \citet{yang2023synbody} introduce the BEDLAM and SynBody datasets respectively, both containing monocular RGB videos of synthetic humans using the SMPL-X model~\cite{pavlakos2019expressive}.
They also demonstrate state-of-the-art accuracy on real-image benchmarks despite training with synthetic data alone, but their pipelines do not include high quality face shape and expression.
We also choose to use synthetic data, given that it is now possible to achieve parity with real data in some tasks, it significantly reduces concerns around privacy, and can easily include complex ground-truth annotations without error.
We extend the pipeline of \citet{wood2021fake} using our parametric human model to be able to generate data including the full human, rather than just the face.

\section{Synthetic Data}
\label{sec:synth_data}

We exclusively use synthetic data to train the DNNs used in our performance capture system.
The use of the same parametric model throughout the entire stack is critical to its success.
As the 2D landmarks correspond directly to 3D points on the mesh generated by the model, we can fit to them directly.
This eliminates the need for error-prone manual tuning of marker mappings~\cite{loper2014mosh} and prevents sliding of predicted landmarks relative to the mesh, as would be expected with DNNs trained on human-annotated landmark datasets or traditional contour tracking.

\begin{figure}
    \centering
    \begin{subfigure}{0.155\linewidth}
        \centering \tiny
        \includegraphics[width=\linewidth]{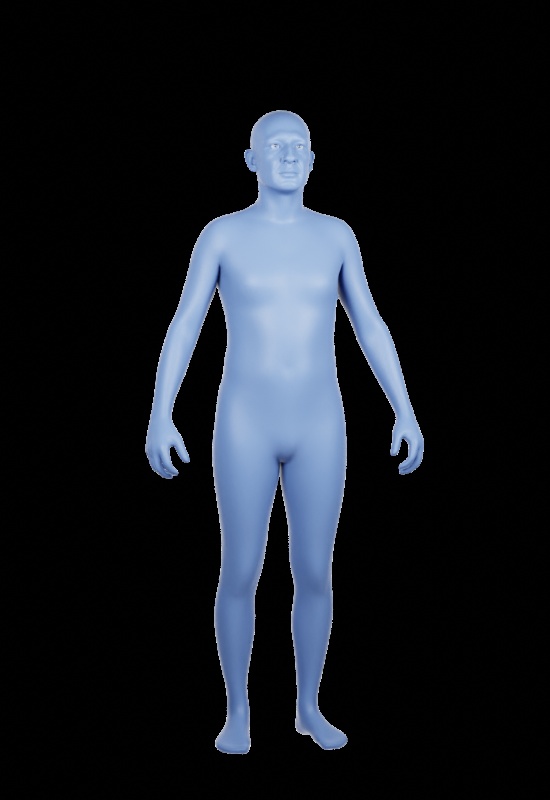}
        Identity
    \end{subfigure}
    \begin{subfigure}{0.155\linewidth}
        \centering \tiny
        \includegraphics[width=\linewidth]{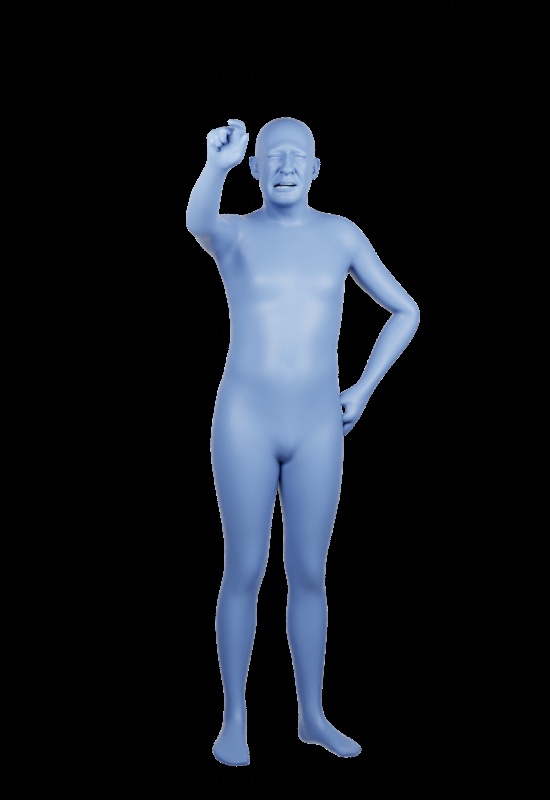}
        + Pose
    \end{subfigure}
    \begin{subfigure}{0.155\linewidth}
        \centering \tiny
        \includegraphics[width=\linewidth]{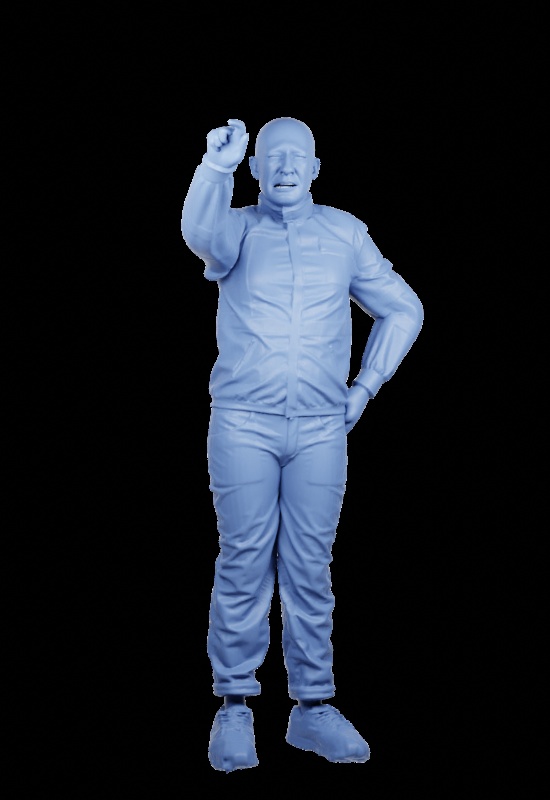}
        + Clothing
    \end{subfigure}
    \begin{subfigure}{0.155\linewidth}
        \centering \tiny
        \includegraphics[width=\linewidth]{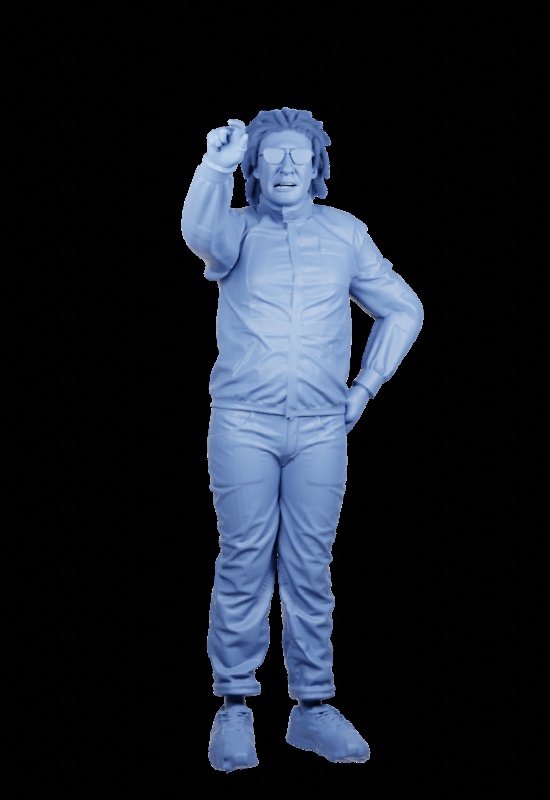}
        + Hair/Accessories
    \end{subfigure}
    \begin{subfigure}{0.155\linewidth}
        \centering \tiny
        \includegraphics[width=\linewidth]{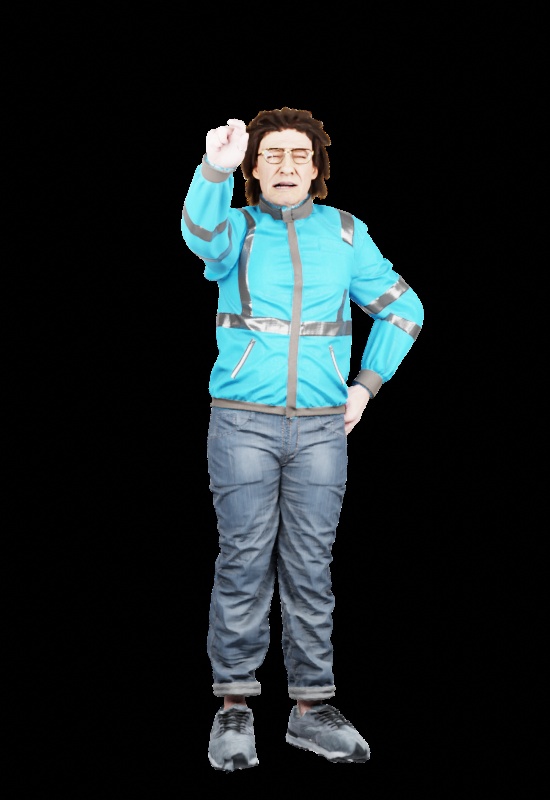}
        + Texture
    \end{subfigure}
    \begin{subfigure}{0.155\linewidth}
        \centering \tiny
        \includegraphics[width=\linewidth]{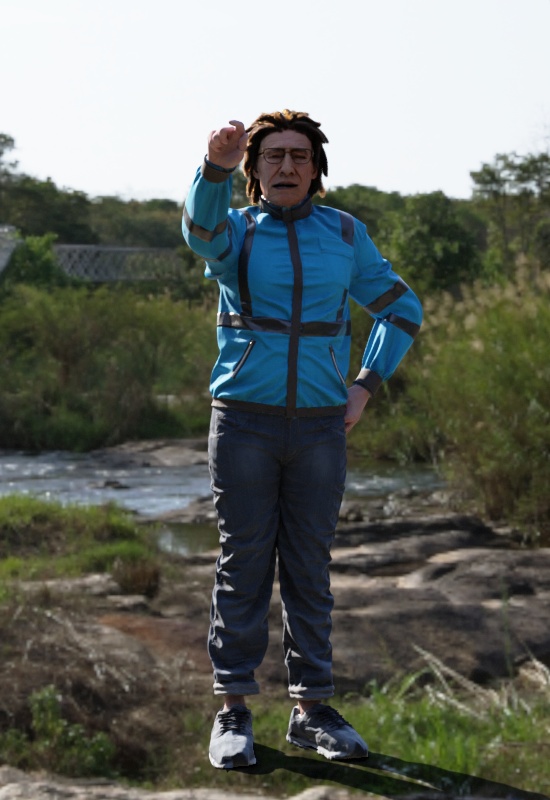}
        + Environment
    \end{subfigure}
    \\
    \begin{subfigure}{0.155\linewidth}
        \centering \tiny
        \includegraphics[width=\linewidth]{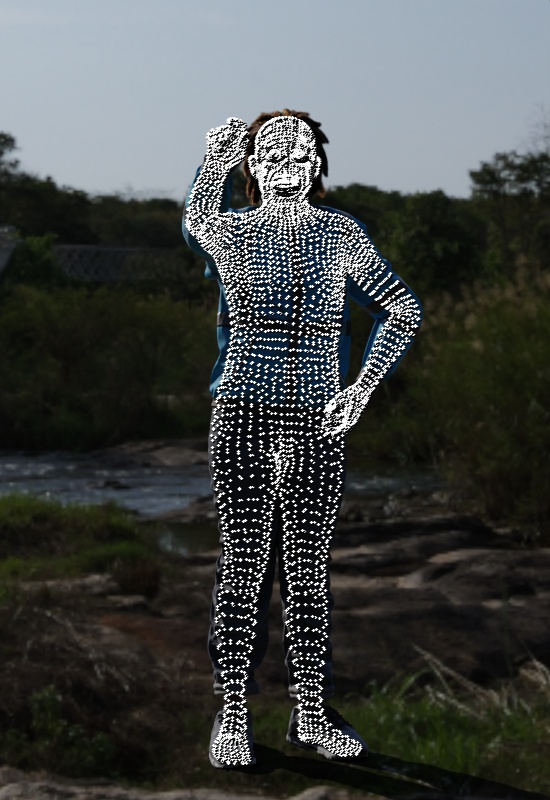}
        Vertices
    \end{subfigure}
    \begin{subfigure}{0.155\linewidth}
        \centering \tiny
        \includegraphics[width=\linewidth]{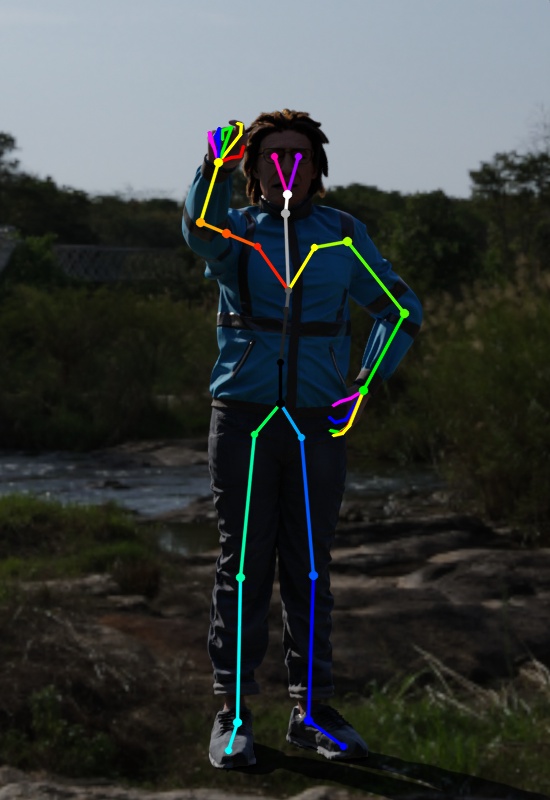}
        Skeleton
    \end{subfigure}
    \begin{subfigure}{0.155\linewidth}
        \centering \tiny
        \includegraphics[width=\linewidth]{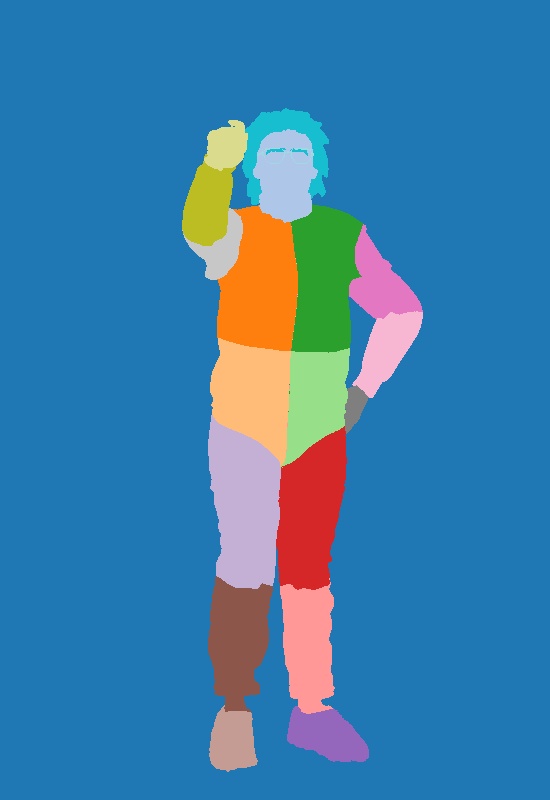}
        Segmentation
    \end{subfigure}
    \begin{subfigure}{0.155\linewidth}
        \centering \tiny
        \includegraphics[width=\linewidth]{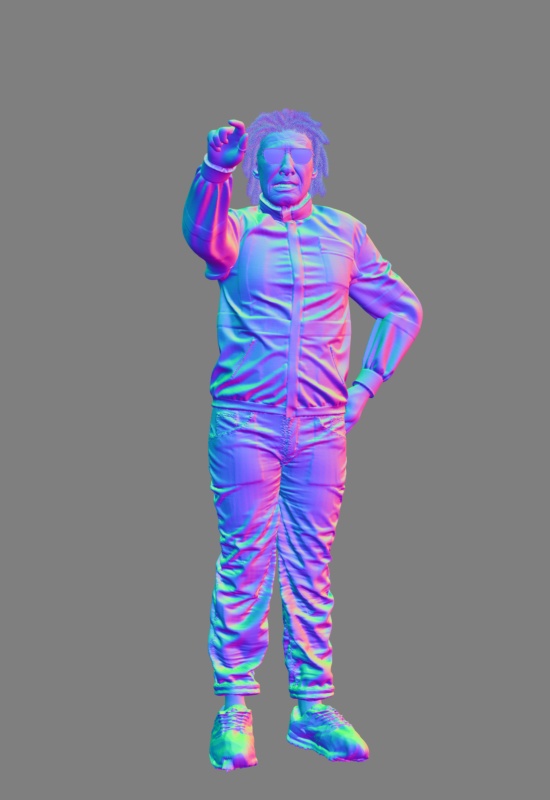}
        Normals
    \end{subfigure}
    \begin{subfigure}{0.155\linewidth}
        \centering \tiny
        \includegraphics[width=\linewidth]{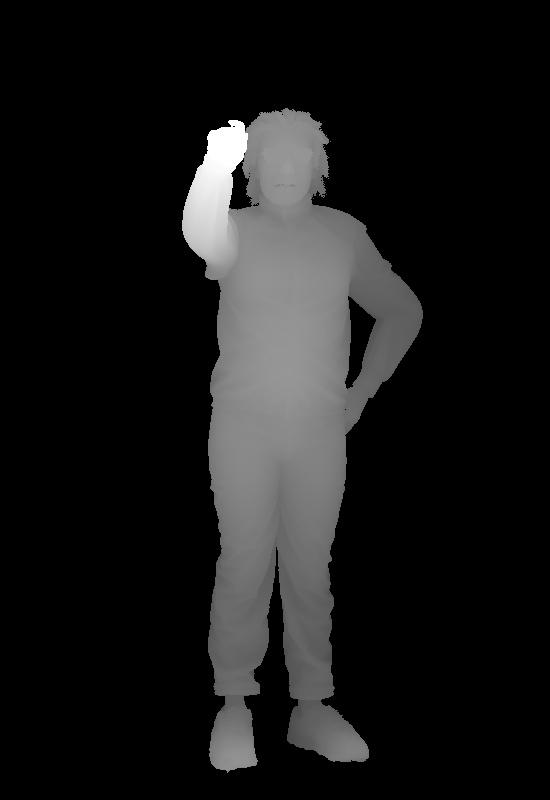}
        Depth
    \end{subfigure}
    \begin{subfigure}{0.155\linewidth}
        \centering \tiny
        \includegraphics[width=\linewidth]{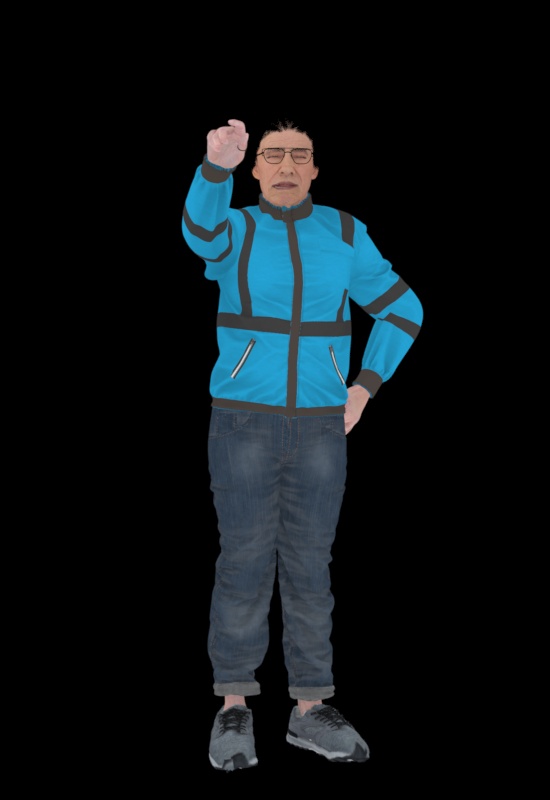}
        Albedo
    \end{subfigure}
    \caption{Construction of synthetic image by sampling random identity and pose for the \SOMA{} parametric model, clothing, hair and accessory assets, textures, shaders and HDRI environments from our asset library (top row). From this we can produce a highly realistic image and a number of corresponding ground-truth annotations (bottom row). \label{fig:construction}}
\end{figure}

\begin{figure}
  \centering
\includegraphics[width=0.325\linewidth]{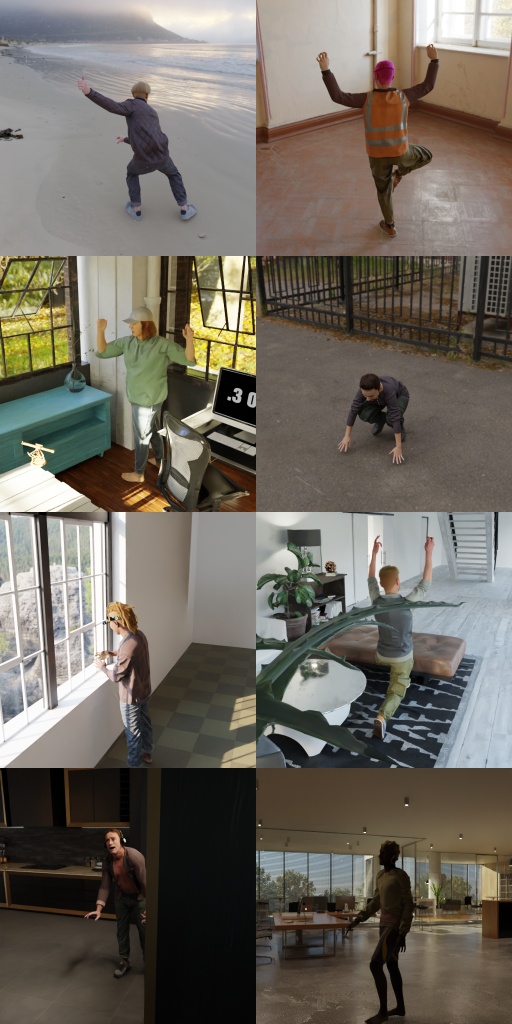}
\hfill
\includegraphics[width=0.325\linewidth]{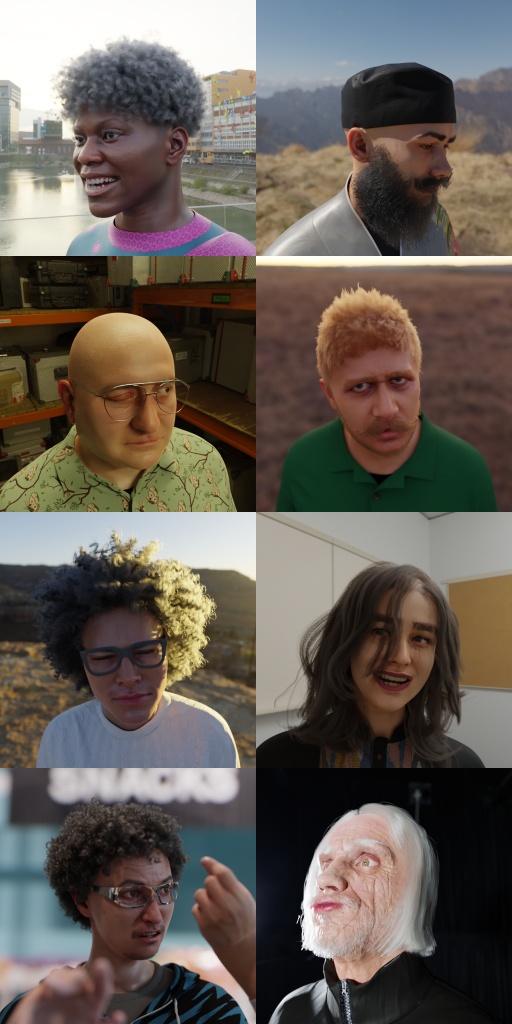}
\hfill
\includegraphics[width=0.325\linewidth]{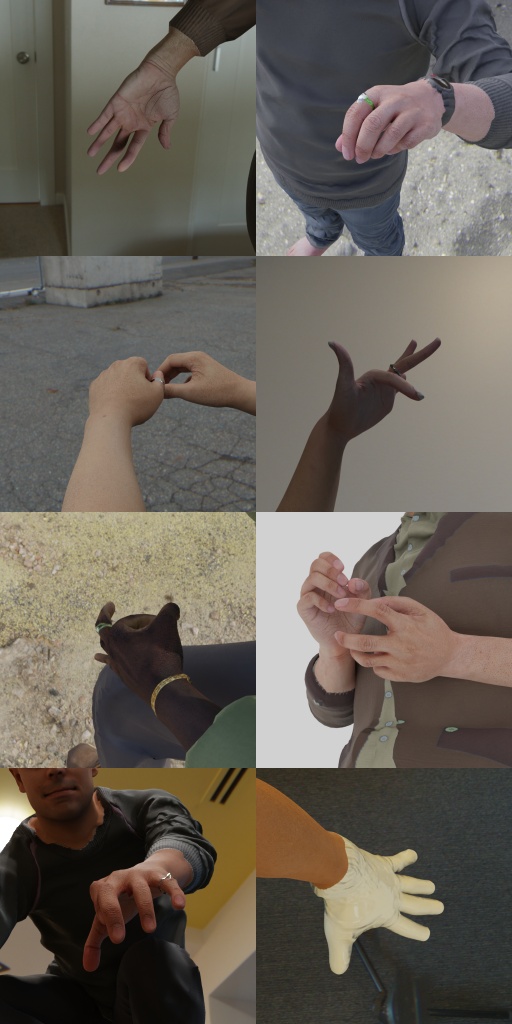}
  \caption{Example images from the \SynthBody{}, \SynthFace{} and \SynthHand{} datasets, all of our neural networks are trained exclusively on these synthetic datasets.}
  \label{fig:synth_data}
\end{figure}

The pipeline starts with a parametric model which captures face and body shape, body and hand pose and facial expression, including articulation of the tongue.
To construct this, we build on the popular \SMPLH{} body model~\cite{romero2017embodied} and the face model of \citet{wood2021fake}.
We then add texture, hair and clothing from a large and diverse asset library.
Finally, we situate the body in an indoor 3D scene or an outdoor scene with environmental lighting from an HDR image~\cite{debevec2006image}.
We render the scene using the Cycles rendering engine~\cite{CyclesRenderer}.
\autoref{fig:construction} shows how we construct a render, as well as some example ground-truth annotations output by our pipeline.
Examples of images rendered using our pipeline can be seen in \autoref{fig:synth_data}.
All data originating from captures of real people used in the pipeline is obtained with explicit consent.
Below we give a brief summary of the parametric model and data generation pipeline; for greater detail, please consult the supplementary material.

\subsection{Parametric Human Model}
\label{sec:model}
We improve the face model of \citet{wood2021fake} by refining and expanding the training data (as described below) and adding tongue articulation.
We combine this updated face model with the \SMPLH{} body model~\cite{romero2017embodied} to produce a unified parametric model of the full human with a high degree of control and expressivity, we call this \SOMA{} (from the Greek for `body').
The \SOMA{} mesh vertex positions are defined by mesh generating function
$\mathcal{M}(\boldsymbol{\beta}, \boldsymbol{\gamma}, \boldsymbol{\zeta}, \boldsymbol{\psi}, \boldsymbol{\theta})$ 
which takes parameters
$\boldsymbol{\beta}\in\mathbb{R}^{300}$ for body shape,
$\boldsymbol{\gamma}\in\mathbb{R}^{256}$ for face shape,
$\boldsymbol{\zeta}\in\mathbb{R}^{9}$ for hand shape,
$\boldsymbol{\psi}\in\mathbb{R}^{224}$ for expression, and
$\boldsymbol{\theta}\in\mathbb{R}^{54\times3}$ for pose. %\TCnote{body pose?}. includes neck/head/eyes so not really just body
% Tongue articulation is achieved through addition of 12 blendshapes solely to control the movement of the tongue.
A complete mathematical definition of the model can be found in the supplementary material.

\begin{figure}
    \centering
    \includegraphics[width=\linewidth]{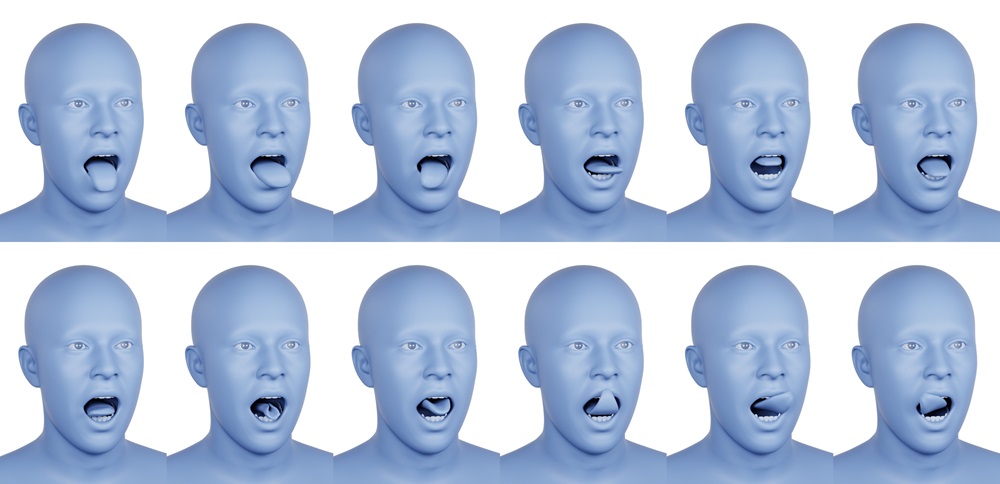}
    \caption{Twelve tongue blendshapes included in the \SOMA{} model, visualized with mouth opening blendshapes also activated. These blendshapes allow for significant coverage of the range-of-motion of the tongue.}
    \label{fig:tongue}
\end{figure}

 Tongue articulation is achieved through addition of 12 blendshapes solely to control the movement of the tongue, covering a large proportion of possible articulations of the tongue. 
Given that the tongue is such a complex muscle, it would be almost impossible to capture its \emph{full} range of motion using a practical number of blendshapes.
Rather than introducing a complex joint based rig which would involve a more challenging fitting procedure, we instead compromise on the expressivity of our model and select a minimal set of blendshapes that provide reasonable coverage, these are visualized in \autoref{fig:tongue}.

\paragraph{Face Shape Basis}
\label{sec:face_id}
We retrain the model of \citet{wood2021fake} with a training set of $N = 1040$ scans, selected to enhance the diversity of the data across multiple demographic axes.
We find the model of \citet{wood2021fake} does not generalize well to populations with thinner or thicker lips and analysis shows inconsistencies in the semantic correspondence of the topology around the lips in the original training data.
As such, we reprocess the training data to enforce consistency in the mouth region using our method and commercial retopologization software~\cite{Wrap3} in a semi-automated approach.
We then carry out photometric refinement~\cite{Nicolet2021Large} to better capture high-frequency features such as wrinkles.
For further details of the improvements made to the training data, see the supplementary material.
To obtain the shape basis we jointly fit the basis and parameters, $[\boldsymbol{\gamma}_0, ..., \boldsymbol{\gamma}_N]$, to the vertices of the training scans, while other components of the model are kept constant, using vector Adam~\cite{ling2022vectoradam}.

\subsection{Asset Library}
\label{sec:asset_lib}

\begin{figure}
    \centering
    \centering
    \includegraphics[height=0.43\linewidth]{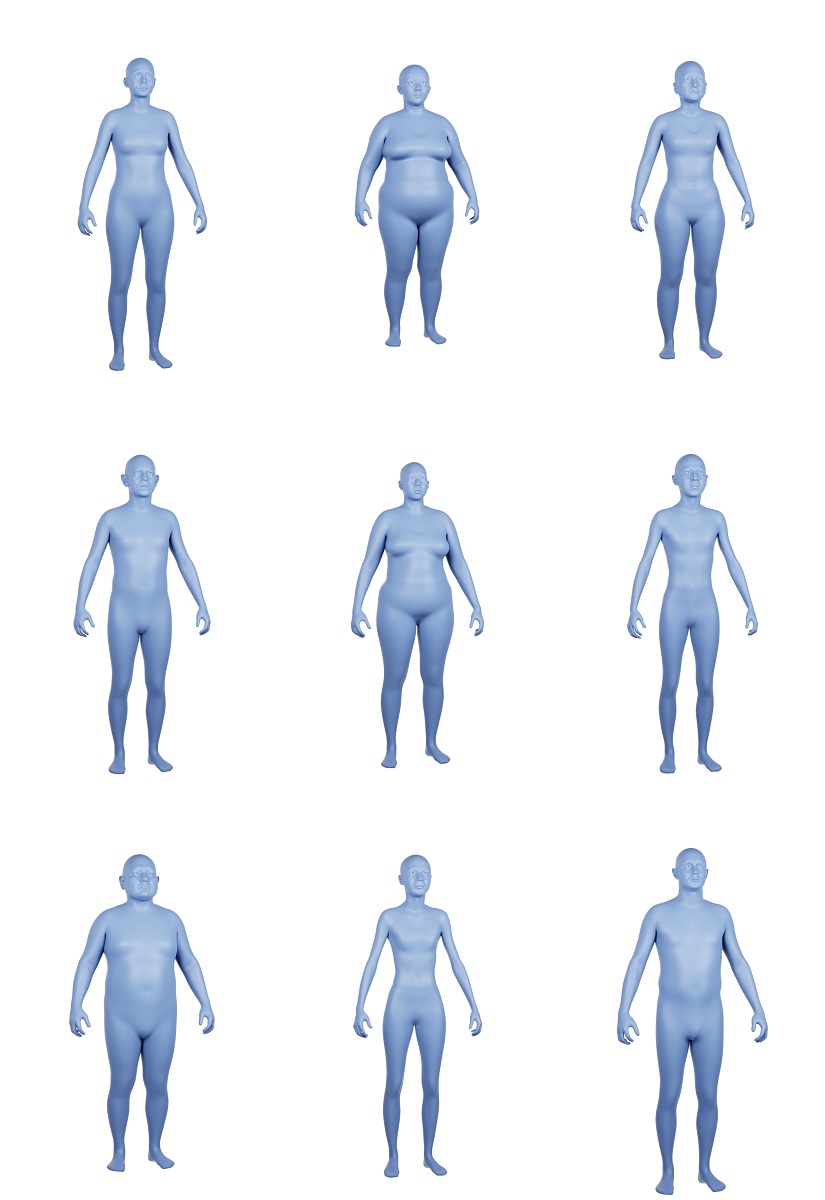}
    \hfill
    \includegraphics[height=0.43\linewidth]{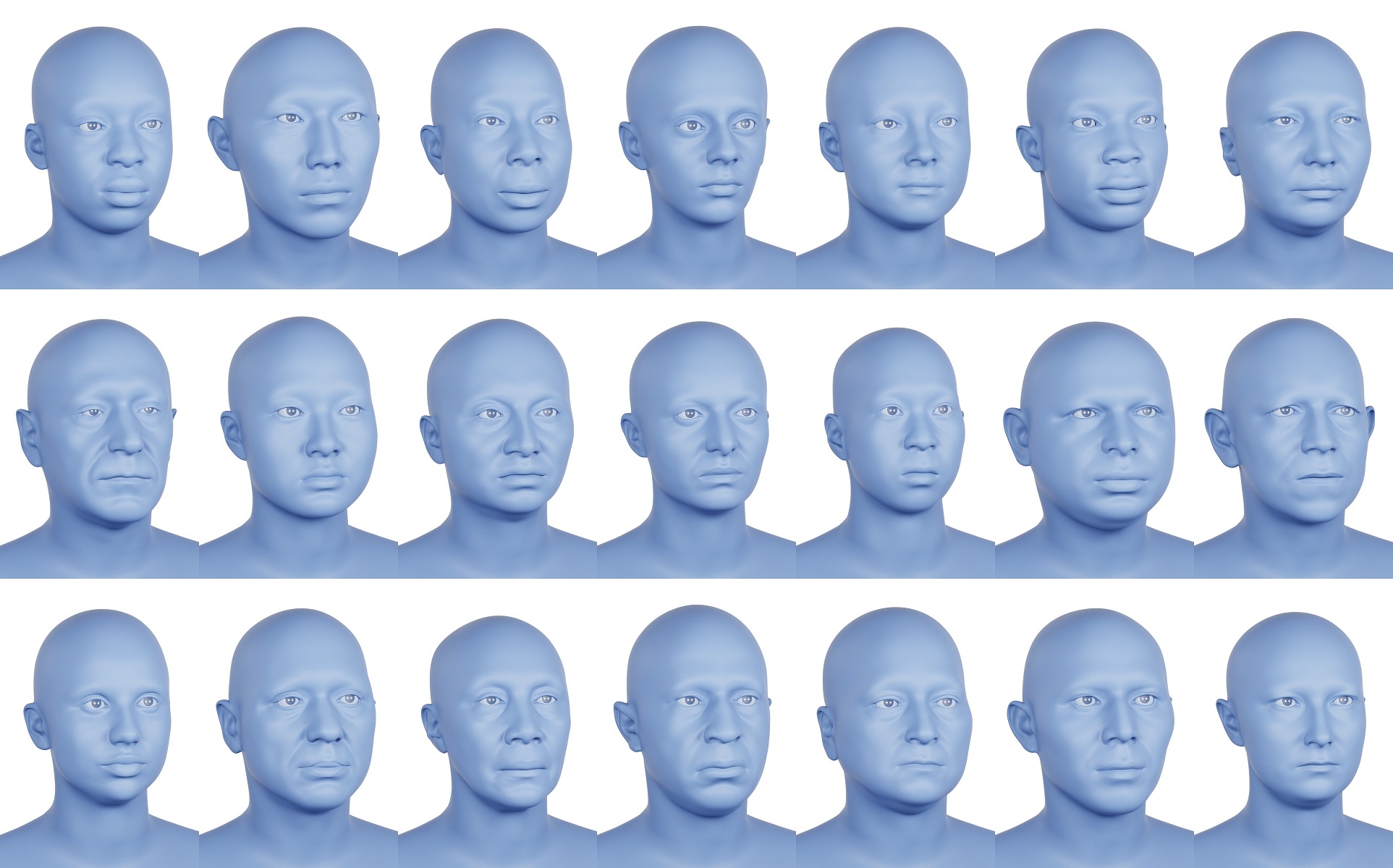}
    \caption{\label{fig:id_samples}Body and face shape samples. We sample at random from male, female and ungendered GMMs. The \SOMA{} model itself has no explicit concept of gender.}
\end{figure}

As outlined in \autoref{fig:construction}, we build up a unique render by sampling shape, pose, hair and clothing assets, texture and environment.
Face and body shape are sampled from GMMs fit to, or directly from, the face training data and a library of 3572 body scans from various sources~\cite{yang2014semantic,mahmood2019amass,renderpeople,3dscanstore}.
Example identities are shown in \autoref{fig:id_samples}.

\begin{figure}
    \centering
    \includegraphics[height=0.43\linewidth]{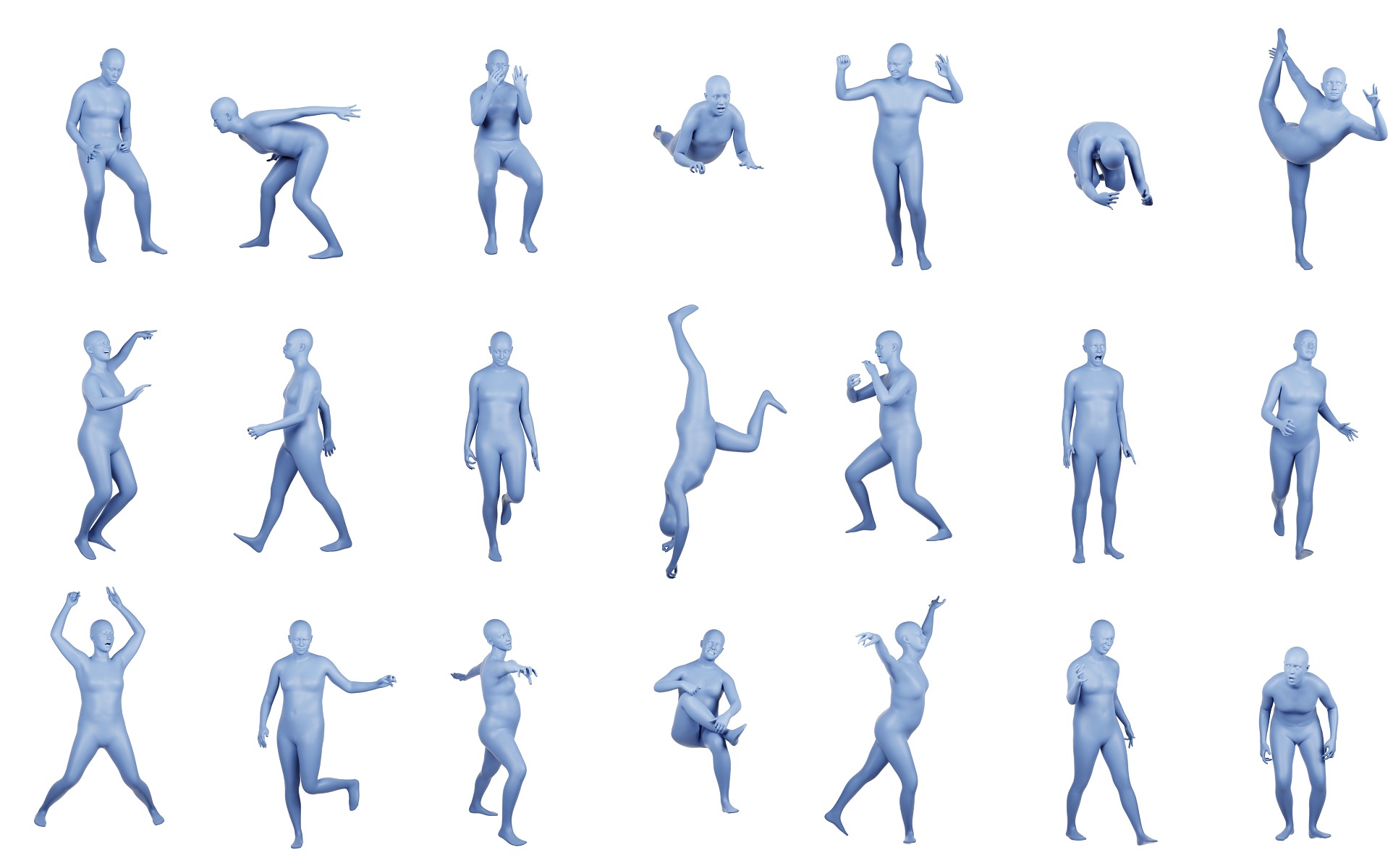}
    \hfill
    \includegraphics[height=0.43\linewidth]{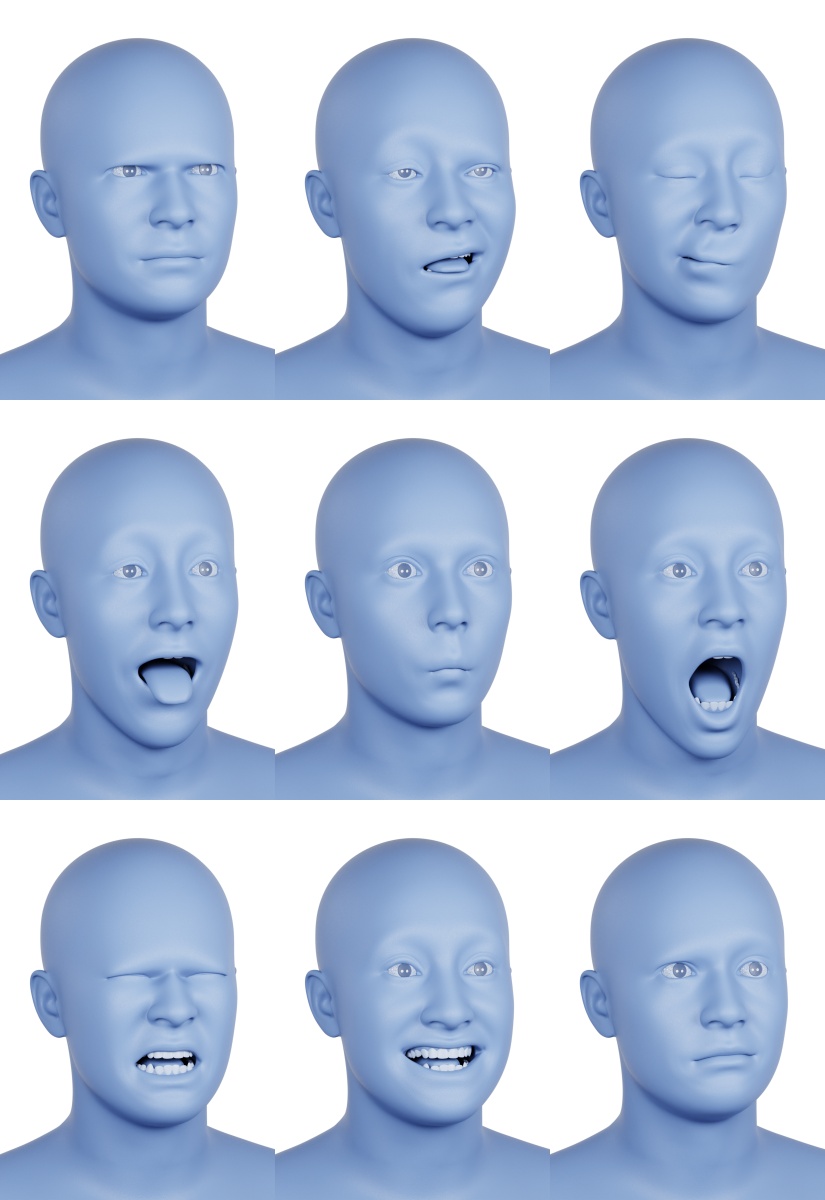}
    \caption{Pose and expression library samples. If not all data is present in the original capture, we splice together body, hand and face pose and expression from separate sequences.}
    \label{fig:pose_samples}
\end{figure}

For facial expression we re-register the library of \citet{wood2021fake} to improve the quality and add tongue articulation, as well as adding additional artist generated sequences including the tongue.
For body and hand pose we sample from a large library including AMASS~\cite{mahmood2019amass}, MANO~\cite{romero2017embodied} and other motion capture data processed using MoSh~\cite{loper2014mosh}.
Example poses and expressions are shown in \autoref{fig:pose_samples}.

Once we have a posed body mesh, we add hair, clothing and accessories and render the result in a realistic environment.
For this, we construct a large library of artist-created assets, from which we can sample to produce a diverse range of realistic scenes.
We use displacement maps to model basic clothing, and mesh assets for glasses and headwear.
Hair is represented using strands, with separate assets for the scalp hair, facial hair, eyebrows and eyelashes.
We use a mixture of HDRIs and 3D room environments to provide a wide variety of environmental lighting and background content.
Example assets can be seen in \autoref{fig:asset_library} and further details of the asset library are given in the supplementary material.

\begin{figure}
    \centering
    \includegraphics[width=\linewidth]{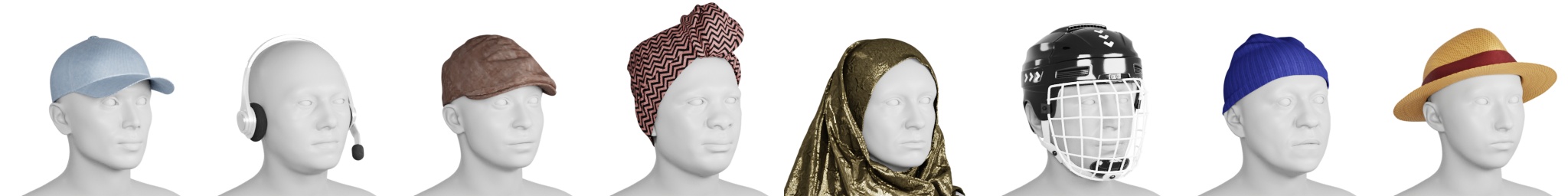}
    \includegraphics[width=\linewidth]{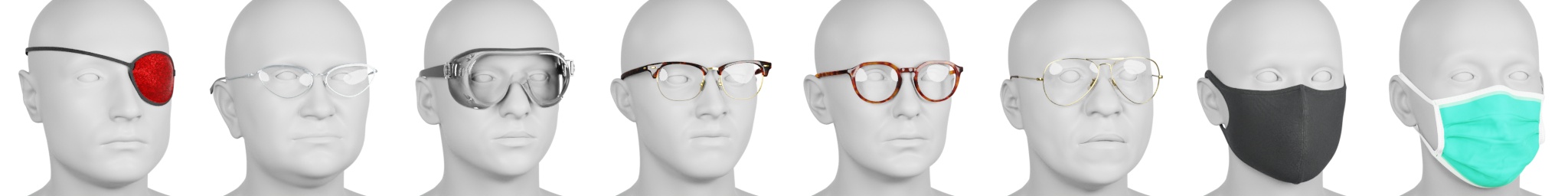}
    \includegraphics[width=\linewidth]{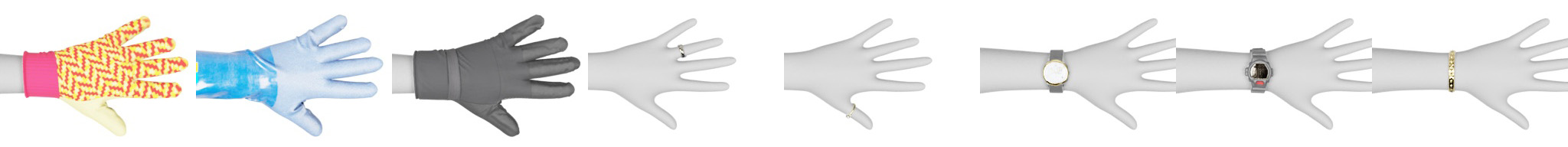}
    \includegraphics[width=\linewidth]{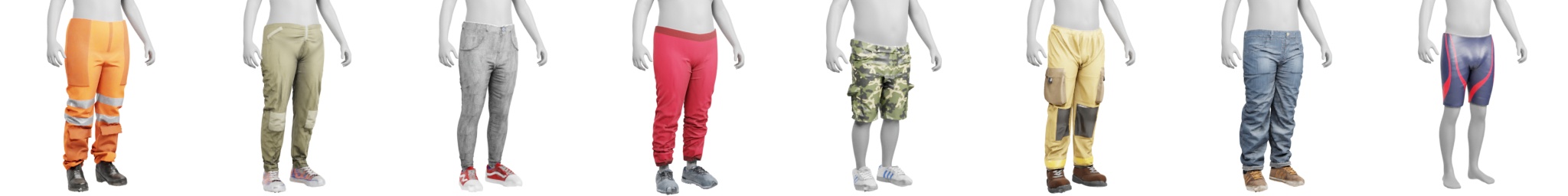}
    \includegraphics[width=\linewidth]{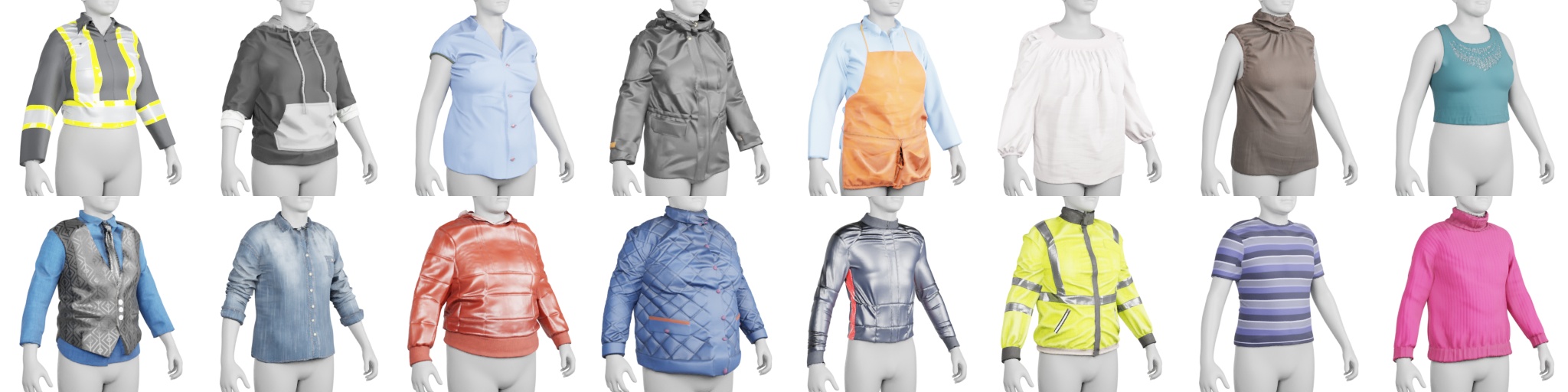}
    \includegraphics[width=\linewidth]{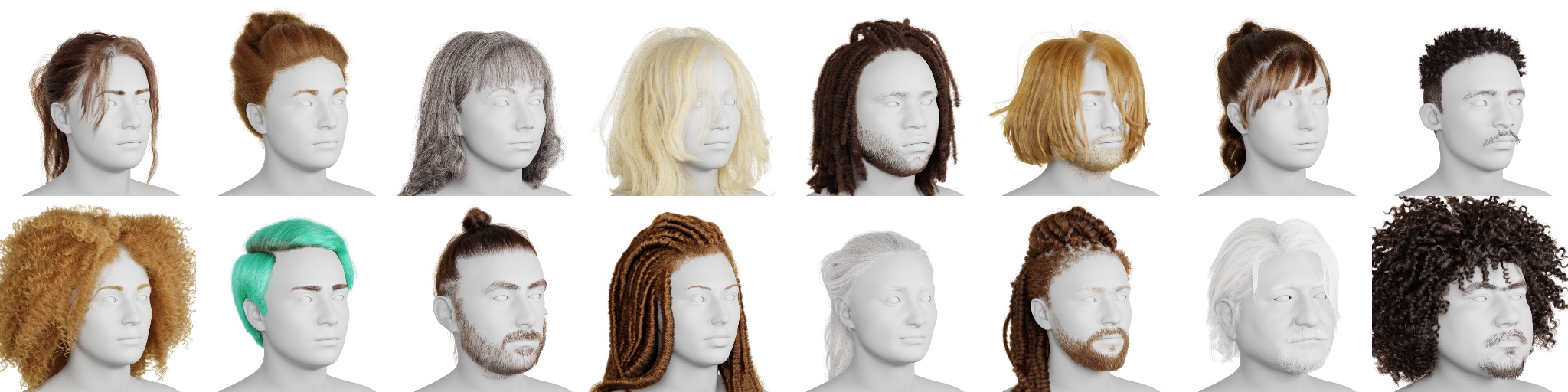}
    \caption{Selection of assets from our library of accessories, clothing and hair. Displacement-based assets automatically adapt to variation in body shape, mesh-based assets and hair are adapted using lattice warping.}
    \label{fig:asset_library}
\end{figure}

\subsection{Training Datasets}
\label{sec:datasets}
To train the DNNs used in our method, we generate three datasets, each containing approximately 100,000 images, one featuring the entire body, one the face, and one the left hand.
We call these datasets \SynthBody{}, \SynthFace{} and \SynthHand{}.
These datasets will be released publicly with sparse annotations upon publication.
Example images from each dataset can be seen in \autoref{fig:synth_data}.
\SynthBody{} and \SynthFace{} are rendered at $512\times512$ pixel resolution, both include 2D landmark and semantic segmentation annotations. 
\SynthBody{} also includes 3D joint locations and \SMPLH{} pose and shape parameters while \SynthFace{} also includes head pose annotations.
\SynthHand{} is rendered at $256\times256$ pixel resolution and includes 2D and 3D joint locations and MANO parameter annotations.
All datasets also include camera parameters, download instructions can be found at {\color{niceblue}{\url{https://aka.ms/SynthMoCap}}}.
Note that all results presented below are from DNNs trained \emph{exclusively} on these synthetic datasets.

\section{Performance Capture}

\begin{figure}
    \centering
    \includegraphics[width=\linewidth]{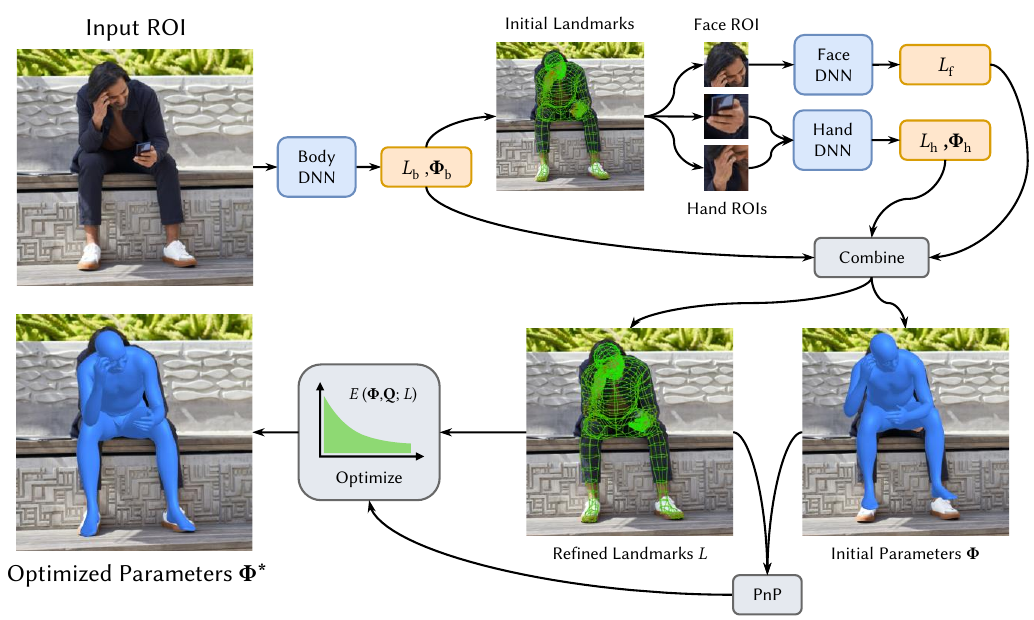}
    \caption{Overview of our method architecture. From an input image we regress landmarks and initial model parameters, we refine these for the face and hands and minimize reprojection error to obtain the output parameters.}
    \label{fig:method_overview}
\end{figure}

We employ a multi-stage technique to fit the \SOMA{} model to image or video data.
The first stage regresses information including 2D landmarks from the input image(s) using DNNs trained on synthetic data generated using the pipeline described in \autoref{sec:synth_data}.
The second stage uses conventional optimization to fit our parametric model to the observed 2D landmarks by minimizing the reprojection error of the 3D mesh vertices into the image(s).
The optimization process involves a series of energy terms to encode prior knowledge of the scenario, as detailed below.
We find that using 2D landmarks alone does not lead to robust results, and that it is critical to initialize the optimization with a good approximation of the body shape and pose.
As such, we also train the DNNs to predict body pose and shape from the input image(s).
The method can therefore also be run in real-time without using optimization to provide a solution for pre-visualization of results.
\autoref{fig:method_overview} provides a graphical representation of our pipeline.

% The first stage involves regression of dense 2D landmarks, body and hand pose and body shape using DNNs trained on synthetic data generated using the pipeline described in \autoref{sec:synth_data}.
% The formulation of this approach naturally extends to arbitrary camera setups as the ML component is concerned solely with image-space predictions. 
% We also find it critical , which we also regress directly from the input image(s) using DNNs.
% Following \citet{wood20223d}, we make use of probabilistic landmarks which enable the fitter to exploit per-landmark confidence measures to account for occlusions or other issues in the input imagery.
 % and \autoref{fig:teaser} shows qualitative results of our method.

\subsection{Neural Networks}
\label{sec:ldmks}
\begin{figure}
    \centering
    \includegraphics[height=0.35\linewidth]{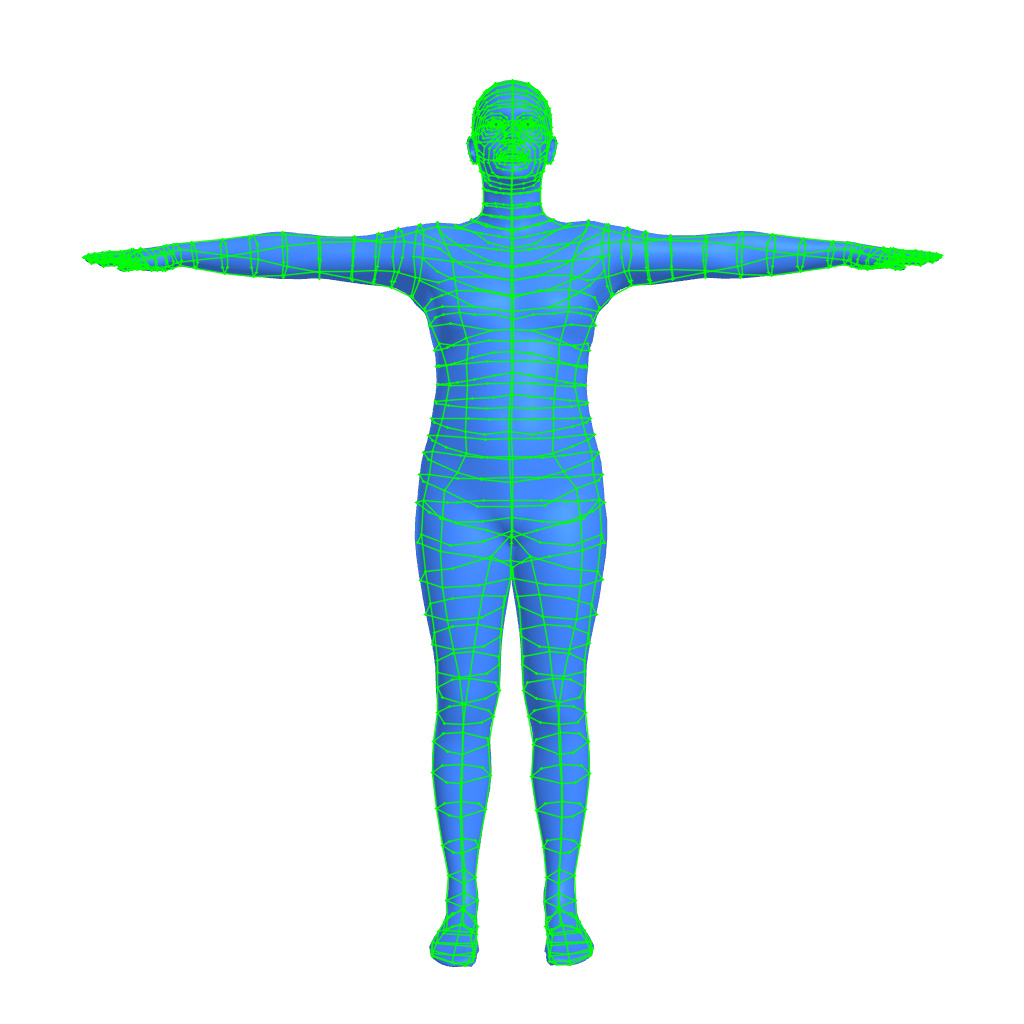}
    \hfill
    \includegraphics[height=0.35\linewidth]{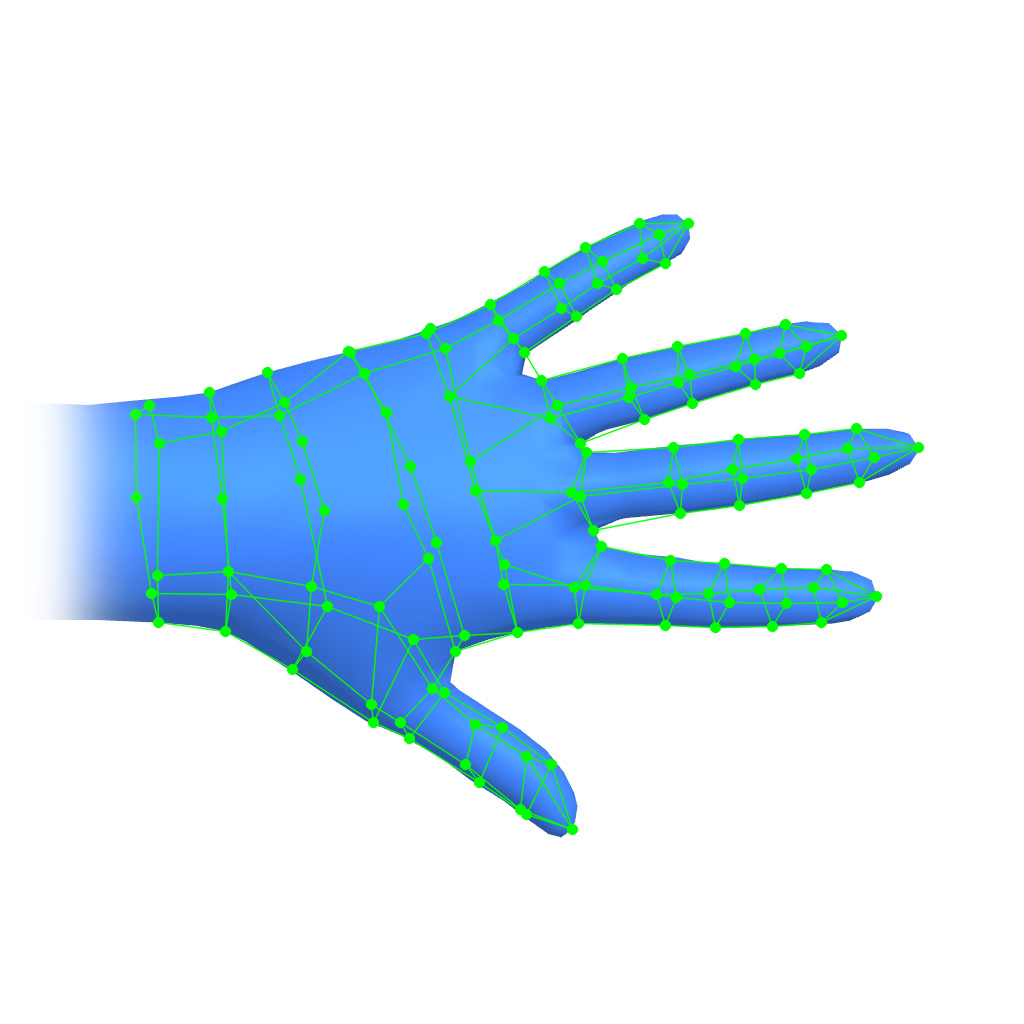}
    \hfill
    \includegraphics[height=0.35\linewidth]{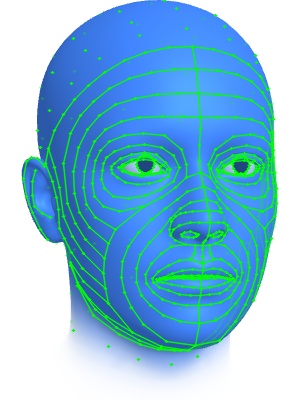}
    \caption{
        Dense landmark definitions for the body, hands and face.
        We use 1428 landmarks for the full body, 141 for each hand and 744 for the face.
        Face landmarks include 16 points on the limbus boundary in each eye, 12 points on the teeth and 9 points on the tongue.
        \label{fig:landmark_defs}
    }
\end{figure}

Our model-fitting pipeline relies on full-body dense landmarks as well as an initialization for pose and shape.
High density of the landmarks is critical to ensure accurate shape reconstruction of the surface of the human~\cite{wood20223d}; we show our landmark set on the template meshes in \autoref{fig:landmark_defs}.
% \TCnote{We don't show tongue landmarks in this figure, but we mentioned tongue a lot in the claims.} 
We observe that DNNs trained to directly regress parametric model parameters from images are very robust and learn a strong implicit prior over the body pose, though struggle to achieve very precise results on their own.
Starting from a reasonable estimate of body pose, however, helps the optimizer to find a better solution, particularly in the under-constrained single-view case.

% While there are some similarities in network architecture design, we train three separate models, each with synthetic training data rendered for capturing variations in each scenario. 
% These DNNs are trained to predict dense landmarks as the main task, along with the option to predict the pose and shape parameters as additional tasks.
% \TBnote{maybe combine both sentences to link better that we choose three separate models explicitly, e.g. "hands, and face, because we found a single DNN was unable to achieve.... This may be due to the low..."}
We find a single DNN is unable to capture adequately high-fidelity results for the face and hands, perhaps due to the limited resolution in a region of interest (ROI) including the whole body. %, or due to high variation in position of these features in the image.
We therefore train three separate DNNs for the body, hands, and face; each model takes ROIs cropped closely to the relevant body part as input.
During training, the ROI is estimated based on the ground-truth landmarks, at inference time, a detector is used to estimate the full-body ROI and sub-ROIs are extracted based on the initial landmark prediction. 
For further details on detection and tracking, details of the training procedure and experiments supporting the above design choices, please see the supplementary material.

\begin{figure}
    \centering
    \includegraphics[width=\linewidth]{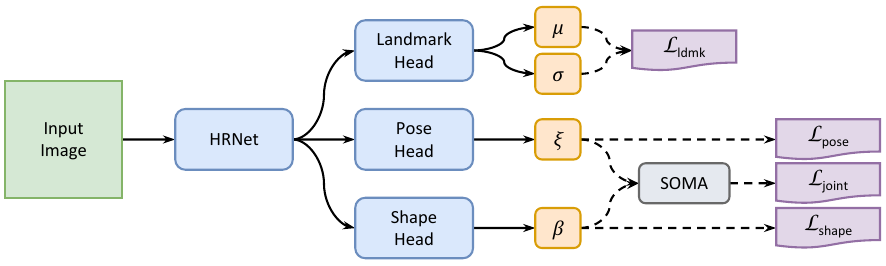}
    \caption{Landmark, pose, and shape estimator DNN architecture. Used in full for the body DNN, without shape head for the hand DNN, and with only landmark head for the face DNN. 
    $\boldsymbol{\theta} = n(\boldsymbol{\xi})$ is input to the \SOMA{} model.
    }
    \label{fig:dnn}
\end{figure}

\paragraph{Architecture and Losses.} 
Following \citet{zhang2023pymaf}, we use a HRNet backbone~\cite{sun2019deep}, pretrained on ImageNet~\cite{deng2009imagenet}, to extract visual features from the input image. 
As shown in \autoref{fig:dnn}, the compressed representation computed with HRNet then serves as input to three MLP-based heads to predict probabilistic landmarks, pose, and shape. 

For predicted landmarks, $\mathbf{L}$, we model each as a circular Gaussian, $L_i = \{\boldsymbol{\mu}_i, \sigma_i\}$, where $\boldsymbol{\mu}_i = [x_i, y_i]$ is the expected 2D position of that landmark, and $\sigma_i$ represents the uncertainty of the prediction~\cite{wood20223d}. 
We denote ground-truth landmark positions $\boldsymbol{\mu}'_i = [x_i', y_i']$.
For training we use Gaussian negative log likelihood loss with per-landmark weights, $\lambda_i$,
\begin{equation}
\mathcal{L}_{\textrm{ldmk}} = \sum_{i=1}^{|\mathbf{L}|} \lambda_i \Bigg(%
    \log\left(\sigma_i^2\right) + 
    \frac{\| \boldsymbol{\mu}_{i} - \boldsymbol{{\mu}'_{i}} \| ^2}{2 \sigma_{i}^2}
\Bigg).
\label{eqn:gnll}
\end{equation}

We represent pose parameters output by the pose head as 6D rotations following~\citet{zhou2019continuity}, and compute the $L_1$ loss $\mathcal{L}_{\textrm{pose}} = \|\boldsymbol{\xi} - \boldsymbol{\xi}'\|_1$ between the ground-truth rotations $\boldsymbol{\xi}'$ and the predicted rotations $\boldsymbol{\xi}$.
We write the forward kinematics of the \SOMA{} model as $\mathbf{J}_T, \mathbf{J}_R = H(\boldsymbol{\theta}; \boldsymbol{\beta}, \boldsymbol{\gamma}, \boldsymbol{\zeta})$, where $H$ computes world-space joint translations $\mathbf{J}_T$ and rotations $\mathbf{J}_R$ for the given pose, $\boldsymbol{\theta}$, and shape parameters, $\boldsymbol{\beta}, \boldsymbol{\gamma}, \boldsymbol{\zeta}$.
Then we have the ground-truth joint transforms $\mathbf{P}_T', \mathbf{P}_R' = H(\boldsymbol{\theta}'; \boldsymbol{\beta}', \boldsymbol{\gamma}', \boldsymbol{\zeta}')$ and the predicted joint transforms $\mathbf{P}_T, \mathbf{P}_R = H(n(\boldsymbol{\xi}); \boldsymbol{\beta}, \boldsymbol{\gamma}', \boldsymbol{\zeta}')$ where $n$ maps 6D rotations to axis-angles.
We use these to compute a pose reconstruction loss decomposed into two parts:
\begin{equation}
\begin{split}
    \mathcal{L}_{\textrm{joint}_T} = \frac{1}{|\mathbf{J}|} \sum_{j=1}^{|\mathbf{J}|}{\|\mathbf{P}^j_T - \mathbf{P}_T'^j\|_1}
\end{split}
\label{eqn:pose_P_t}
\end{equation}
\begin{equation}
\begin{split}
    \mathcal{L}_{\textrm{joint}_R}  = \frac{1}{|\mathbf{J}|} \sum_{j=1}^{|\mathbf{J}|}{\arccos\Bigg(\frac{tr(\mathbf{P}_R^j(\mathbf{P}_R'^j)^T)-1}{2}\Bigg)}
\end{split}
\label{eqn:pose_P_r}
\end{equation}
where $|\mathbf{J}|$ is the number of joints and $tr(.)$ is the trace operation.

The shape head predicts \SOMA{} shape parameters for which we compute the $L_1$ error between the ground truth shape parameters and the prediction, $\mathcal{L}_{\textrm{shape}} = \|\boldsymbol{\beta} - \boldsymbol{\beta}'\|_1$. 
Note that shape parameters are also implicitly optimized in \autoref{eqn:pose_P_t}.
Our final loss function is therefore
\begin{equation}
    \mathcal{L} = \alpha_{l} \mathcal{L}_{\textrm{ldmk}} + \alpha_{p} \mathcal{L}_{\textrm{pose}} + \alpha_{t}\mathcal{L}_{\textrm{joint}_T} + \alpha_{r} \mathcal{L}_{\textrm{joint}_R} + \alpha_{s} \mathcal{L}_{\textrm{shape}}
\end{equation}
where $\alpha_l, \alpha_p, \alpha_{t}, \alpha_{r}, \alpha_s$ are the weights assigned to each loss term.

\paragraph{Models.} 
We train the body DNN on the \SynthBody{} dataset using the full architecture and losses described above.
The model predicts 1428 landmarks on the body (see \autoref{fig:landmark_defs}), body pose, $\boldsymbol{\xi} \in \mathbb{R}^{23\times 6}$, (excluding the pelvis and hands) 
and the first 10 components of the \SOMA{} body shape parameters which explain the majority of shape variance.
The hand DNN is trained on the \SynthHand{} dataset with the shape head and corresponding loss omitted ($\alpha_s = 0$).
We train only a single DNN for the left hand. 
During inference, we mirror the input image and predicted landmarks in the $x$-direction and mirror the predicted pose around the $y$-axis for the right hand. 
The hand DNN predicts 141 dense landmarks on the hand (see \autoref{fig:landmark_defs}) and 15 joints for the hand pose (excluding the wrist).
The face DNN is trained on the \SynthFace{} dataset and directly regresses 744 landmarks on the face (see \autoref{fig:landmark_defs}).
We omit the pose and shape heads ($\alpha_s = \alpha_p = \alpha_t = \alpha_p = 0$) as variation in the facial appearance due to shape and expression is of far lower magnitude than for the body, and the optimizer is able to fully determine these in the model fitting stage.
% \BLnote{It would be great if you could go into more detail about this, either here or in the supp, especially since you mention this again in just a couple of paragraphs.}

\subsection{Model Fitting}
\label{sec:model_fitting}
We use L-BFGS~\cite{liu1989limited} to find the optimal parameters for the \SOMA{} model to explain the observed landmarks, starting from our pose and shape initialization.
If multiple views are present then we regress pose and shape from the view with the lowest mean sigma of predicted probabilistic landmarks.
We first initialize the global position and orientation using PnP~\cite{fischler1981random}, then employ a three-stage fitting procedure: (1) finalizing the PnP initialization by optimizing only the translation and rotation (2) fitting the \SOMA{} model by optimizing pose, expression and shape parameters, and optionally (3) refining the camera extrinsics (multi-view) or focal length (single-view), assuming camera calibration is not provided.
% If a very precise calibration is available then (3) is unnecessary, but if calibration was not provided or it was imprecise then this refinement can lead to significantly improved results, though can also introduce ambiguity of scale.

In multi-view scenarios where no camera calibration is provided, the camera positions and orientations can be estimated accurately using PnP with the template 3D head mesh and the 2D face landmarks for each view~\cite{szymanowicz2022photo}.
The head is used instead of the full body due to the minimal impact on overall shape caused by variation in pose and expression, leading to more reliable camera positioning.
Unless provided, we assume no camera distortion is present and that the optical center is aligned to the image center.

The objective function for the optimization is:
\begin{equation}
\resizebox{0.91\hsize}{!}{%
    $E(\boldsymbol{\Phi},\mathbf{Q}; \mathbf{L}) =
    \underbrace{E_{
        \vphantom{identity}%
        \textrm{ldmks}}
    }_{\textrm{Data term}}
    +
    \underbrace{
        E_{\textrm{shape}} +
        E_{\textrm{exp}} +
        E_{\textrm{pose}} +
        E_{\textrm{temp}} +
        E_{\textrm{intsct}} +
        E_{\textrm{cam}}
    }_{\textrm{Regularizers}}$%
}
\end{equation}
where $\boldsymbol{\Phi} = \{\boldsymbol{\beta}, \boldsymbol{\gamma}, \boldsymbol{\zeta}, \boldsymbol{\psi}, \boldsymbol{\theta}\}$ are the \SOMA{} parameters, $\boldsymbol{Q}$ are the camera extrinsics in the multi-view case, or focal length in the single-view case, which are constant through the sequence, and $\mathbf{L}$ are the predicted probabilistic landmarks.
% \paragraph{Landmark Term.}

The only data term is the re-projection error for $\mathbf{L}$.
Critically, we exploit the predicted uncertainty, $\sigma$, to weight the loss for each landmark, so for $F$ frames over $C$ cameras:
\begin{equation}
E_{\textrm{ldmks}} =
  \sum_{i, j, k}^{F, C, |\mathbf{L}|}
  \frac{\| \mathbf{x}_{ijk} - \boldsymbol{\mu}_{ijk} \| ^2}{2 \sigma_{ijk}^2}
\end{equation}
where $[\boldsymbol{\mu}_{ijk}, \sigma_{ijk}]$ is the 2D location and uncertainty predicted for the $k$\textsuperscript{th} landmark, and $\boldsymbol{x}_{ijk}$ is the 2D projection of that landmark on the posed \SOMA{} mesh, $\mathcal{M}(\boldsymbol{\Phi})$, as seen by the $j$\textsuperscript{th} camera in the $i$\textsuperscript{th} frame~\cite{wood20223d}.
This weighting of the loss based on uncertainty of the predicted landmarks is particularly valuable in the full-body scenario where body parts can be easily self-occluded based on the pose or viewing angle.
In the multi-view case, this lets the view with the best visibility of a certain body part take precedence over those with occlusions.
In the single-view case it allows the priors to exert greater influence for occluded body parts, while the landmarks take precedence for unoccluded regions.
% In the multi-view case one view may be unable to see, for example, an arm due to self-occlusion, while it is observed clearly from another perspective.
% The predicted sigma values will let the optimizer fit the confidently observed landmarks with precedence over the inaccurate signal from the occluded view, leading to far more robust fitting results.
% This benefit also applies to the prior terms, particularly so in the monocular case; e.g., if a limb is occluded then we can rely more on the prior terms to push the optimizer towards the most likely pose for that body part, while the landmarks will take precedence for the rest of the body.

\paragraph{Regularizers.}
We use a number of priors to regularize the optimization~\cite{wood20223d,pavlakos2019expressive}.
$E_{\textrm{shape}}$ is a combination of GMM priors;  minimizing the log probability of the given body and face shape, $\boldsymbol{\beta}$ and $\boldsymbol{\gamma}$.
%\TCnote{Which $\beta$ is this? $\beta \in \mathbb{R}^10$ is described half a page before, but I think this is something different!}
For faces we fit a GMM to the shape parameters of the training scans described in \autoref{sec:face_id}, for bodies we fit a GMM to a library of \SMPLH{} identities (\autoref{sec:asset_lib}).
$E_{\textrm{exp}}$ is an $L_1$ loss on expression parameters to promote sparsity, combined with quartic barrier loss to keep coefficients in the range $[0, 1]$.
$E_{\textrm{temp}}$ is an $L_2$ loss on frame-to-frame distance between 3D vertex locations, this helps to reduce jitter that might arise from noise in the predicted 2D landmarks.
$E_{\textrm{intsct}}$ is to prevent intersection of the eyeballs, teeth and tongue with the surface of the face \cite{wood20223d}.
$E_{\textrm{cam}}$ is an $L_2$ loss on the position of the cameras compared to their initial location, to prevent them moving too far from the initialization. % under the assumption that it was approximately correct. 
% This helps to prevent drastic re-scaling of the scene.
% Unlike \citet{pavlakos2019expressive}, we don't perform explicit gender classification which can lead to significant failure cases.
% Instead, 
% If the gender of the subject is known then this can be provided as an additional input to our system, in which case we use GMMs fit to only identities of that gender. %, providing a more applicable prior. 
It is trivial to extend the system with additional priors in the case where we have further information on the scene. 
For example we can add an $L_2$ loss on the height of the subject to reduce 
% completely remove\TCnote{completely remove? That sounds like a prior with zero uncertainty.} any 
ambiguity in scale, or restrict the shape prior based on the gender of the subject.

\paragraph{Neural Pose Prior.}
A key challenge is to ensure the plausibility of the estimated body and hand poses, particularly in single-view scenarios where the problem is ill-posed.
The level of variation and self-occlusion due to body and hand pose is much higher than for facial expression, which we find does not require complex priors.
To address this, we use a neural pose prior, specifically a Normalizing Flow~\cite{rezende2015variational}.
% We use neural pose priors that can expressively model the distribution of natural poses. Specifically, we use Normalizing Flows~\cite{rezende2015variational}, which are likelihood-based models that can transform a simple distribution into a complex one using a series of invertible mappings. 

Our prior consists of a known base distribution and a sequence of bijective transformations that enable the model to map the data to the latent space and vice versa. 
The body (or hand) poses are modeled as a result of applying a transformation $T$ to a latent code $z$ sampled from a base distribution $p_z(z)$, which is a multivariate normal in our case. 
The transformation $T$ is invertible, %(hence $z=T^{-1}(q)$),
which allows us to use the change of variable formula to compute the density of the poses as 
\begin{equation}
    p_{q}(q) = p_z\big(T^{-1}(q)\big)\big|\det \mathbb{J}_{T^{-1}}(q)\big|
\end{equation}
where $\mathbb{J}_T$ is the Jacobian
% \TCnote{$J$ for Jacobian clashes with earlier definition for set of joints} 
of the transformation and $q$
% \todo[size=\tiny]{x is overloaded}
is the pose. 
More specifically $q=[\boldsymbol{\xi}, \mathbf{P}_T, \mathbf{P}_R]$,
% \TCnote{Earlier didn't we have $P_T$ and $P_R$ for global joint transforms?} 
the concatenation of local joint rotations and global joint transforms of a normalized pose (front facing root orientation positioned at $[0,0,0]$).

To capture the complexity of the body and hand pose distributions, we use a series of simple transformations $T=T_K \circ T_{K-1} \circ ... \circ T_1$ to form a complex one, where each $T_i$ maps $z_{i-1}$ to $z_i$, $z_0$ is the latent variable in the base distribution, and $z_K$ is the data $q$. 
This composition is modeled by a neural network (with Real-NVP~\cite{dinh2016density} architecture) and is trained to maximize the data log-likelihood, as 
\begin{equation}
    \log p_{q}(q) = \log p_z(z_0) - \sum_{i=1}^K \log \det \bigg|\frac{\partial T_i}{\partial z_i}\bigg|
    \label{eq:log_p}
\end{equation}

For model fitting, we define the pose prior energy term as $E_{\textrm{pose}} = - \alpha_b \log p_{q_b}(q_b) - \alpha_h \log p_{q_h}(q_h)$, wherein $q_b$ and $q_h$ are the body and hand pose representations, respectively, and $\alpha_b$ and $\alpha_h$ are the corresponding negative log-likelihood weights for body and hands.

\section{Experiments}
Given that holistic performance capture is not a common technique, there are few datasets available to evaluate the full method.
As such, we use a number of different benchmarks to evaluate sub-tasks in isolation.
Taken together, and in combination with our qualitative results, these demonstrate the high quality achieved by our method, and its robustness across diverse input data and scenarios.
Further results, details of the benchmark datasets and experimental procedure, and comprehensive ablation experiments are provided in the supplementary material.

It is important to note that while many benchmarks provide both training and test sets, we use the same DNNs (trained on the \SynthBody{}, \SynthHand{} and \SynthFace{} datasets only) for all experiments; we do not use \emph{any} of the benchmark training sets.
% Within-dataset experiments provide an upper-bound estimation as large DNNs can specialize to the domain of the specific dataset.
It is very difficult to achieve comparable metrics when evaluating cross-dataset~\cite{choudhury2023tempo}, though cross-dataset evaluation provides a much better estimate of generalization of the approach.
For benchmarks with training sets, the best results using the training set are denoted in \hlcol{green!25}{{\textbf{bold}}} and the best results not using the training set are shown \hlcol{yellow!25}{{\underline{underlined}}}.
% NoW, EHF and SSP-3D are fair comparisons for our method, while other benchmarks introduce a significant quantitative disadvantage; qualitative assessment as shown in \autoref{fig:qual_body_fits} may be more meaningful.
The ability of our method to generalize well across \emph{all} of these benchmarks suggests high diversity and fairness in our synthetic data, and that we are not biased to any particular scenario.

% \TBnote{we may want to say that cross-dataset provides a much better estimate of model generalization, and within-dataset experiments provide an overly optimistic or upper bound estimation}
% It is very hard to achieve comparable metrics when evaluating cross-dataset~\cite{choudhury2023tempo}; it is easy for large DNNs to specialize to the domain of a certain dataset, particularly when many of these datasets have quite limited visual diversity due to the fixed capture scenarios used.
%\TBnote{maybe reminder that we use the same DNNs and same fitter in each of the experiments, so our method does not need to be retrained for each scenario compared to other work.}
% \TBnote{I feel the above paragraph is really important to get right, and at the moment it feels like it could be polished and tightened up.}

% \todo[size=\tiny]{@Julien - are these benchmarks/metrics good for performance capture anyway - accuracy vs plausibility.}
% \todo[size=\tiny]{Qual comparison to faceware and moveai?}

\begin{table}
\footnotesize
\centering
  \caption{Single-view hand reconstruction errors on the FreiHAND dataset~\cite{zimmermann2019freihand}, \emph{ours is the only method not trained on FreiHAND}. Qualitative results are shown in \autoref{fig:metro_comp}.}
  \resizebox{\linewidth}{!}{%
    \begin{tabular}{lccc}
    \toprule
    \multirow{2}{*}{Method} & \multirow{2}{*}{PA-MPVPE $\downarrow$} & \multirow{2}{*}{PA-MPJPE $\downarrow$} & F-Scores $\uparrow$ \\
    & & & 5 mm / 10 mm \\
    \midrule
    \emph{Non-parametric}\\
    I2L-MeshNet~\cite{moon2020i2l} & 7.6  & 7.4  & 0.681 / 0.973 \\
    METRO~\cite{lin2021end} & \sota{6.7} & \sota{6.8} & \sota{0.717} / \sota{0.981} \\
    % \midrule
    \emph{Parametric}\\
    FrankMocap~\cite{rong2020frankmocap} & 11.6 & 9.2  & 0.553 / 0.951 \\
    PIXIE~\cite{feng2021collaborative} & 12.1 & 12.0 & 0.468 / 0.919 \\
    Hand4Whole~\cite{moon2022Hand4Whole} & 7.7  & 7.7 & 0.664 / 0.971 \\
    PyMAF-X~\cite{zhang2023pymaf} & 8.1 & 8.4 & 0.638 / 0.969 \\
    PyMAF-X \tiny{(extra data)} \footnotesize \cite{zhang2023pymaf} & 7.5 & 7.7 & 0.671 / 0.974 \\
    \midrule
    Ours & \dsota{8.1} & \dsota{8.6} & \dsota{0.653} / \dsota{0.967} \\
    \bottomrule
    \end{tabular}%
  \label{tab:hand_eval}%
}%
\end{table}%
 
\begin{figure}
    \centering
    \footnotesize
    \begin{tabularx}{\linewidth}{@{}YYYYYY@{}}
         \multirow{2}{*}{Input} & METRO & \multirow{2}{*}{Ours} & \multirow{2}{*}{Input} & METRO & \multirow{2}{*}{Ours} \\
         \multicolumn{3}{c}{\tiny \cite{lin2021end}} & \multicolumn{3}{c}{\tiny \cite{lin2021end}}\\
    \end{tabularx}
    \includegraphics[width=\linewidth]{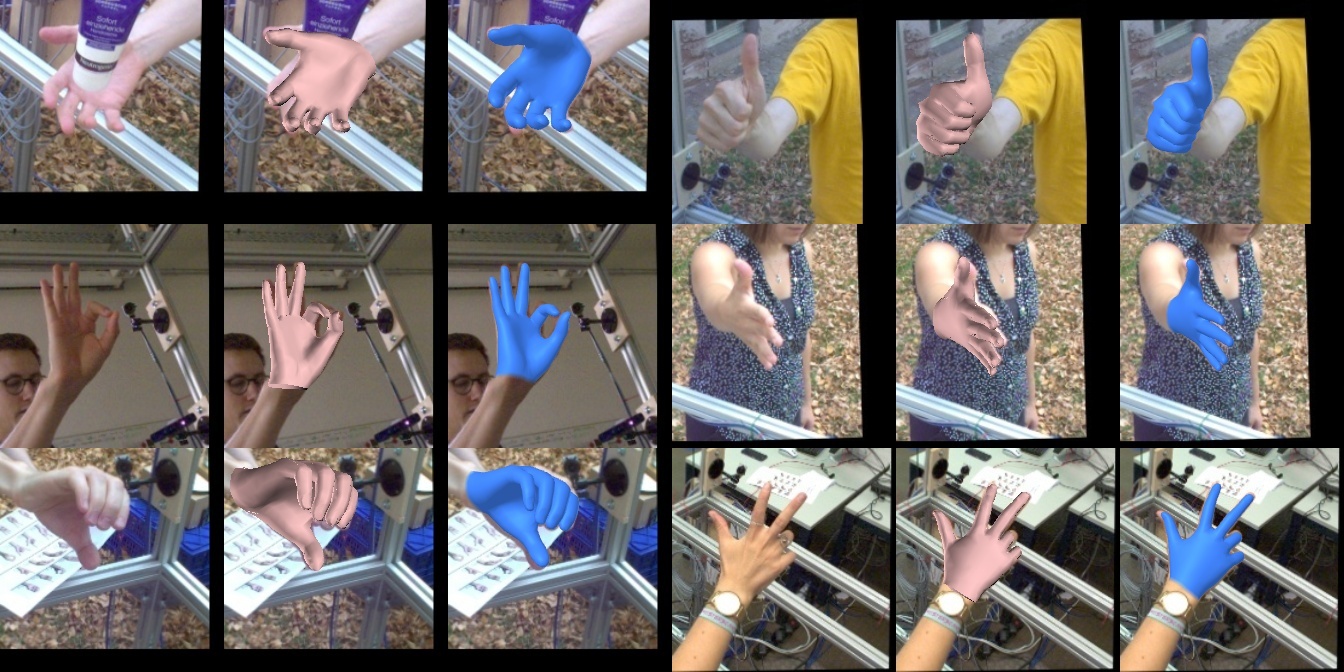}
    \caption{Qualitative comparison to METRO~\cite{lin2021end} for hand reconstruction, on images from the FreiHAND dataset~\cite{zimmermann2019freihand}. \copylabel{FreiHAND}{LMB, University of Freiburg}.} 
    \label{fig:metro_comp}
\end{figure}

\paragraph{Hands}
To evaluate performance capture for the hands we use the FrieHAND dataset~\cite{zimmermann2019freihand}.
We fit just the MANO hand model~\cite{romero2017embodied} using the prediction of the hand DNN and $E_{\textrm{pose}}$ only.
% \BLnote{Again, this is a large enough gap that I feel like you do really need to do the work to justify the proposed method over just asserting it "performs comparably". Can you finetune on this dataset and see how things change?}
Quantitative results are shown in \autoref{tab:hand_eval}, where our method performs comparably to recent approaches.
Note that ours is the only method that is \emph{not} trained on the FreiHAND training set and that recent non-parametric methods~\cite{moon2020i2l,lin2021mesh} lead to superior metrics but are not useful for performance capture.
Qualitative results and comparison to recent work are shown in \autoref{fig:metro_comp} and the supplementary material.

\begin{figure*}
    \centering
    \begin{minipage}{.33\linewidth}
    \footnotesize
    \begin{tabularx}{\linewidth}{@{}YYY@{}}
         \multirow{2}{*}{Input}  & MICA & \multirow{2}{*}{Ours} \\
         & \tiny\cite{zielonka2022towards} & \\
    \end{tabularx}
    \includegraphics[width=\linewidth]{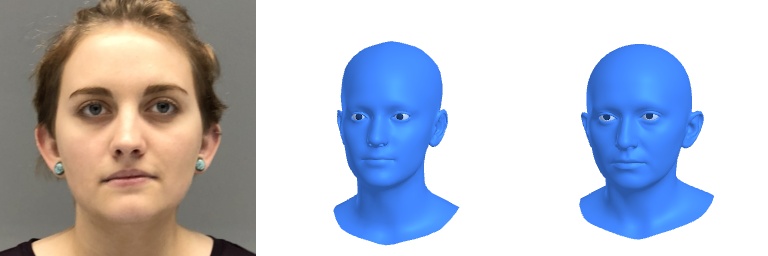}\\
    \includegraphics[width=\linewidth]{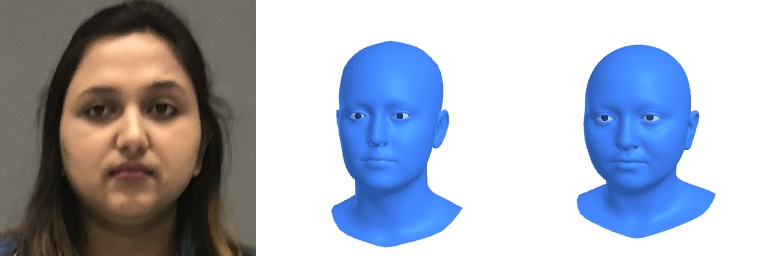}
    \end{minipage}
    \hfill
    \begin{minipage}{.33\linewidth}
    \footnotesize
    \begin{tabularx}{\linewidth}{@{}YYY@{}}
         \multirow{2}{*}{Input}  & MICA & \multirow{2}{*}{Ours} \\
         & \tiny\cite{zielonka2022towards} & \\
    \end{tabularx}
    \includegraphics[width=\linewidth]{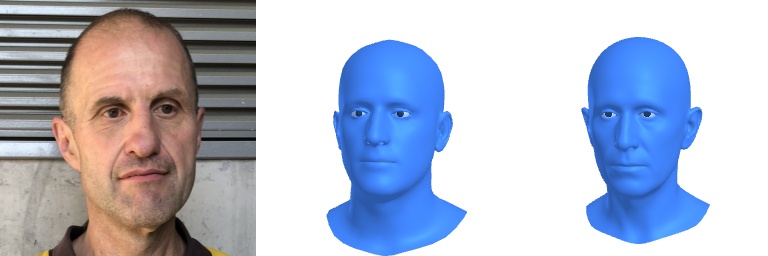}\\
    \includegraphics[width=\linewidth]{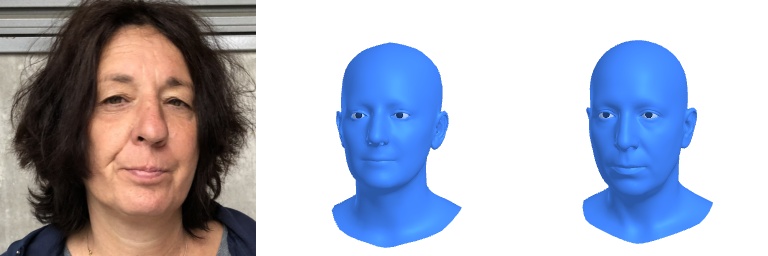}
    \end{minipage}
    \hfill
    \begin{minipage}{.33\linewidth}
    \footnotesize
    \begin{tabularx}{\linewidth}{@{}YYY@{}}
         \multirow{2}{*}{Input}  & MICA & \multirow{2}{*}{Ours} \\
         & \tiny\cite{zielonka2022towards} & \\
    \end{tabularx}
    \includegraphics[width=\linewidth]{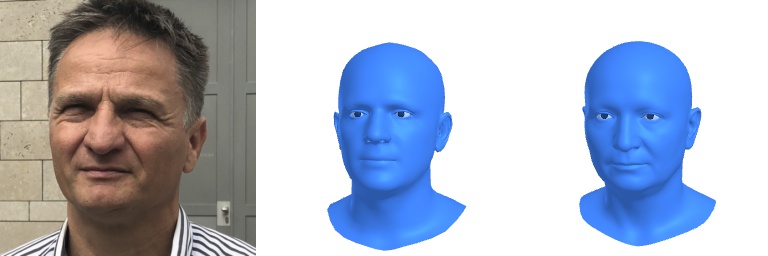}\\
    \includegraphics[width=\linewidth]{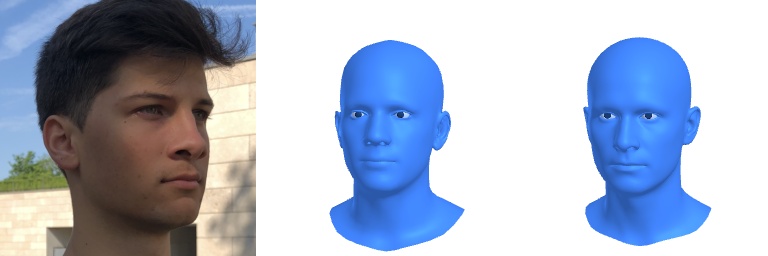}
    \end{minipage}
    \caption{Qualitative comparison to MICA~\cite{zielonka2022towards} for face reconstruction on a number of images from the NoW challenge~\cite{Sanyal2019_ringnet}. We observe greater variation in nose, lip, ear and head shape, along with more accurate representation of wrinkles, in our method. \copylabel{NoW}{MPI}.} 
    \label{fig:now_comp}
\end{figure*}

\begin{table}
    \centering
    \footnotesize
    \caption{Face shape reconstruction results on the NoW Challenge~\cite{Sanyal2019_ringnet} validation and test sets. Qualitative results are shown in \autoref{fig:now_comp}.} 
    \begin{tabularx}{\linewidth}{Xcccccc}
        \toprule
        Method & \multicolumn{3}{c}{Val Error (mm)} & \multicolumn{3}{c}{Test Error (mm)}\\
        \midrule
        Single-view & Median & Mean & Std & Median & Mean & Std\\
        \midrule
        PyMAF-X~\tiny{\cite{zhang2023pymaf}} & - & - & - & 1.13 & 1.42 & 1.20 \\
        \citet{wood20223d} & 0.95 & 1.17 & 0.97 & 1.02 & 1.28 & 1.08\\
        AlbedoGAN~\tiny{\cite{rai2023towards}} & 0.90 & 1.12 &  0.96 & 0.98 & 1.21 &  0.99 \\
        MICA~\tiny{\cite{zielonka2022towards}} & 0.91 & 1.13 & 0.95 & 0.90 & 1.11 & 0.92 \\
        TokenFace~\tiny{\cite{zhang2023accurate}} & - & - & - & \textbf{0.76} & \textbf{0.95} & \textbf{0.82} \\
        Ours & \textbf{0.87} & \textbf{1.08} & \textbf{0.89} & 0.89 & 1.11 & 0.93 \\
        \midrule
        Multi-view & Median & Mean & Std & Median & Mean & Std\\
        \midrule
        \citet{Bai_2021_CVPR} & - & - & - & 1.08 & 1.35 & 1.15\\ 
        \citet{wood20223d} & 0.75 & 0.90 & 0.73 & 0.81 & 1.01 & 0.84 \\
        Ours & \textbf{0.68} & \textbf{0.83} & \textbf{0.68} & \textbf{0.70} & \textbf{0.88} & \textbf{0.74} \\
        \bottomrule
    \end{tabularx}
    \label{tab:now_results}
\end{table}
\paragraph{Face}
We evaluate facial performance capture on the NoW benchmark \cite{Sanyal2019_ringnet} by fitting just the face of the \SOMA{} model, omitting any unnecessary energy terms; results are shown in \autoref{tab:now_results}.
Most full-body benchmarks have low-quality face ground truth, so this lets us compare our face model with other recent methods, including the model of \citet{wood2021fake}, more directly.
% asses the impact of the improvements \TBnote{do we want to call these improvements or just evaluating our face model vs. one from Wood et al? Maybe we can instead rephrase to be more similar to hand section, we evaluate face performance in isolation, this highlights the benefits of our parametric model vs Wood et al and our approach vs other recent face reconstruction methods.} to the face model described in \autoref{sec:face_id}.
% We use the model fitting method of \citet{wood20223d} to fit just the face of the \SOMA{} model, with our face DNN to regress landmarks.\TCnote{Any reason why we can't say that we use our model fitter? Seems to confuse the message to say that we use Erroll's.}
We also evaluate a variant of the NoW benchmark where we reconstruct the face shape of each subject using \emph{all} images of that subject together as input to our system, i.e., the multi-view scenario.
% \autoref{tab:now_results} shows that we significantly outperform \citet{wood20223d}, demonstrating the value of the improvements made to the face shape basis while the method is unchanged.
While our method does not outperform \citet{zhang2023accurate} in the single-view case, our approach extends to multi-view input which provides more accurate reconstructions.
% \BLnote{I feel like you have to address the gap between the proposed method and TokenFace. It's large enough that you can't really claim "on par," so it immediately raises the question: Why didn't you just use their approach for the face?}
Qualitative results and comparison to recent work are shown in \autoref{fig:now_comp} and the supplementary material.
The NoW challenge only considers the face, not the full head, so improved quality of head/ear shape is not captured in the quantitative results.

\paragraph{Body}
\hyphenation{Pav-la-kos}
For the full body we evaluate on three datasets: 
EHF \cite{pavlakos2019expressive} which evaluates both shape and pose reconstruction,
SSP-3D \cite{sengupta2020straps} which evaluates shape reconstruction, and
Human3.6M \cite{ionescu2014human3} which evaluates skeletal pose estimation.
EHF is most directly applicable to the performance capture use case, %s we envisage for our method,
though only covers the single-view scenario.
Human3.6M does not allow for assessment of shape reconstruction, but enables us to test our multi-view effectiveness.

\begin{figure*}
    \centering
    \footnotesize
    \begin{tabularx}{\linewidth}{@{}YYYYYYY@{}}
         \multirow{2}{*}{Input} & PIXIE &  HybrIK-X & OSX & PyMAF-X & \multirow{2}{*}{Ours} & \multirow{2}{*}{GT} \\
         & \cite{feng2021collaborative} & \cite{li2023hybrik} & \cite{lin2023one} & \cite{zhang2023pymaf} & \\
    \end{tabularx}
    \includegraphics[width=\linewidth]{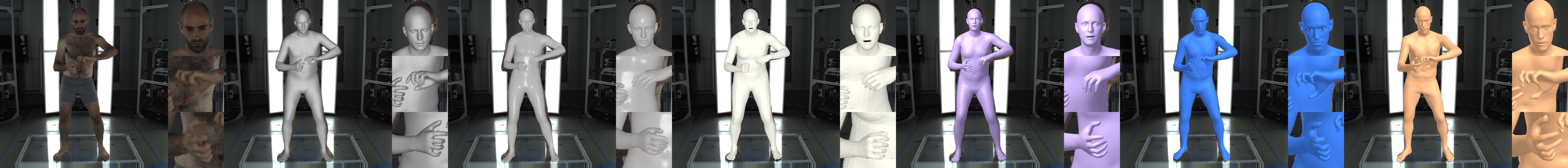}
    \includegraphics[width=\linewidth]{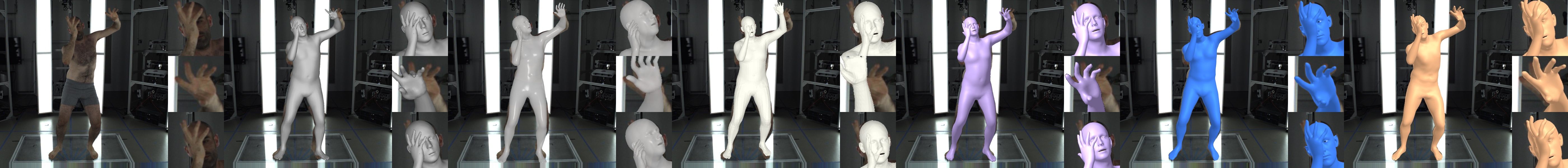}
    \includegraphics[width=\linewidth]{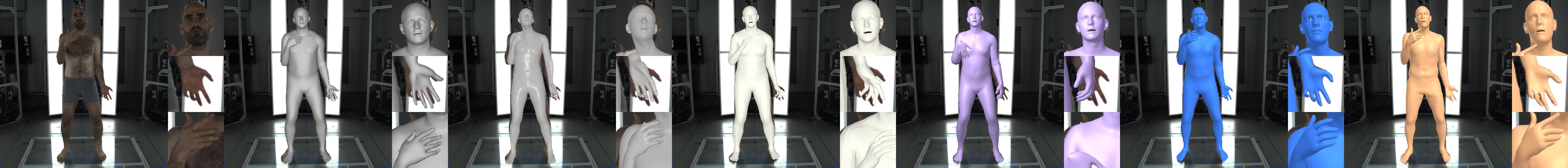}
    \includegraphics[width=\linewidth]{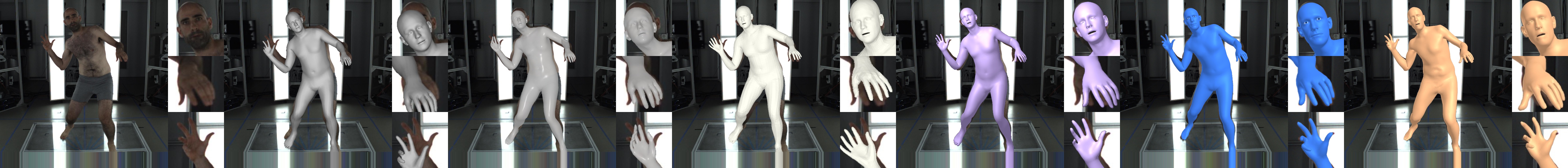}
    \includegraphics[width=\linewidth]{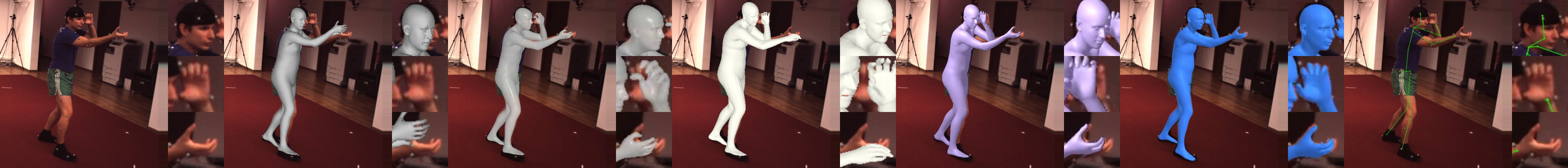}
    \includegraphics[width=\linewidth]{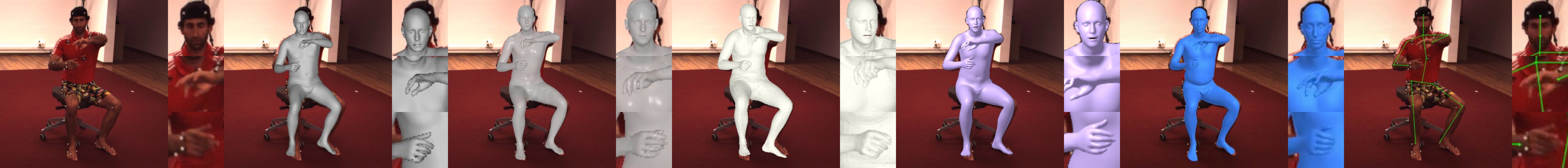}
    \includegraphics[width=\linewidth]{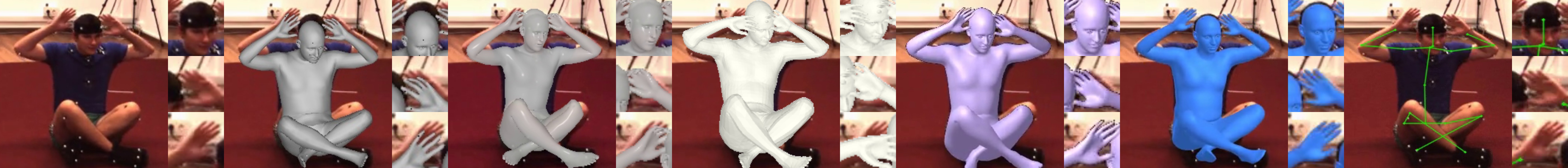}
    \caption{Qualitative examples of single-view full body reconstruction for images from the EHF~\cite{pavlakos2019expressive} (top 4 rows) and Human3.6M~\cite{ionescu2014human3} (bottom 3 rows) datasets compared with recent methods. 
    Additional qualitative results including failure cases can be found in the supplementary material.
    Our method achieves more precise results and significantly higher fidelity reconstruction of the face. \copylabel{EHF}{MPI}, \copylabel{Human3.6M}{IMAR}.}
    \label{fig:qual_body_fits}
\end{figure*}

Results for the EHF dataset~\cite{pavlakos2019expressive} are shown in \autoref{tab:ehf_eval}.
We outperform recent methods in most metrics or are on par with concurrent works, and qualitatively achieve significantly more accurate results as shown in \autoref{fig:qual_body_fits}.
We can easily introduce additional supervisory signals to our method; for example if we introduce a single measurement of the actor's height we achieve a full-body MPVPE (\emph{without} pelvis alignment) of 43.1mm, as the scale ambiguity inherent to the single-view scenario is removed.

\begin{table*}
    \centering
    \footnotesize
    \caption{Single-view holistic reconstruction errors on the EHF dataset~\cite{pavlakos2019expressive}. MPVPE is pelvis aligned, PA-MPVPE is aligned with Procrustes analysis. Qualitative results are shown in \autoref{fig:qual_body_fits}.\label{tab:ehf_eval}}
    \begin{tabular}{lccccccccc}
\toprule
\multicolumn{1}{c}{\multirow{2}{*}{Method}} & \multicolumn{3}{c}{MPVPE} & \multicolumn{4}{c}{PA-MPVPE} & \multicolumn{2}{c}{PA-MPJPE} \\
\cmidrule(lr){2-4} \cmidrule(lr){5-8} \cmidrule(lr){9-10}
 & Full-body & Hands & Face & Full-body & Body & Hands & Face & Body & Hands \\
\midrule
% MTC~\cite{xiang2019monocular} & -     & -     & -     & 67.2  & -     & -     & -     & 107.8 & 16.7 \\
% SMPLify-X~\cite{pavlakos2019expressive} & -     & -     & -     & 65.3  & 75.4  & 12.3  & 6.3   & 87.6  & 12.9 \\
% ExPose~\cite{choutas2020monocular} & 77.1  & 51.6  & 35.0  & 54.5  & 52.6  & 12.8  & 5.8   & 62.8  & 13.1 \\
% FrankMocap~\cite{rong2020frankmocap} & 107.6 & 42.8  & -     & 57.5  & 52.7  & 12.6  & -     & 62.3  & 12.9 \\
% Zhou et al~\cite{zhou2021monocular} & 90.8  & 51.7  & 28.1  & 66.2  & 60.3  & 14.6  & 7.0   & 70.9  & 14.3 \\
% PIXIE~\cite{feng2021collaborative} & 89.2  & 42.8  & 32.7  & 55.0  & 53.0  & 11.1  & 4.6 & 61.5  & 11.6 \\ from their paper - wrong for face!
PIXIE~\cite{feng2021collaborative}* & 89.1  & 42.8  & 32.6  & 55.0  & 53.8  & 11.2  & 5.5 & 65.5  & 11.1 \\
% Hand4Whole~\cite{moon2022Hand4Whole} & 76.8  & 39.8  & 26.1  & 50.3  & -     & 10.8  & 5.8   & 60.4  & 10.8 \\
% HMR2~\cite{goel2023humans}* & 85.5 & 58.5 & 42.4 & 51.2 & 49.8 & 16.3 & 7.2 & 59.8 & 16.2 \\
HybrIK-X~\cite{li2023hybrik}* & 121.4 & 52.1 & 41.9 & 59.8 & 50.0 & 17.6 & 8.1 & 60.8 & 17.9 \\
OSX~\cite{lin2023one} & 70.8 & 53.7 & 26.4 & 48.7 & - & 15.9 & 6.0 & - & - \\
PyMAF-X (Res50)~\cite{zhang2023pymaf} & 68.0 & 29.8 & 20.5 & 47.3  & 45.9 & 10.1  & 5.6 & 52.9 & 10.3 \\     
PyMAF-X (HR48)~\cite{zhang2023pymaf} & 64.9 & 29.7 & 19.7 & 50.2  & 44.8 & 10.2  & 5.5 & 52.8 & 10.3 \\
SMPLer-X-L20~\cite{cai2024smpler}$^\dagger$ & 65.4 & 49.4 & 17.4 & 37.8  & - &  15.0 & 5.1 & - & - \\
SMPLer-X-L32~\cite{cai2024smpler}$^\dagger$ & 62.4 & 47.1 & 17.0 & 37.1 & - & 14.1 & 5.0 & &  \\
Multi-HMR-ViT-L/14~\cite{multi-hmr2024}$^\dagger$ & \textbf{42.0}& 28.9 & \textbf{18.0} & \textbf{28.2} & - & 10.8 & 5.3 & - & - \\
\midrule
Ours & 51.0 & \textbf{21.6} & 18.4 & 34.2 & \textbf{33.0} & \textbf{9.8} & \textbf{4.9} & \textbf{38.0} & \textbf{9.7} \\
\bottomrule
\end{tabular}\\[3pt]
    \parbox{\linewidth}{\centering \tiny * Results evaluated for this paper using official implementation - \citet{feng2021collaborative} use a different evaluation protocol for the face, \citet{li2023hybrik} do not report on EHF. $^\dagger$ Results from concurrent works.}
\end{table*}

Results for shape reconstruction on the SSP-3D dataset are shown in \autoref{tab:ssp_results}.
Our method outperforms the closest method also trained exclusively on synthetic data~\cite{black2023bedlam}, as well as beating the overall state-of-the-art approach~\cite{sengupta2021hierarchical}.

\begin{table}
    \centering
    \footnotesize
    \caption{Single-view body shape reconstruction errors on the SSP-3D dataset~\cite{sengupta2020straps}.\label{tab:ssp_results}}
    \begin{tabular}{lc}
  \toprule
  Method & PVE-T-SC \\
  \midrule
  HMR~\cite{Kanazawa2018_hmr} & 22.9 \\ 
  SPIN~\cite{Kolotouros2019_spin} & 22.2 \\
  SHAPY~\cite{Shapy:2022} & 19.2 \\
  STRAPS~\cite{sengupta2020straps} & 15.9 \\
  \citet{sengupta2021probabilistic} & 15.2 \\
  \citet{sengupta2021hierarchical} & 13.6\\
  BEDLAM-CLIFF~\cite{black2023bedlam} & 14.2 \\
  \midrule
  Ours & \textbf{11.8} \\
  \bottomrule
\end{tabular}

\end{table}

Results for multi-view reconstruction on the Human3.6M dataset \cite{ionescu2014human3} are shown in \autoref{tab:h36m_results_multi}.
Our method significantly outperforms other multi-view methods which are also not trained on Human3.6M.
Given the limited variety of the data there is a high chance that other methods over-fit; this is supported by the results of \citet{choudhury2023tempo} reproduced in \autoref{tab:h36m_results_multi}.
The Human3.6M benchmark evaluates only joint positions which do not fully specify human pose, leaving some degrees of freedom (e.g., rotation about a limb) undefined~\cite{zhang2021we}.
% Our method recovers the full body mesh, and so fully specifies the pose. 
Our method, which recovers the full body mesh, often leads to more compelling results, supported by the qualitative results shown in \autoref{fig:qual_body_fits}.

\begin{table}
  \centering
  \footnotesize
  \caption{
Multi-view body pose errors on the Human3.6M validation set~\cite{ionescu2014human3} excluding sequences with incorrect ground-truth annotations following~\citet{iskakov2019learnable}. MPJPE is \emph{not} pelvis aligned.
}
\begin{tabular}{lcc}
    \toprule
    Multi-view Method & Trained on H36M & MPJPE \\
    \midrule
    % \citet{martinez20173dbaseline} & \checkmark & 57.0\\
    \citet{pavlakos2017harvesting} & $\times$ & 56.9 \\
    \citet{iskakov2019learnable} & \checkmark & \sota{17.7} \\
    % VoxelPose~\cite{voxelpose} & \checkmark & 19.0 \\ 
    % \citet{kadkhodamohammadi2021generalizable} & \checkmark & 49.1\\
    % MvP~\cite{wang2021mvp} & \checkmark & 18.6\\
    % Faster VoxelPose~\cite{fastervoxelpose} & \checkmark & 19.8\\
    % \citet{ma2022ppt} & \checkmark & 24.4\\
    TEMPO~\cite{choudhury2023tempo} & \checkmark & 18.5\\
    TEMPO~\cite{choudhury2023tempo} & $\times$ & 63.0 \\
    \midrule
    Ours & $\times$ & \dsota{27.9} \\
    \bottomrule
  \end{tabular}
  \label{tab:h36m_results_multi}
\end{table}

\section{Discussion}

\subsection{Limitations}
Our method has a number of limitations, both in the synthetic training data and model fitting approach.
Due to the probabilistic nature of the landmarks we can ignore any frames where the average confidence of the landmarks is too low, resulting in smoothly interpolated results from the adjacent frames.
This produces plausible results, but at significant cost to accuracy in those frames.
The system as a whole is reliant on the quality of the predicted 2D landmarks which can lead to failures for interlocking hands or if poor sub-ROIs are selected.
Tuning of priors to regularize the optimization can also be cumbersome, though in most cases these are transferable.
In the future we can extend to an ML-based iterative optimization approach to help reduce the reliance on bespoke priors and increase robustness to errors~\cite{choutas2022learning}.
% \BLnote{Again, qualitative examples here would be cool. Or even an example where you do poorly, add some assets/poses into the synthetics library, then do really well.}
We often find that failures can be traced back to the synthetic training data. %, for example very unusual poses or clothing not present in our data can be problematic.
Particular shortcomings of our synthetic data include absence of particularly unusual poses, loose-fitting clothing and hair simulation.
Using our method, gaps in pose data can be quite easily addressed through capture and manual cleanup.
Gaps in the asset library can be addressed through artist effort.
% While neither of these is ideal, we note that through the data multiplication effect of the synthetic data it can be much \emph{less} costly to conduct these seemingly labour intensive additions than it would be to collect equivalent real data. \TBnote{this feels a bit wordy and can be shrunk, or even deleted (although I like the sentiment)}
% For the face model we still see some issues in accuracy of eyelid shape, this can be addressed by reprocessing the data to ensure consistency in the eye region.
% While our landmarks are very dense compared to other work, they are still a relatively sparse signal.
% It is likely that this sparsity lessens the impact of the remaining domain gap between synthetic data and real data.
% Exploring regression of denser signals such as surface normals is an interesting area of future work which may put more pressure on any domain gap, though there have been some promising initial results~\cite{raman2023mesh}.
% Our detrection and tracking methods are simple and can result in poor performance in cases of multiple people interacting in the scene. 
% Execution time could be significantly reduced using a single-shot detection method such as YOLO~\cite{redmon2016you}, and tracking performance could be improved with a simple method such as a Kalmann filter, or more modern methods like PHALP~\cite{rajasegaran2022tracking}.
% \TBnote{If looking for space I would be tempted to delete above? although reviewers might ask about speed, we do not report details anywhere?}
While we can support moving cameras when provided with calibration, the current implementation does not support jointly optimizing for subject \emph{and} camera positions that vary through time.
This would be a powerful extension of the system to allow truly arbitrary camera input and has been shown to be effective in recent work~\cite{shin2024wham}. %; our system could quite easily be extended given better priors of global human and camera motion.

\subsection{Conclusion}
We have demonstrated a system which enables holistic, high-quality performance capture of humans, including the body, face and hands.
Our parametric model improves on models of the body and hands \cite{romero2017embodied} and face \cite{wood2021fake} to represent the complete human at high fidelity.
We use this model to generate synthetic data to train DNNs for 2D landmark regression and direct pose and shape prediction from images.
We then use optimization to fit our model to images or videos of people, providing high-quality holistic performance capture.
Our approach focuses on 
\emph{(1) Robustness} enabled by DNNs directly regressing pose and shape;
\emph{(2) Accuracy} enabled by DNNs predicting image-space features;
\emph{(3) Adaptability} of model fitting to support arbitrary numbers and positions of cameras.
Because of this, our system can operate with no markers or proprietary hardware, arbitrary cameras, without calibration, and produces combined results for the body, hands and face from a single shot.
We show state-of-the-art results on benchmarks for human reconstruction, demonstrating that it is possible to achieve high-quality, holistic performance capture without the hassle.

\begin{acks}
The authors would like to thank Rodney Brunet, Kendall Robertson and Jon Hanzelka for their work on the clothing asset library; Steve Hoogendyk for his work on the tongue blendshapes; and Ben Lundell and Erroll Wood for their comments and suggestions.
\end{acks}

\newpage
\appendix

\section{Model Definition}
The \SOMA{} mesh is made up of $N = 12943$ vertices and 12726 polygons with a skeleton of $K = 54$ joints: 22 for the body (the SMPL skeleton), 15 per hand (as in \SMPLH{}) and 2 for the eyes, from the face model of \citet{wood2021fake}, represented as axis-angles.

The mesh vertex positions are defined by mesh generating function
$\mathcal{M}(\boldsymbol{\beta}, \boldsymbol{\gamma}, \boldsymbol{\zeta}, \mathbf{\psi}, \boldsymbol{\theta})\!:\!\mathbb{R}^{|\boldsymbol{\beta}|+|\boldsymbol{\gamma}|+|\boldsymbol{\zeta}|+|\boldsymbol{\psi}|+|\boldsymbol{\theta}|}\!\to\!\mathbb{R}^{N\times3}$ which takes parameters
$\boldsymbol{\beta}\in\mathbb{R}^{300}$ for body shape,
$\boldsymbol{\gamma}\in\mathbb{R}^{256}$ for face shape,
$\boldsymbol{\zeta}\in\mathbb{R}^{9}$ for hand shape,
$\boldsymbol{\psi}\in\mathbb{R}^{224}$ for expression, and
$\boldsymbol{\theta}\in\mathbb{R}^{54\times3}$ for skeletal pose.
\begin{equation}
    \mathcal{M}(\boldsymbol{\beta}, \boldsymbol{\gamma}, \boldsymbol{\zeta}, \boldsymbol{\psi}, \boldsymbol{\theta}) =
    \mathcal{L}(\mathcal{T}(\boldsymbol{\beta}, \boldsymbol{\gamma}, \boldsymbol{\zeta}, \boldsymbol{\psi}, \boldsymbol{\theta}), \boldsymbol{\theta}, \mathcal{J}(\boldsymbol{\beta}, \boldsymbol{\gamma}, \boldsymbol{\zeta}); \mathbf{W})
\end{equation}
where $\mathcal{L}(\mathbf{X}, \boldsymbol{\theta}, \mathbf{J}; \mathbf{W})$ is a standard linear blend skinning (LBS) function that rotates
vertex positions $\mathbf{X}\in\mathbb{R}^{N\times3}$ about joint locations $\mathbf{J}\in\mathbb{R}^{K\times3}$ by local
joint rotations $\mathbf{\boldsymbol{\theta}}$, with per-vertex hand-authored skinning weights $\mathbf{W}\in\mathbb{R}^{K\times N}$ determining how rotations are interpolated across the mesh.

$\mathcal{T}(\boldsymbol{\beta}, \boldsymbol{\gamma}, \boldsymbol{\zeta}, \boldsymbol{\psi}, \boldsymbol{\theta})\!:\!\mathbb{R}^{|\boldsymbol{\beta}| + |\boldsymbol{\gamma}| + |\boldsymbol{\zeta}| + |\boldsymbol{\psi}| + |\boldsymbol{\theta}|}\to\mathbb{R}^{N\times3}$
constructs an unposed body mesh by adding displacements to the template mesh $\mathbf{\overline{T}}\!\in\!\mathbb{R}^{N \times 3}$, which represents the average body in T-pose with neutral expression:
\begin{equation}
    \mathcal{T}(\boldsymbol{\beta}, \boldsymbol{\gamma}, \boldsymbol{\zeta}, \boldsymbol{\psi}, \boldsymbol{\theta})^j_{\:k} =
    \overline{T}^j_{\:k} +
    \beta_i S^{ij}_{\;\;k} +
    \gamma_i U^{ij}_{\;\;k} +
    \zeta_i V^{ij}_{\;\;k} +
    \psi_i E^{ij}_{\;\;k} +
    P(\mathbf{\theta})^j_{\:k}
\end{equation}
With linear 
body shape basis, $\mathbf{S}\!\in\!\mathbb{R}^{|\boldsymbol{\beta}| \times N \times 3}$,
face shape basis, $\mathbf{U}\!\in\!\mathbb{R}^{|\boldsymbol{\gamma}| \times N \times 3}$,
hand shape basis, $\mathbf{V}\!\in\!\mathbb{R}^{|\boldsymbol{\zeta}| \times N \times 3}$,
expression basis, $\mathbf{E}\!\in\!\mathbb{R}^{|\boldsymbol{\psi}| \times N \times 3}$, and
$P(\boldsymbol{\theta})$ which represents pose-dependent blendshape offsets for pose parameters $\boldsymbol{\theta}$ .
% $\mathbf{\overline{T}}$ is constructed by hand based on the \SMPLH{} neutral template and template mesh of \citet{wood2021fake}.
Finally, $\mathcal{J}(\boldsymbol{\beta}, \boldsymbol{\gamma}, \boldsymbol{\zeta})\!:\!\mathbb{R}^{|\boldsymbol{\beta}| + |\boldsymbol{\gamma}|+ |\boldsymbol{\zeta}|}\to\mathbb{R}^{K\times3}$ moves the joint locations to account for changes in shape:
\begin{equation}
    \mathcal{J}(\boldsymbol{\beta}, \boldsymbol{\gamma}, \boldsymbol{\zeta})^j_{\:k} =
    J(\overline{T}^j_{\:k} + \beta_i S^{ij}_{\;\;k} + \gamma_i U^{ij}_{\;\;k} + \zeta_i V^{ij}_{\;\;k})
\end{equation}
Where $J$ is a modified version of the \SMPLH{} joint regressor.
The body shape basis, $\mathbf{S}$, is that from neutral \SMPLH{}~\cite{romero2017embodied} masked to only the body.
The hand shape basis, $\mathbf{V}$, is that from MANO~\cite{romero2017embodied} excluding the first component which broadly represents scale, as this is already captured in $\mathbf{S}$.
Both are PCA bases learned from scans of humans.
The pose-dependent blendshapes, $P(\boldsymbol{\theta})$, are also taken from neutral \SMPLH{}.
The face shape basis, $\mathbf{U}$, is learned from a library of face scans as described in the main paper.
The face expression basis, $\mathbf{E}$, is taken from \citet{wood2021fake}, but with an additional 12 blendshapes added to provide articulation of the tongue.

\begin{figure}
    \includegraphics[width=0.5\linewidth]{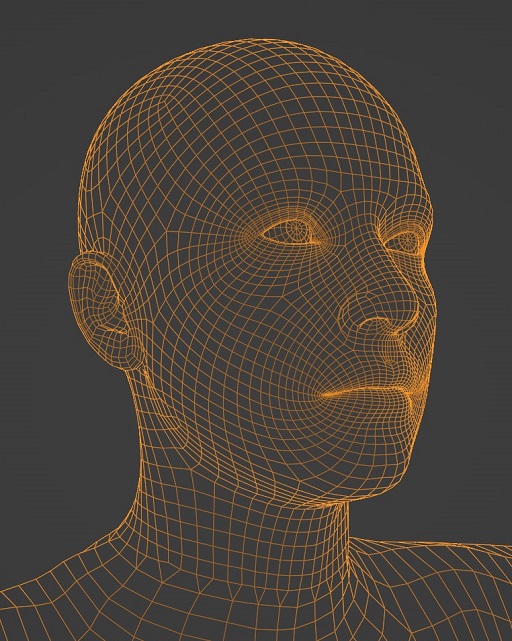}
    \caption{Head and neck topology of the \SOMA{} model.}
    \label{fig:neck_topo}
\end{figure}

To construct the template mesh, $\mathbf{\overline{T}}$, we manually align the template of \citet{wood2021fake} to the head of the \SMPLH{} template.
Once aligned, the head of \SMPLH{} and lower neck of the new head are removed and the two partial meshes merged.
The topology around the join is hand-crafted to create a smooth transition given the different density of the two meshes, see \autoref{fig:neck_topo}.
A new UV-map is also hand-authored based on the \SMPLH{} UV space.

Given that the topology of the template mesh differs significantly from both \citet{wood2021fake} and \SMPLH{}, the bases associated with these models must be adapted to work with the \SOMA{} model.
To do this, we calculate a mapping function from the respective source topologies to the \SOMA{} topology.
Specifically, we map data from the source model to the \SOMA{} by iterating through each \SOMA{} vertex and taking the sum of the values for each vertex in the closest triangle of the source mesh, weighted according to the barycentric coordinate of the closest surface point on the source mesh to the target \SOMA{} vertex.
We restrict the bases of \citet{wood2021fake} to affect on the head and neck, and those of \SMPLH{} to affect only the body, using a hand-crafted mask.

\section{Automatic Face Retopologization}
\label{app:autowrap}

To add training data to the model of \citet{wood20223d}, one would typically retopologize new head scans into the target topology using manual correspondence point selection in a tool such as Wrap \cite{Wrap3}.
This is slow and prone to human labeling inconsistencies. 
To reach our total of 1040 samples we automatically retopologized new face scans using the following method.

Firstly, due to poor consistency in the mouth region of the training data we annotate points on the lip margin of each scan.
We then render each scan with uniform shading from many viewpoints. 
We use our multi-view fitting method to fit a 3D mesh that best reconstructs the scan in the face model's topology. 
% This fitted mesh could be used as a training sample but has limited quality and landmark inaccuracies which do not add sufficient diversity in the face model in training.
We use this as an initialization to Wrap which can correct inaccuracies and improve quality of surface details compared to the scan.
Supervising the Wrap retopologization process requires sparse correspondence points between the raw scan and the target face model topology for accurate surface alignment. 
We automatically select a sparse set on the fitted mesh's ears and eyes and add these to our manually annotated lip correspondences.
Since the fitted mesh is close to the scan surface, we select corresponding points on the scan by finding the nearest scan vertices that have similar normals to those on the fitted mesh.
To improve accuracy of back-of-ear correspondence points, we displace these points on the fitted mesh by 2cm behind the ear before finding the nearest point with similar normals on the scan mesh.
Once the fitted mesh is retopologized onto the scan mesh using these correspondences, we use photometric refinement \cite{Nicolet2021Large} to better capture high-frequency features such as wrinkles in the same topology.

\begin{figure}
    \centering
    \includegraphics[width=\linewidth]{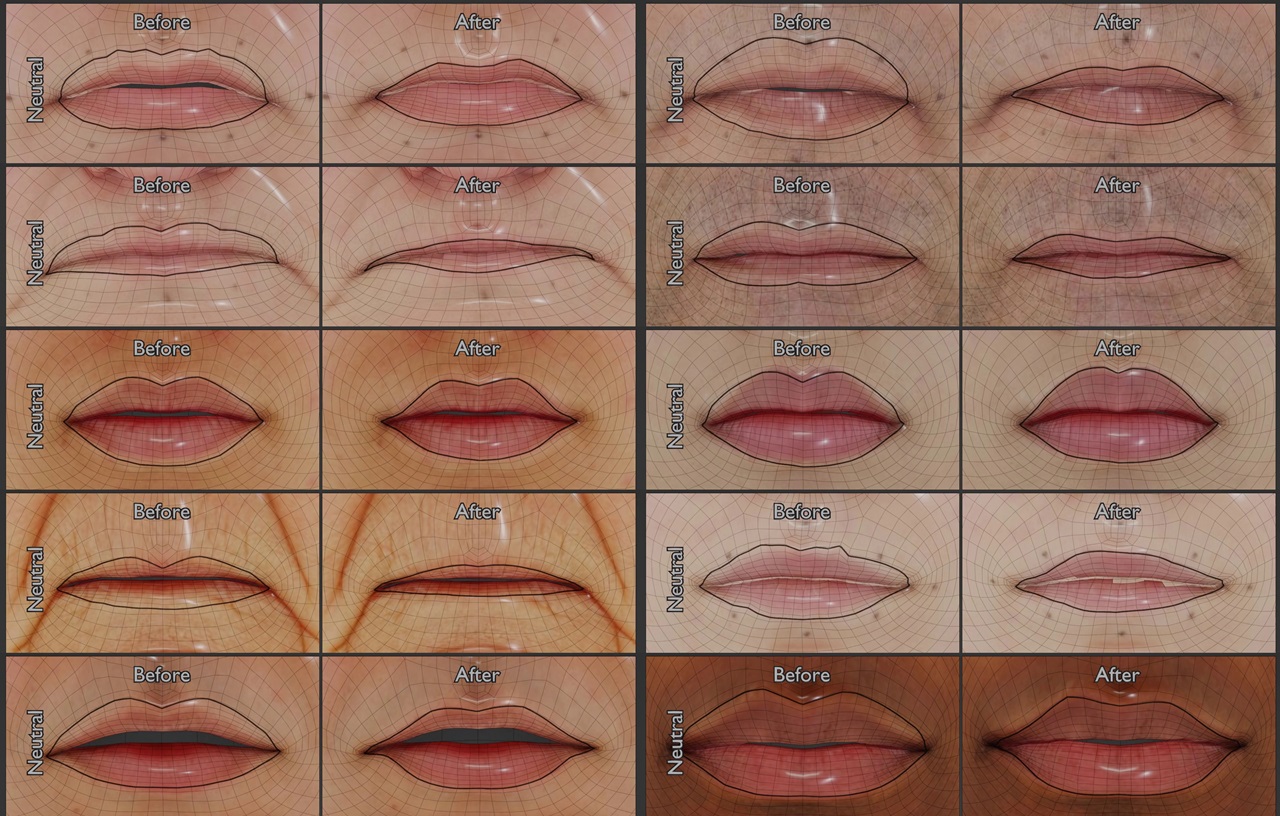}
    \caption{Improvements to semantic consistency in the mouth region as part of our updated face identity basis training. The same edge loop is shown in bold for the original retopologized scans and our updated versions. The new training data has topology that is significantly more consistent with the semantic region of the lips shown by the texture.}
    \label{fig:lip_consistency}
\end{figure}

Examples showing the poor semantic consistency in the mouth region described in the main paper, and improvements resulting from the above process, are shown in \autoref{fig:lip_consistency}.

\section{Synthetic Data Details}

\subsection{Pose Library}

In order to pose the body realistically we collect a large library of poses and expressions from a variety of sources.
For the facial expression and eye motion, we start with the library of \citet{wood2021fake}, which is captured using the technique of \citet{wood20223d}.
Using the updated model and techniques outlined here we reprocess the raw multi-view video data to extract higher quality results including tongue motion.
We also add a number of artist-animated expression sequences making use of the tongue articulation in order to boost representation in the generated data.
In total this gives us almost 165,000 frames of expression and eye gaze data to sample from.
For body pose data we combine the AMASS library~\cite{mahmood2019amass} with a diverse library of additional pose sequences captured using conventional optical motion capture technologies and processed using MoSh~\cite{loper2014mosh}.
Most of these libraries include only the body pose, but some also include articulated hands.
In total this comprises over 2 million frames.
We also use the MANO dataset~\cite{romero2017embodied} to provide additional hand poses in isolation.

We sample randomly from these libraries, combining face expression and body pose from independent sequences.
If the body pose sequence lacks articulated hands then we splice these in from the hand pose library.
For body pose, we weight the sampling by the log probability of an individual pose under a GMM fit to the entire library.
For facial expression, we weight the sampling by the sum of the absolute value of the expression coefficients, and additionally randomize gaze direction and eye closing.
This ensures that we pick difficult and diverse poses and expressions from the library. %, which has an over-representation of ``easy'' poses, e.g., T-pose, standing, walking.
% Examples are shown in \autoref{fig:pose_samples}.

\subsection{Clothing}
To model clothing in a way that adapts to a wide variety of body shapes and poses we use displacement maps~\cite{ma2020learning}.
The clothing assets are artist authored as textured meshes using Marvelous Designer~\cite{mervelous} on our template body mesh and then baked to displacement maps.
We also author albedo, normal, metallic and roughness for each asset, providing a high level of realism.
At render time we use adaptive subdivision to retain high frequency details of the clothing.
Clothing assets are split into tops and bottoms, as well as a number of accessories such as bracelets, rings, gloves and watches.
We also model prosthetic limbs using displacement.
In total we have 384 displacement clothing assets and 148 displacement accessories.

We also adapt and extend some of the mesh clothing assets of \citet{wood2021fake} to fit our new template mesh and use the same lattice-based deformation to fit these to different identities at render time.
Our mesh clothing library includes 36 glasses assets, 56 headwear items and 7 masks and eyepatches.

While displacement clothing adapts very well to diverse body shapes and poses, it has significant limitations in terms of representation.
It is not possible to represent loose-fitting clothing such as dresses and skirts using this technique, leading to a significant gap in our synthetic data.
We hope to address this in the future using cloth simulation similar to that used by \citet{black2023bedlam}.

\subsection{Hair}
We adapt the hair library of \citet{wood2021fake} and extend it with a particular aim towards diversity of hair types and styles.
The hair grooms are artist-authored strand-based assets which we adapt to fit our new template mesh.
At generation time the base assets are morphed to fit the head shape of the specific identity.
In total our hair library includes 479 head hair styles, 81 eyebrows, 71 eyelashes and 91 facial hair styles.

Given the complexity of these assets, we currently do not perform any simulation to produce more realistic interaction of the hair with gravity or the body.
We mostly observe that this isn't problematic in terms of our results as the DNNs are able to learn to generalize sufficiently, however it is undeniably a component of the domain gap between our synthetic and real images.

\subsection{Textures}
We extend the face texture library of \citet{wood2021fake} to include 207 textures, enhancing the diversity of the selection in terms of ethnicity and age.
To obtain pore-level detail bumpmaps we also high-pass filter the albedo texture, as in \citet{wood2021fake}. However, this creates inconsistent appearance of skin specular roughness for different skin tones due to albedo textures differing in contrast. We weight the bumpmap strength by the skin tone brightness resulting in more natural and consistent appearance of skin specular highlights across skin tones.
For the body we use a library of 25 high quality textures balanced by gender.
When sampling textures we first select the face texture and then select a body texture with in some bound of perceptual similarity to the face texture.
Following this we recolor the body texture to precisely match the face texture.

\subsection{Environment}

We use two techniques to provide lighting and background for our renders: 1) HDR-image-based lighting for indoor and outdoor environments \cite{debevec2006image,wood2021fake}, 2) and mesh-based scenes for indoor environments.
For HDR environments we randomly sample from 914 images~\cite{polyhaven} and randomly rotate them and increase/decrease brightness to increase environment diversity.
For mesh-based scenes we use 47 indoor environments~\cite{chocofur,evermotion}.
The indoor environments are illuminated by area and point lights that simulate indoor lighting and an additional outdoor HDRI environment chosen from 477 outdoor HDRIs.
To find a valid placement of a subject in the scene we randomly sample both the location (on predefined ground planes) and the pose of a clothed mesh until we find a combination without any intersections between the subject and the scene.
As the mesh scene can cause occlusions, we randomly sample the camera until we find a position where at least $75\%$ of joints are visible.
We include secondary people to serve as occluders/distractors in a proportion of scenes.

\section{Detection and Tracking}

For detection and tracking of the body in images we use a sliding-window approach similar to that described by \citet{wood20223d}.
We train a light-weight DNN to detect 36 landmarks across the full body, with a ResNet18~\cite{he2016deep} backbone.

For the first frame in a video, or for a single image, we use the sliding-window approach, executing this DNN on each window and taking the window with the highest confidence over a threshold.
We then refine this candidate ROI by executing the DNN additional times to get a more precise bounding box around the predicted landmarks.
This refined ROI then serves as input to the dense landmark DNN.

In a video we use the bounding box of the dense landmarks from the previous frame as the candidate ROI instead of running the sliding-window method every frame.
If the median confidence for the predicted landmarks in the new frame is too low then we re-run the detection process in full for that frame.
We can optionally carry out the iterative refinement of the candidate ROI for each frame in the video.
To extract the face and hand ROIs in a given frame we take bounding boxes around subsets of the predicted dense full-body landmarks for the frame.

% Both the detection and tracking approaches outlined above are rudimentary and leave significant room for improvement.
% Execution time could be significantly reduced using a single-shot detection method such as YOLO~\cite{redmon2016you}, and tracking performance could be improved with a simple method such as a Kalmann filter, or more modern methods like PHALP~\cite{rajasegaran2022tracking} \todo{others?}.

\section{DNN Training Details}

For training the DNNs we use single frames of our synthetic datasets, which are cropped closely to the relevant body part, i.e., body, hand, and face. 
To this end, we extract a squared region of interest (ROI) around the specific set of landmarks. 
The crop is aligned with the image axes and tightly contains the 2D landmarks. 
We extract the squared ROI of size $128\times128$ for the hand model and of size $256\times256$ for body and face models. 
For evaluating the \SOMA{} model in $\mathcal{L}_{\textrm{joint}_T}$ and $\mathcal{L}_{\textrm{joint}_R}$ for partial predictions we use ground-truth values for shape, translation and any missing pose parameters.

\paragraph{Augmentation.}
During training, we enhance the ROI extraction process by applying rotation, scaling, and translation. 
This augmentation simulates scenarios where the ROI is not perfectly defined during inference.
In addition to ROI augmentations, we incorporate appearance augmentations into the training pipeline. 
In particular, we randomly blur the image to simulate motion blur effects caused by camera motion. 
This is achieved using directional blur kernels. 
It is important to note that this simulated motion blur is not due to actual scene movement; rather, it is a post-processing effect. 
This operation is proportional to image size and the direction of blur is randomly chosen. 
We also adjust brightness by adding a constant offset to all channels within a specified range. 
Similarly, we randomly increase or decrease contrast via $\text{img} = (\text{img} - 0.5) \times (1 + \text{contrast}) + 0.5$. 
% To mimic sensor noise, we independently add random Gaussian noise to each pixel in the image. 
Additionally, we alter the hue and saturation of the image, apply JPEG compression, and convert the image from BGR to grayscale. 
The application of each of these augmentations depends on a given probability.
Furthermore, for training our hand model, we introduce random occlusions by overlaying occluders from a library of images onto the original image. 
These occluders are not physically present in the scene; they are simply layered over the image. 
The occluders undergo random shifts, scaling, and rotations. 
Lastly, we observe that incorporating random ISO noise~\cite{alvar2022practical} (inspired by real camera noise) into the input image benefits our body model training. 
This noise is approximated as a combination of image intensity-dependent Poissonian noise and image intensity-independent Gaussian noise.

\paragraph{Network architecture.}
Our backbones are constructed from models in the PyTorch Image Models library~\cite{rw2019timm}. 
In particular we use \textit{hrnet-w48} backbone pre-trained on ImageNet1k for our face and body DNNs and \textit{hrnet-w18} backbone pre-trained on ImageNet1k for our hand DNN. We define the hidden dimension of size 512 to get the visual features from these backbones. 
For the landmark, pose, and shape heads, we use a simple feed-forward neural network. 
The network is composed of two fully-connected layers separated by a Leaky ReLU activation function. 
We use a hidden dimension of size 512 and the output dimension is defined separately based on the number of outputs for each task.

\paragraph{Training hyper-parameters.}
We train the body and hand DNNs for 600 epochs and the face DNN for 300 epochs with a batch size of (64$\times$4), distributed over four A100 GPUs. 
All networks are trained using the AdamW~\cite{loshchilov2017decoupled} optimizer with a learning rate of $10^{-4}$. 
The learning rate is adjusted using a Cosine Annealing scheduler for improved convergence. 
For training body and hand models, we set the landmark loss weight ($\alpha_l$) to 10.0 and the pose and shape weights to 1.0 where applicable. 
We set $\alpha_{t}$ to 2.0 and $\alpha_{r}$ to 1.0 for both body and hand DNNs.
For the neural pose prior, we train the body and hand priors for 600 epochs with a batch size of 1024. For both models, we use the AdamW optimizer, and we use a learning rate of $10^{-4}$ and $10^{-5}$ for the body and hand priors, respectively. Similar to body and hand DNNs, the learning rate is adjusted using a Cosine Annealing scheduler for improved convergence. 

\section{Registration Details}

\subsection{Energy Terms}
A number of energy terms not fully defined in the main paper are provided here.
The shape prior is made of two parts, one for the face and one for the body:
$$
  E_{\textrm{shape}} = E_{\textrm{face\_shape}} + E_{\textrm{body\_shape}}
$$
For each of face and body shape we have a multivariate Gaussian Mixture Model (GMM) of $G$ components:
$$
  E_{\textrm{part\_shape}} = -\log\left(p(\boldsymbol{\beta})\right)
$$
where
$p(\boldsymbol{\beta}) = \sum_{i=1}^{G} \alpha_i\; \mathcal{N}\! \left(\boldsymbol{\beta} | \boldsymbol{\nu}_i,\, \boldsymbol{\Sigma}_i\right)$.
$\boldsymbol{\nu}_i$ and $\boldsymbol{\Sigma}_i$ are the mean and covariance matrix of the $i^{th}$ component, and $\alpha_i$ is the weight of that component.
$E_{\textrm{exp}}$ prevents excessive use of expression blendshapes
$$
E_{\textrm{exp}} = \| \boldsymbol{\psi} \| + E_{\textrm{quartic}}
$$
where
$$
  E_{\textrm{quartic}} = \left\{
  \begin{array}{ll}
    (t_0 - t)^4 & t < 0\\
    0          & 0 \le t \le 1\\
    (t - t_1)^4 & t > 1
  \end{array} \right.
$$
$E_{\textrm{temp}}$ reduces jitter by encouraging face mesh vertices to remain still between neighboring frames $i-1$ and $i$ in a sequence of length $F$ with $C$ cameras.
$$
  E_{\textrm{temp}} = \sum_{i=2, j, k}^{F, C, |\mathbf{L}|} \| x_{i,j,k} - x_{i-1,j,k} \| ^2
$$
$E_{\textrm{cam}}$ penalizes movement of the cameras from their initial position:
$$
  E_{\textrm{cam}} = \| \mathbf{Q} - \mathbf{Q}_{\textrm{init}} \|^2
$$
where $\mathbf{Q}$ is the current position of all cameras and $\mathbf{Q}_{\textrm{init}}$ is the initial position of all cameras.
$E_{\textrm{intsct}}$ prevents intersection of the eyes and teeth with the face mesh, following \citet{wood20223d}. 
$$
E_{\textrm{intsct}} = E_{\textrm{eyes}} + E_{\textrm{teeth}}
$$
$E_{\textrm{eyes}}$ penalizes the squared distance between the vertices in the eyelid and the surface of the sphere, $E_{\textrm{teeth}}$ instead uses a convex hull of the teeth geometry and penalizes the distance inside this.
$$
  E_{\textrm{teeth}} = \sum_{i \in I} D_i^2
$$
where $D_i$ measures the distance the $i^{th}$ skin vertex is inside the convex hull,
$$
  D_{i} = \min_{h \in H} \left\{ d_{i,1}, \ldots, d_{i,H} \right\}
$$
and $d_{i,h}$ measures the internal distance between the $i^{th}$ skin vertex and the $h^{th}$ plane of the convex hull, defined by position $p_h$ and normal $\hat{n}_h$.
$$
  d_{i,j} = -\textrm{min} \left( \hat{n}_h \cdot x_i + p_h, 0 \right)
$$

\subsection{Pose Prior}

Our method utilizes a flow-based model, with Real-NVP~\cite{dinh2016density} architecture, as the pose prior. We deliberately choose Normalizing Flows~\cite{kolotouros2021probabilistic,aliakbarian2022flag} over other commonly used priors based on Variational Autoencoders (VAEs)~\cite{kingma2013auto} for two reasons: first, VAEs tend to require careful tuning of the trade-off between the KL divergence loss and the reconstruction loss, which may vary depending on the scenario~\cite{pavlakos2019expressive,rempe2021humor}. 
Second, unlike VAEs, which rely on approximate inference of the posterior distribution, Normalizing Flows can perform exact likelihood evaluations, enabling more robust optimization when used as a prior~\cite{aliakbarian2022flag}. This capability of Normalizing Flows (compared to VAEs) has been extensively evaluated in the literature~\cite{aliakbarian2022flag} wherein Normalizing Flows not only provided more accurate likelihood estimates, but also more robust estimates when evaluated on in- and out-of-distribution pose samples.

\section{Experimental Details}

\paragraph{Face}
The NoW benchmark contains 2054 2D images of 100 subjects, with ground-truth 3D head scan for each subject.
The validation set consists of 20 subjects and the test set consists of 80 subjects.
The objective is to reconstruct the 3D shape of the subjects' faces as accurately as possible from each single image.
We use the official evaluation process to evaluate our output meshes.

\paragraph{Hands}
The FreiHAND~\cite{zimmermann2019freihand} test set contains 3960 $224\times224$ pixel RGB images of the right hands of a number of subjects in various poses and environmental conditions, with ground-truth MANO parameters.
We use the official implementation to calculate the standard metrics for the 3D reconstruction benchmark on this dataset; PA-MPVPE, PA-MPJPE and F-scores at 5 and 10mm error.

\paragraph{Body}
The EHF dataset~\cite{pavlakos2019expressive} contains 100 monocular images of a single subject in a studio setting, with ground truth SMPL-X~\cite{pavlakos2019expressive} meshes.
We register our \SOMA{} model to the data and then retopologize to enable us to evaluate against the SMPL-X ground-truth.
This retopologization will itself introduce error as some geometry is not common between the two meshes, so the results we report here are a conservative approximation of the true quality.
We report MPVPE and PA-MPVPE for the entire body and various body parts in isolation, as well as PA-MPJPE for body and hands independently.

The SSP-3D dataset~\cite{sengupta2020straps} contains 311 single-view images of sportspeople in various poses with SMPL~\cite{loper2015smpl} ground-truth annotations.
Again we register the \SOMA{} model and retopologize to match the ground-truth, in this case there is negligible error in this process.
We report the PVE-T-SC metric, vertex error for the scale-corrected T-pose mesh, which is standard for this dataset.

The Human3.6M dataset~\cite{ionescu2014human3} contains captures of a number of subjects from four cameras with ground-truth skeletal 3D landmarks.
We use every fifth frame for subjects 9 and 11 for our evaluation, both in the single-view and multi-view case.
For multi-view evaluation, we omit some sequences from subject 9 which have invalid ground truth~\cite{iskakov2019learnable}.
For the single-view case, MPJPE with pelvis alignment and PA-MPJPE are reported, while for multi-view we report MPJPE \emph{without} pelvis alignment.
We evaluate the standard 14 joint configuration for the dataset, but a mapping is required from the \SOMA{} mesh to these 14 joints as the joints of the \SOMA{} models are not consistent with the Human3.6M annotation scheme.
We follow the approach of \citet{tan2016fits} and solve for the affine combination of just four vertices of the mesh that best predict each joint on an equally-spaced 5\% subset of frames. 
Note that the flexibility of this mapping is extremely limited and this procedure simply automates an otherwise manual and biased process.

\paragraph{Performance}
We run our experiments either locally on a machine with an NVIDIA RTX 2080 GPU or on a virtual machine with an NVIDIA A100 GPU.
On the RTX 2080 machine execution of the hand, face and body DNNs for a single image takes approximately 100ms
The model fitting process for the EHF dataset (100 single-view frames, 200 iterations) takes 10 minutes on an A100 VM. 
Model fitting is performed jointly on the whole sequence and a large amount of time is spent copying data to and from the GPU, so execution time does not scale linearly with the number of frames.
For example, fitting a 3000 frame sequence from Human3.6M (200 iterations) takes 30 minutes on an A100 VM.

\section{Additional Results}

\begin{figure}
    \centering
    \includegraphics[height=0.325\linewidth]{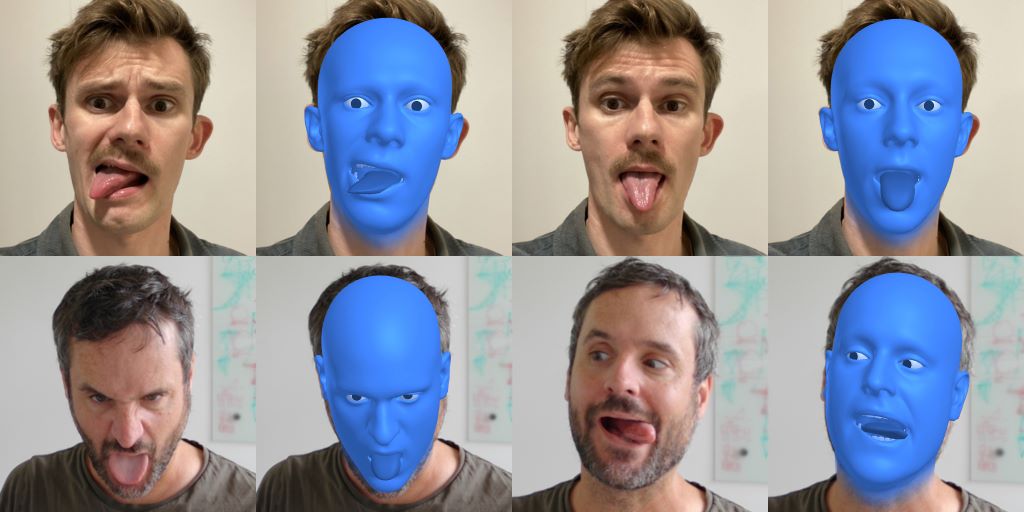}
    \hfill
    \includegraphics[height=0.325\linewidth]{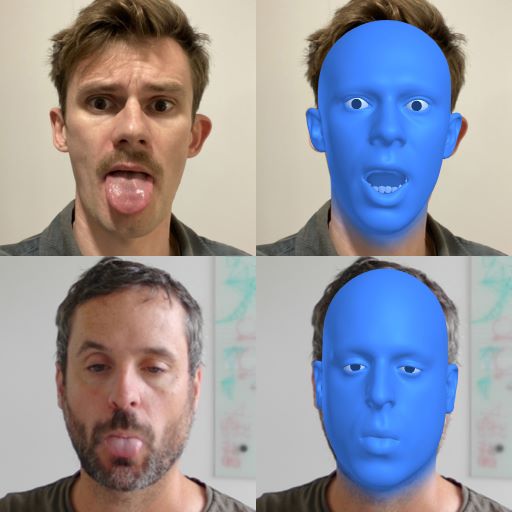}
    \caption{Results including visible tongue. In many cases we achieve accurate results (left), although we do see a number of failure cases (right). For example, where the landmark DNN misses the tongue or confuses it for thicker lips, or the tongue blendshapes can't capture extreme tongue shapes.}
    \label{fig:tongue_res}
\end{figure}

\autoref{fig:tongue_res} shows some results for faces including visible tongues. 
In many cases the reconstruction is of high quality, though we note failures due to lack of expressivity of the blendshapes and failures of the landmark DNN to accurately regress landmarks on the tongue.
Common failure cases include confusion of the tongue for thicker lips, and failure to identify the visibility of the tongue at all.
Addition of the tongue blendshapes also demonstrates how we could quite easily add further blendshapes based on quality gaps noticed in reconstruction quality for a given capture, or based on an enrolment sequence for a specific actor with idiosyncratic facial motion.

% \begin{figure}
%     \centering
%     \includegraphics[height=0.655\linewidth]{figures/tongue_good.jpg}
%     \hfill
%     \includegraphics[height=0.655\linewidth]{figures/tongue_bad.jpg}\\
% \vspace{-0.75em}
%     \caption{Registration results including visible tongues. In many cases we achieve an accurate reconstruction (left), although we do see a number of failure cases (right). For example, where the tongue blendshapes can't capture extreme tongue shapes, or the landmark DNN misses the tongue or confuses it for thicker lips.}
%     \label{fig:tongue_res}
% \end{figure}

\begin{figure*}[p!]
    \centering
    \footnotesize
    \begin{tabularx}{\linewidth}{@{}YYYYYYY@{}}
         \multirow{2}{*}{Input} & PIXIE &  HybrIK-X & OSX & PyMAF-X & \multirow{2}{*}{Ours} & \multirow{2}{*}{GT} \\
         & \cite{feng2021collaborative} & \cite{lin2023one} & \cite{zhang2023pymaf} & \cite{li2023hybrik} & \\
    \end{tabularx}
    \includegraphics[width=\linewidth]{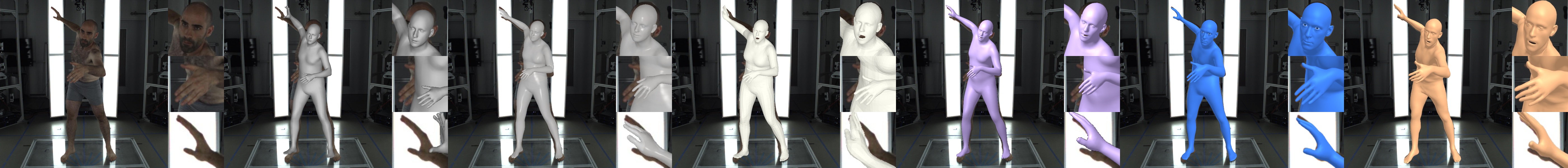}
    \includegraphics[width=\linewidth]{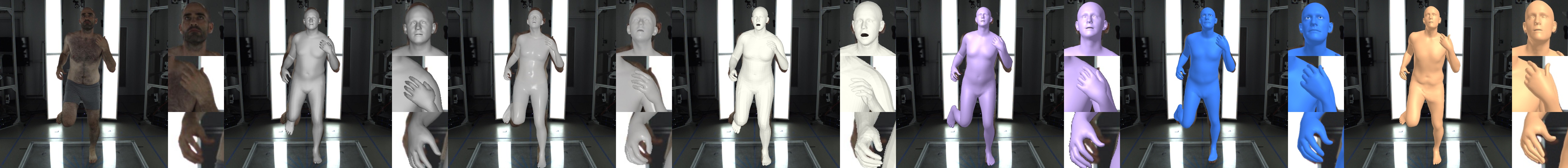}
    \includegraphics[width=\linewidth]{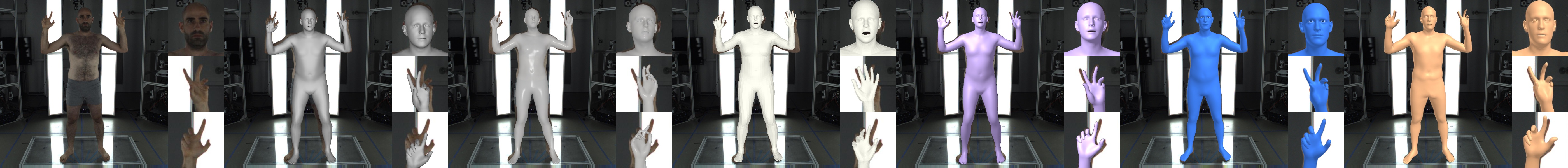}
    \includegraphics[width=\linewidth]{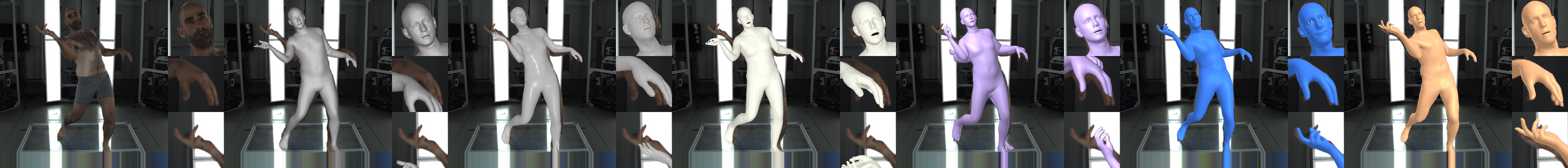}
    \includegraphics[width=\linewidth]{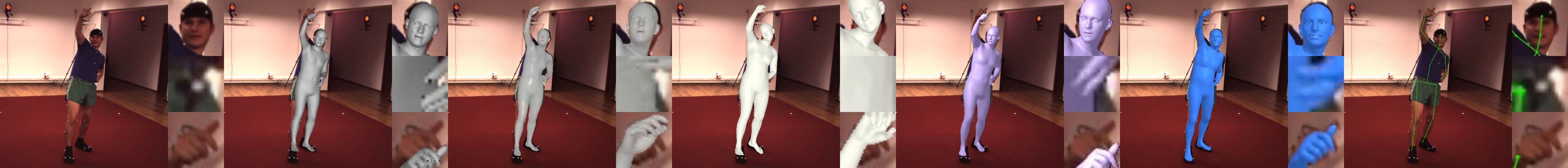}
    \includegraphics[width=\linewidth]{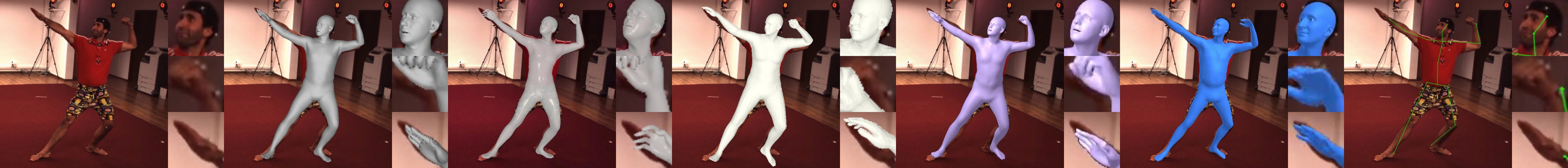}
    \includegraphics[width=\linewidth]{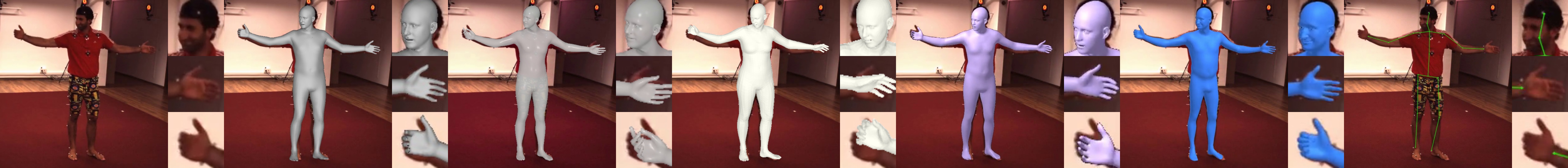}
    \includegraphics[width=\linewidth]{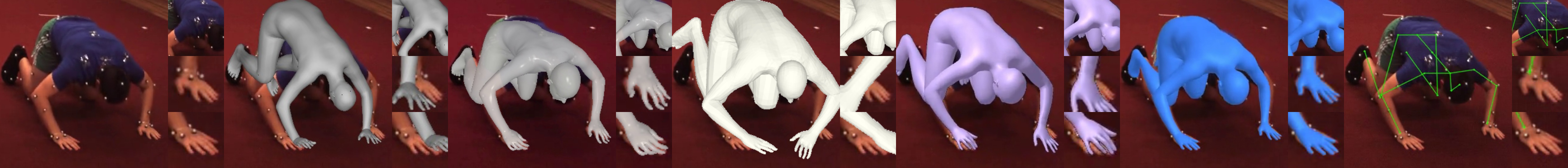}\\\vspace{0.5em}
    \includegraphics[width=\linewidth]{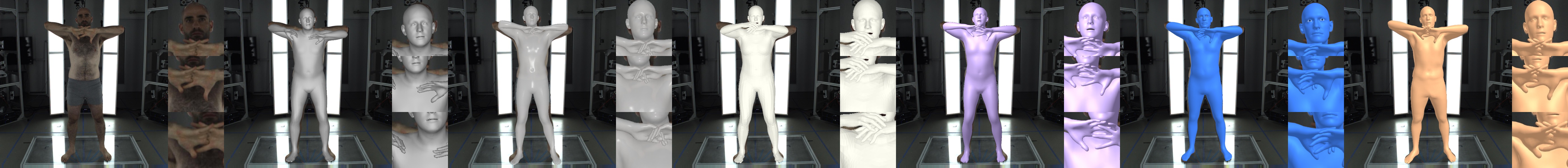}\\
    \includegraphics[width=\linewidth]{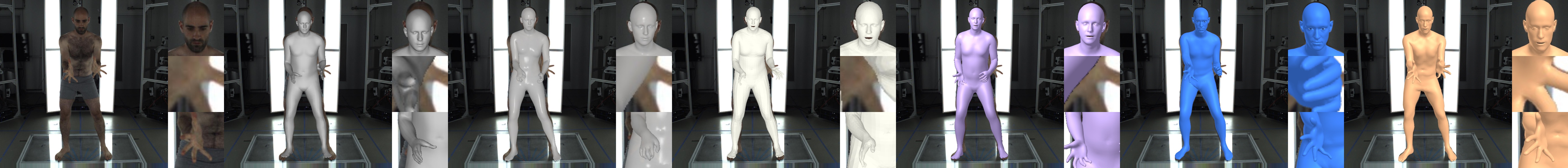}
    \caption{Further qualitative comparisons for full-body reconstruction on the EHF~\cite{pavlakos2019expressive} and Human3.6M~\cite{ionescu2014human3} datasets. Failure cases are shown in the bottom two rows; our method can perform poorly for interlocking hand poses, and in cases where the hand network fails, here due to a bad ROI. \copylabel{EHF}{MPI}, \copylabel{Human3.6M}{IMAR}} 
    \label{fig:extra_comp}
\end{figure*}

Results for the single-view case on the Human3.6M dataset~\cite{ionescu2014human3} are shown in \autoref{tab:h36m_results_mono}.
Our method performs comparatively to other methods despite ours being the only method \emph{not} trained on the Human3.6M training set.
As highlighted in the main paper, the qualitative results shown in the main paper and in \autoref{fig:extra_comp} demonstrate that our method recovers significantly higher-quality body meshes.
This however, is not reflected in the MPJPE metric calculated on sparse ground-truth annotations.

\begin{table}
  \centering
  \footnotesize
  \caption{Single-view body pose errors on the Human3.6M validation set~\cite{ionescu2014human3}. MPJPE is pelvis aligned, PA-MPJPE is aligned with Procrustes-analysis. \emph{Note that ours is the only method not trained on the Human3.6M training set.} Qualitative results are shown in the main paper and \autoref{fig:extra_comp}.}
  \begin{tabular}{lcc}
    \toprule
    Single-view Method & MPJPE & PA-MPJPE \\
    \midrule
    % \multirow{5}[2]{*}{\begin{sideways}Temporal\end{sideways}} & Kanazawa et al~\cite{kanazawa2019learning} & -     & 56.9 \\
    %       & Arnab et al~\cite{arnab2019exploiting} & 77.8  & 54.3 \\
    %       & DSD~\cite{sun2019human} & 59.1  & 42.4 \\
    %       & VIBE~\cite{kocabas2020vibe} & 65.9  & 41.5 \\
    %       & TCMR~\cite{choi2021beyond} & 62.3  & 41.1 \\
    % \midrule
     % \multirow{10}[10]{*}{\begin{sideways}Single-view\end{sideways}} & Pavlakos et al~\cite{pavlakos2018learning} & -     & 75.9 \\
          % & HMR~\cite{kanazawa2018end} & 88.0  & 56.8 \\
          % & NBF~\cite{omran2018neural} & -     & 59.9 \\
          % & LearnedGD~\cite{song2020human} & -     & 56.4 \\
          % & HUND~\cite{zanfir2021neural} & 69.5  & 52.6 \\
          % & GraphCMR~\cite{kolotouros2019convolutional} & -     & 50.1 \\
          % & Pose2Mesh~\cite{choi2020pose2mesh} & 64.9  & 46.3 \\
          % & HMR-EFT~\cite{joo2021exemplar} & -     & 46.0 \\
    SPIN~\cite{Kolotouros2019_spin} & 62.5  & 41.1 \\
    I2L-MeshNet~\cite{moon2020i2l} &  55.7  & 41.1 \\
    HybrIK~\cite{li2021hybrik} & 54.4  & 34.5 \\
    Graphormer~\cite{lin2021mesh} & 51.2  & 34.5 \\
    PyMAF-X (Res50)~\cite{zhang2023pymaf} & 58.8  & 40.2 \\
    PyMAF-X (HR48)~\cite{zhang2023pymaf} & 54.2  & 37.2 \\
    HMR2~\cite{goel2023humans} & \sota{45.3} & 33.4 \\
    HybrIK-X~\cite{li2023hybrik} & 47.0 & \sota{29.8} \\
    \midrule
    Ours & \dsota{56.2} & \dsota{40.1} \\
\bottomrule
\end{tabular}
\label{tab:h36m_results_mono}
\end{table}

Further qualitative comparisons for full-body reconstruction on the EHF~\cite{pavlakos2019expressive} and Human3.6M~\cite{ionescu2014human3} datasets are shown in \autoref{fig:extra_comp}.
One failure case of our method is for complex interlocking hand poses.
We suspect that with higher-quality pose data (e.g., \citet{Moon_2020_ECCV_InterHand2.6M}) and consequently better synthetic data (e.g., \citet{moon2023reinterhand}) we might improve robustness in these cases.
It is possible though that the model fitting method itself may need to be augmented to adequately deal with these scenarios, e.g., to account for intersection or contact points~\cite{taheri2020grab,Taheri_2022_CVPR}.

\begin{figure}
    \centering
    \includegraphics[width=\linewidth]{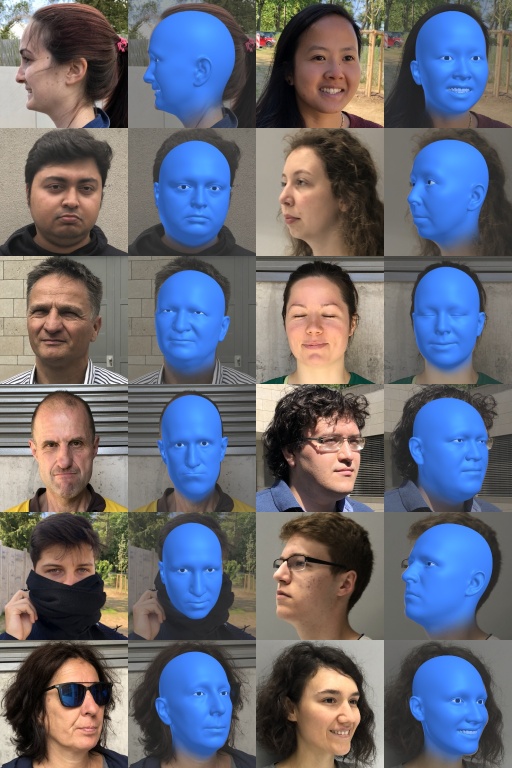}
    \caption{Qualitative results for single-view face reconstruction on images from the NoW benchmark~\cite{Sanyal2019_ringnet}. \copylabel{NoW}{MPI}.} 
    \label{fig:now_qual}
\end{figure}

\begin{figure}
    \centering
    \includegraphics[width=\linewidth]{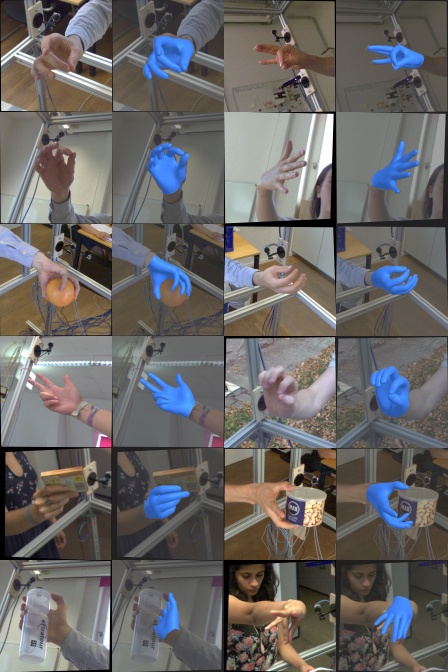}
    \caption{Qualitative results for hand reconstruction on images from the FreiHAND dataset~\cite{zimmermann2019freihand}. Our method fails for some cases of significant occlusion and challenging poses (bottom row). \copylabel{FreiHAND}{LMB, University of Freiburg}} 
    \label{fig:freihand_qual}
\end{figure}

Qualitative results of our method for face reconstruction on the NoW benchmark~\cite{Sanyal2019_ringnet} are shown in \autoref{fig:now_qual}.
Qualitative results of our method for hand reconstruction on the FreiHAND dataset~\cite{zimmermann2019freihand}, including some failure cases, are shown in \autoref{fig:freihand_qual}.

% Our method can struggle in cases of significant occlusion, or extreme poses.
% \autoref{fig:metro_comp} shows comparisons of our method to the state-of-the-art non-parametric approach METRO~\cite{lin2021end}.
% We observe that thumb articulation is less well captured in our method, possible due to lack of representation in our hand pose libray.
% Using the confidence values for the landmarks predicted by the DNN enables the filtering out of data in frames where uncertainty is too high.
% For these single view reconstruction benchmarks this is not a viable strategy, but for video data can help to provide the most plausible results with graceful degradation in challenging cases.

\section{Ablations}

In this section we present and discuss a number of ablation experiments to verify the different components of our approach.

\begin{table*}
\footnotesize
\centering
\caption{Ablations on the EHF dataset~\cite{pavlakos2019expressive}. MPVPE is pelvis aligned.}
    \begin{tabular}{lccccccccc}
    \toprule
    \multicolumn{1}{c}{\multirow{2}{*}{Method}} & \multicolumn{3}{c}{MPVPE} & \multicolumn{4}{c}{PA-MPVPE} & \multicolumn{2}{c}{PA-MPJPE} \\
    \cmidrule(lr){2-4} \cmidrule(lr){5-8} \cmidrule(lr){9-10}
     & Full-body & Hands & Face & Full-body & Body & Hands & Face & Body & Hands \\
    \midrule
    No DNN init. & 105.5 & 60.5 & 20.1 & 100.1 & 63.4 & 16.6 & 5.2 & 95.0 & 16.5 \\
    No optimization & 61.7 & 56.0 & \textbf{17.9} & 49.3 & 42.6 & 10.4 & 5.9 & 52.5 & 10.2 \\
    No prob. landmarks & 67.4 & 25.6 & 18.2 & 37.0 & 34.9 & 11.8 & 5.2 & 40.7 & 11.7 \\
    % No hand or face DNN & 55.7 & 48.0 & \textbf{16.7} & 42.3 & 36.4 & 15.0 & 6.1 & 44.4 & 14.8 \\
    % No pose prior & 52.0 & 22.0 & 18.8 & 34.7 & 33.0 & 10.5 & 4.9 & 38.2 & 10.4 \\
    \midrule
    Full & \textbf{51.0} & \textbf{21.6} & 18.4 & \textbf{34.2} & \textbf{33.0} & \textbf{9.8} & \textbf{4.9} & \textbf{38.0} & \textbf{9.7} \\
    \bottomrule
    \end{tabular}
\label{tab:ehf_ablations}
\end{table*}

\begin{table}
  \centering
  \footnotesize
  \caption{Single-view ablations on the Human3.6M validation set~\cite{ionescu2014human3}. MPJPE is pelvis aligned.}
  \begin{tabular}{lcc}
    \toprule
    Single-view Method & MPJPE & PA-MPJPE \\
    \midrule
    No DNN init. & 92.8 & 78.6 \\
    No optimization & 66.6 & 47.1 \\
    No prob. landmarks & 59.6 & 41.4 \\
    No pose prior & 56.5 & 40.6 \\
    No identity prior & 56.8 & 41.0\\
    No temporal prior & 57.2 & 41.0 \\
    \midrule
    Full & 56.2 & 40.1 \\
\bottomrule
\end{tabular}
\label{tab:h36m_ablations_mono}
\end{table}

\begin{table}
  \centering
  \footnotesize
  \caption{Multi-view ablations on the Human3.6M validation set~\cite{ionescu2014human3} excluding sequences with incorrect ground-truth annotations following~\citet{iskakov2019learnable}. MPJPE is \emph{not} pelvis aligned.}
\begin{tabular}{lc}
    \toprule
    Multi-view Method & MPJPE \\
    \midrule
    No DNN init. & 28.7 \\
    % No optimization & \\
    No prob. landmarks & 32.9 \\
    No pose prior & 27.9 \\
    No identity prior & 28.1 \\
    No temporal prior & 27.8 \\
    \midrule
    Full & 27.9 \\
    \bottomrule
  \end{tabular}
  \label{tab:h36m_ablations_multi}
\end{table}

\subsection{Initialization and Optimization}

Our method uses direct regression by DNNs to initialize an optimization process, while many prior works use \emph{either} DNNs to regress pose and/or shape, or optimization without dynamic initialization to reconstruct the human.

In \autoref{tab:ehf_ablations}, \autoref{tab:h36m_ablations_mono} and \autoref{tab:h36m_ablations_multi} we compare the performance without DNN initialization to the full method.
When not using DNN initialization we initialize with T-pose and template shape.
In the single-view case (\autoref{tab:ehf_ablations}, \autoref{tab:h36m_ablations_mono}), for the same number of iterations, we see significantly worse results, demonstrating that the initialization is critical to enable to optimizer to find a good minimum.
Results are far less stable in the absence of initialization, drastically affecting the robustness of the method.
In the multi-view case (\autoref{tab:h36m_ablations_multi}), the impact of initialization seems to be minor, likely because there is a great deal more data informing the optimization and therefore much less ambiguity for the optimizer to deal with.

\begin{figure*}
    \centering
    \footnotesize
    \begin{tabularx}{\linewidth}{@{}YYYYY@{}}
         Input & No Hand/Face DNNs & No Optimization & Full Method & GT \\
    \end{tabularx}
    \includegraphics[width=\linewidth]{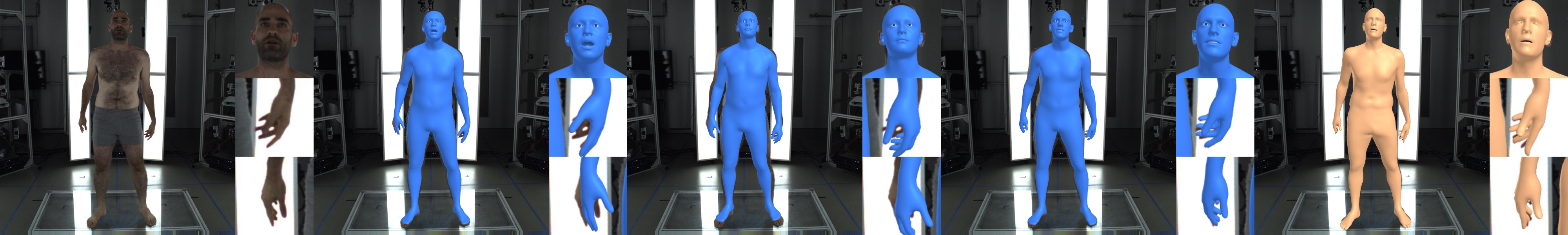}
    \includegraphics[width=\linewidth]{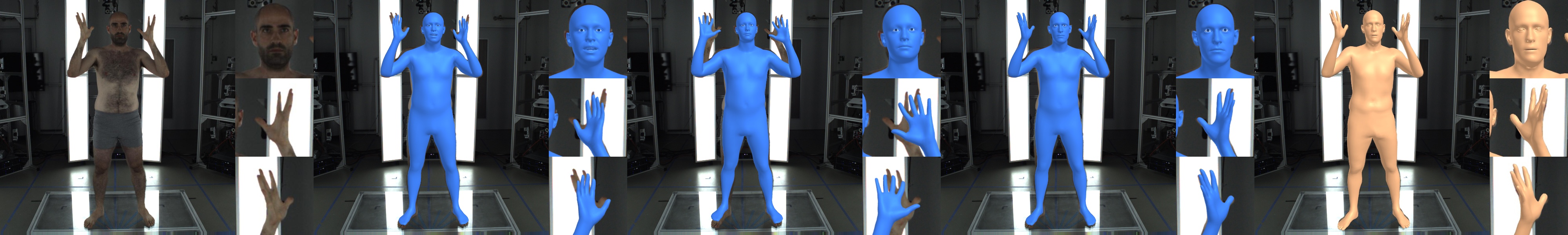}\\
    \includegraphics[width=\linewidth]{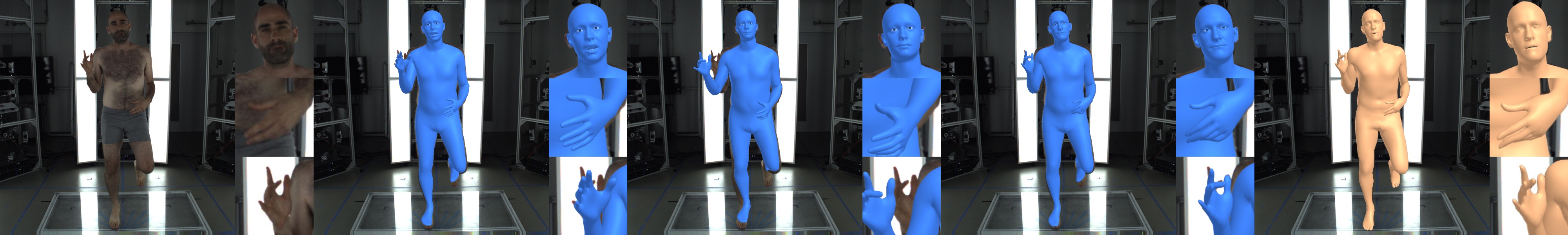}\\
    \vspace{-0.75em}
    \caption{Comparison of our method without hand or face DNNs, without optimization and with the full method on the EHF dataset~\cite{pavlakos2019expressive}.
    Without hand and face DNNs the quality of these areas is drastically reduced, without optimization image alignment is significantly poorer, and the face shape and expression is not captured at all as we do not initialize this using a DNN. \copylabel{EHF}{MPI}.} 
    \label{fig:opt_comp}
\end{figure*}

In \autoref{tab:ehf_ablations} and \autoref{tab:h36m_ablations_mono} we compare the performance of the DNN only (i.e., the method without optimization) to the full method.
The impact of optimization is generally small as the DNNs are able to directly produce quite accurate results.
The models are robust to diverse poses and image conditions, and appear to have learned a strong prior over both body pose and shape.
The optimization serves to fine-tune the precision of the results in terms of image-space alignment, fine-grained pose and shape detail and temporal stability.
This makes a large difference to perceived quality of the results, though this is not clearly expressed in the metrics used, see \autoref{fig:opt_comp}.

\subsection{Hand and Face DNNs}

Our method uses dedicated DNNs to refine the face and hand predictions.
\autoref{fig:opt_comp} shows results when not performing this refinement and just using the body DNN in isolation.
This significantly degrades performance for the face and hands.

\subsection{Probabilistic Landmarks}

We make use of dense probabilistic landmarks in our method to provide additional freedom to the optimizer based on the confidence of the DNN prediction.
In \autoref{tab:ehf_ablations}, \autoref{tab:h36m_ablations_mono} and \autoref{tab:h36m_ablations_multi} we show the effect of ignoring these confidence predictions and treating all landmarks with uniform certainty during optimization.
In all cases this has a negative impact on results, as the optimizer is `distracted' by poor landmark predictions and unable to fine-tune for those landmarks that are highly accurate.
In the multi-view case (\autoref{tab:h36m_ablations_multi}) the benefit of probabilistic landmarks appears to be greater, allowing the optimizer to use the best available viewpoint, while in the single-view case (\autoref{tab:h36m_ablations_mono}) the benefit is less significant.

\begin{figure}
    \centering
    \footnotesize
    \begin{minipage}{0.66\linewidth}\includegraphics[height=5.4cm]{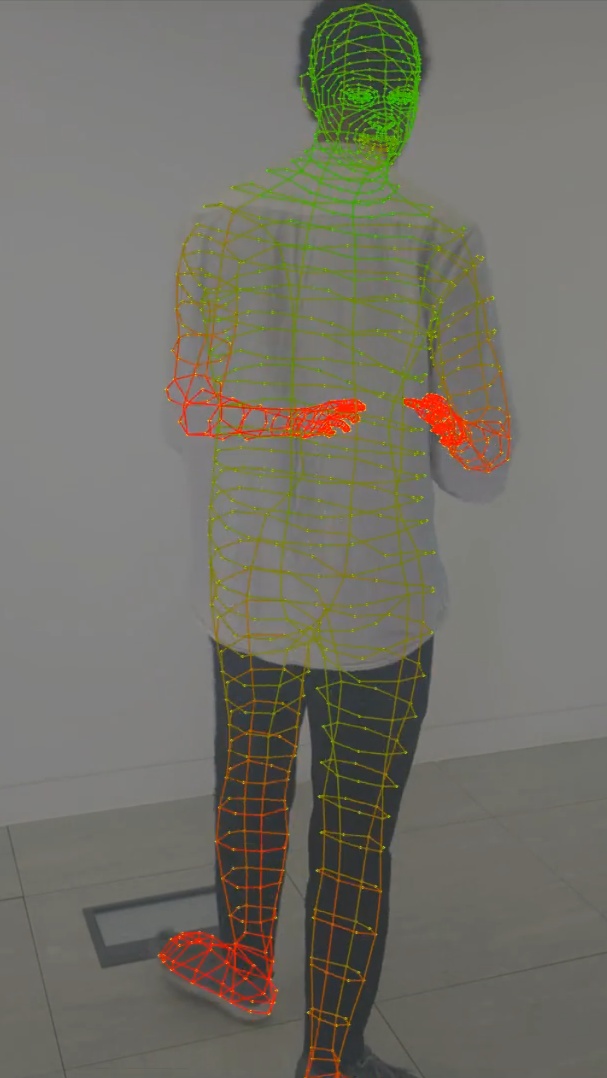}\includegraphics[height=5.4cm]{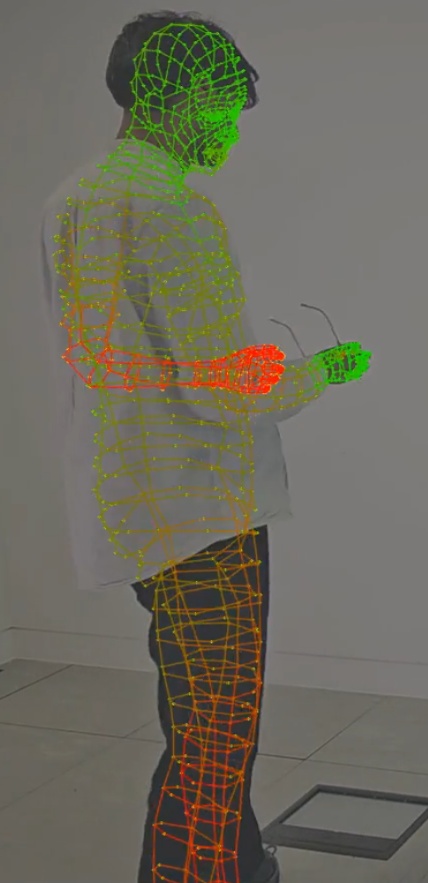}\end{minipage}
    \hfill
    \begin{minipage}{2.69cm}\includegraphics[height=2.69cm]{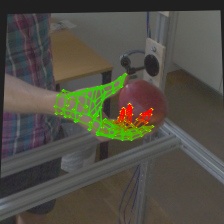}\\\includegraphics[height=2.69cm]{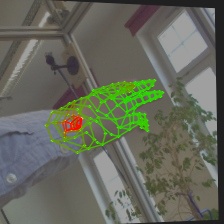}\end{minipage}
    \caption{Predicted dense landmarks for a multi-view, full-body scenario (left) and monocular hand scenario (right), landmarks are colored from least confident (red) to most confident (green). Occluded parts are less confident than visible parts meaning during model-fitting landmarks from views without occlusion (multi-view) or priors can be more heavily relied upon. \copylabel{FreiHAND}{LMB, University of Freiburg}.}
    \label{fig:prob_ldmks}
\end{figure}

\autoref{fig:prob_ldmks} shows the benefit of probabilistic landmarks in dealing with self-occlusions due to variation in body pose.
Parts that are occluded can be informed more strongly by observations from other views, or priors, leading to more plausible results in these cases.

\subsection{Priors}

\autoref{tab:h36m_ablations_mono} and \autoref{tab:h36m_ablations_multi} show the impact of the pose, shape and temporal priors on registration quality in monocular and multi-view scenarios.
In the monocular case all of these priors have a small beneficial effect on results.
In the multi-view case, however, the impact is negligible and in the case of the temporal prior it actually increases the MPJPE very slightly.
This is not unexpected; in the multi-view case there is sufficient data from the dense landmarks to almost completely specify the result, and we need to rely very little on priors.
In the monocular case there is often self-occlusion or depth ambiguity where we rely more heavily on the pose prior, or scale ambiguity where we rely more on the shape prior.
The temporal prior promotes smoothness of the motion, something that is perceptually important for performance capture, but not reflected at all in the MPJPE metric.

\subsection{Number of Views}

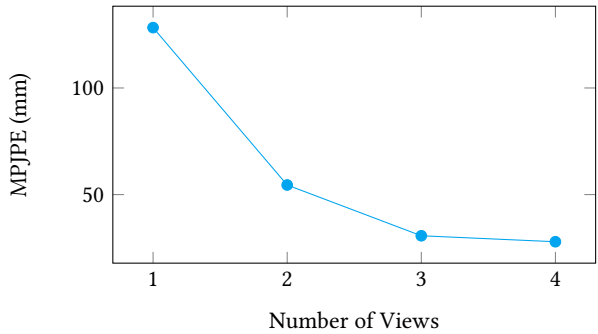
\begin{figure}
    \centering
    \begin{tikzpicture}
    \definecolor{ms_blue}{RGB}{0,164,239}
    \definecolor{ms_red}{RGB}{242,80,34}
    \definecolor{ms_green}{RGB}{127,186,0}
    \begin{axis}[
        xlabel=Number of Views,
        ylabel=MPJPE (mm),
        xtick={1,2,3,4},
        height=5cm,
        width=8cm,
    ]
    \addplot[mark=*,color=ms_blue] plot coordinates {
        (1, 128.3)
        (2, 54.5)
        (3, 30.7)
        (4, 27.9)
    };
    \end{axis}
    \end{tikzpicture}
    \caption{Variation in error for the Human3.6M benchmark~\cite{ionescu2014human3} with different numbers of cameras used as input. The more cameras the better the results, though with diminishing gains. MPJPE is \emph{not} pelvis-aligned.}
    \label{fig:h36m_num_cams}
\end{figure}

We use the Human3.6M benchmark~\cite{ionescu2014human3} to evaluate the impact of number of views on our method.
As our approach can take arbitrary input viewpoints, we run the reconstruction process with subsets of the four views available in the dataset.
As shown in \autoref{fig:h36m_num_cams}, results improve with each camera added, although the benefit diminishes with greater numbers of cameras.
In the monocular case there is a significant scale/translation ambiguity present, so it is no surprise that world-space (not pelvis-aligned) MPJPE is high in this case.
Adding a second view gives a strong cue for scale so drastically improves results.
After this each camera can help to reduce the impact of self-occlusions and further refine results, but with far more limited effect.
Going beyond four cameras we would expect the impact to further diminish as it is largely redundant information that is being added to the objective function.

\bibliographystyle{ACM-Reference-Format}
\bibliography{refs}


\begin{thebibliography}{87}

%%% the Web address.

\ifx \showCODEN    \undefined \def \showCODEN     #1{\unskip}     \fi
\ifx \showDOI      \undefined \def \showDOI       #1{#1}\fi
\ifx \showISBNx    \undefined \def \showISBNx     #1{\unskip}     \fi
\ifx \showISBNxiii \undefined \def \showISBNxiii  #1{\unskip}     \fi
\ifx \showISSN     \undefined \def \showISSN      #1{\unskip}     \fi
\ifx \showLCCN     \undefined \def \showLCCN      #1{\unskip}     \fi
\ifx \shownote     \undefined \def \shownote      #1{#1}          \fi
\ifx \showarticletitle \undefined \def \showarticletitle #1{#1}   \fi
\ifx \showURL      \undefined \def \showURL       {\relax}        \fi
% invisible to TeX
\providecommand\bibfield[2]{#2}
\providecommand\bibinfo[2]{#2}
\providecommand\natexlab[1]{#1}
\providecommand\showeprint[2][]{arXiv:#2}

\bibitem[Aliakbarian et~al\mbox{.}(2022)]%
        {aliakbarian2022flag}
\bibfield{author}{\bibinfo{person}{Sadegh Aliakbarian}, \bibinfo{person}{Pashmina Cameron}, \bibinfo{person}{Federica Bogo}, \bibinfo{person}{Andrew Fitzgibbon}, {and} \bibinfo{person}{Thomas~J Cashman}.} \bibinfo{year}{2022}\natexlab{}.
\newblock \showarticletitle{Flag: Flow-based 3d avatar generation from sparse observations}. In \bibinfo{booktitle}{\emph{CVPR}}. \bibinfo{pages}{13253--13262}.
\newblock


\bibitem[Allen et~al\mbox{.}(2003)]%
        {allen2003space}
\bibfield{author}{\bibinfo{person}{Brett Allen}, \bibinfo{person}{Brian Curless}, {and} \bibinfo{person}{Zoran Popovi{\'c}}.} \bibinfo{year}{2003}\natexlab{}.
\newblock \showarticletitle{The space of human body shapes: reconstruction and parameterization from range scans}.
\newblock \bibinfo{journal}{\emph{ACM TOG}} \bibinfo{volume}{22}, \bibinfo{number}{3} (\bibinfo{year}{2003}), \bibinfo{pages}{587--594}.
\newblock


\bibitem[Alvar and Baji{\'c}(2022)]%
        {alvar2022practical}
\bibfield{author}{\bibinfo{person}{Saeed~Ranjbar Alvar} {and} \bibinfo{person}{Ivan~V Baji{\'c}}.} \bibinfo{year}{2022}\natexlab{}.
\newblock \showarticletitle{Practical noise simulation for rgb images}.
\newblock \bibinfo{journal}{\emph{arXiv preprint arXiv:2201.12773}} (\bibinfo{year}{2022}).
\newblock


\bibitem[Anguelov et~al\mbox{.}(2005)]%
        {anguelov2005scape}
\bibfield{author}{\bibinfo{person}{Dragomir Anguelov}, \bibinfo{person}{Praveen Srinivasan}, \bibinfo{person}{Daphne Koller}, \bibinfo{person}{Sebastian Thrun}, \bibinfo{person}{Jim Rodgers}, {and} \bibinfo{person}{James Davis}.} \bibinfo{year}{2005}\natexlab{}.
\newblock \showarticletitle{{SCAPE:} Shape Completion and Animation of People}.
\newblock \bibinfo{journal}{\emph{ACM TOG}}  \bibinfo{volume}{24} (\bibinfo{year}{2005}), \bibinfo{pages}{408--416}.
\newblock


\bibitem[Bai et~al\mbox{.}(2021)]%
        {Bai_2021_CVPR}
\bibfield{author}{\bibinfo{person}{Ziqian Bai}, \bibinfo{person}{Zhaopeng Cui}, \bibinfo{person}{Xiaoming Liu}, {and} \bibinfo{person}{Ping Tan}.} \bibinfo{year}{2021}\natexlab{}.
\newblock \showarticletitle{{Riggable 3D Face Reconstruction via In-Network Optimization}}. In \bibinfo{booktitle}{\emph{CVPR}}.
\newblock


\bibitem[Baradel* et~al\mbox{.}(2024)]%
        {multi-hmr2024}
\bibfield{author}{\bibinfo{person}{Fabien Baradel*}, \bibinfo{person}{Matthieu Armando}, \bibinfo{person}{Salma Galaaoui}, \bibinfo{person}{Romain Br{\'e}gier}, \bibinfo{person}{Philippe Weinzaepfel}, \bibinfo{person}{Gr{\'e}gory Rogez}, {and} \bibinfo{person}{Thomas Lucas*}.} \bibinfo{year}{2024}\natexlab{}.
\newblock \showarticletitle{Multi-HMR: Multi-Person Whole-Body Human Mesh Recovery in a Single Shot}. In \bibinfo{booktitle}{\emph{ECCV}}.
\newblock


\bibitem[Black et~al\mbox{.}(2023)]%
        {black2023bedlam}
\bibfield{author}{\bibinfo{person}{Michael~J Black}, \bibinfo{person}{Priyanka Patel}, \bibinfo{person}{Joachim Tesch}, {and} \bibinfo{person}{Jinlong Yang}.} \bibinfo{year}{2023}\natexlab{}.
\newblock \showarticletitle{BEDLAM: A Synthetic Dataset of Bodies Exhibiting Detailed Lifelike Animated Motion}. In \bibinfo{booktitle}{\emph{CVPR}}. \bibinfo{pages}{8726--8737}.
\newblock


\bibitem[{Blender Foundation}(2021)]%
        {CyclesRenderer}
\bibfield{author}{\bibinfo{person}{{Blender Foundation}}.} \bibinfo{year}{2021}\natexlab{}.
\newblock \bibinfo{title}{Cycles Renderer}.
\newblock \bibinfo{howpublished}{\url{https://www.cycles-renderer.org/}}.
\newblock


\bibitem[Cai et~al\mbox{.}(2024)]%
        {cai2024smpler}
\bibfield{author}{\bibinfo{person}{Zhongang Cai}, \bibinfo{person}{Wanqi Yin}, \bibinfo{person}{Ailing Zeng}, \bibinfo{person}{Chen Wei}, \bibinfo{person}{Qingping Sun}, \bibinfo{person}{Wang Yanjun}, \bibinfo{person}{Hui~En Pang}, \bibinfo{person}{Haiyi Mei}, \bibinfo{person}{Mingyuan Zhang}, \bibinfo{person}{Lei Zhang}, {et~al\mbox{.}}} \bibinfo{year}{2024}\natexlab{}.
\newblock \showarticletitle{Smpler-x: Scaling up expressive human pose and shape estimation}.
\newblock \bibinfo{journal}{\emph{NeurIPS}}  \bibinfo{volume}{36} (\bibinfo{year}{2024}).
\newblock


\bibitem[{Chocofur}(2024)]%
        {chocofur}
\bibfield{author}{\bibinfo{person}{{Chocofur}}.} \bibinfo{year}{2024}\natexlab{}.
\newblock \bibinfo{title}{Chocofur Interior Scene}.
\newblock \bibinfo{howpublished}{\url{https://store.chocofur.com/scenes-interior}}.
\newblock


\bibitem[Choudhury et~al\mbox{.}(2023)]%
        {choudhury2023tempo}
\bibfield{author}{\bibinfo{person}{Rohan Choudhury}, \bibinfo{person}{Kris~M Kitani}, {and} \bibinfo{person}{L{\'a}szl{\'o}~A Jeni}.} \bibinfo{year}{2023}\natexlab{}.
\newblock \showarticletitle{TEMPO: Efficient multi-view pose estimation, tracking, and forecasting}. In \bibinfo{booktitle}{\emph{ICCV}}. \bibinfo{pages}{14750--14760}.
\newblock


\bibitem[Choutas et~al\mbox{.}(2022a)]%
        {choutas2022learning}
\bibfield{author}{\bibinfo{person}{Vasileios Choutas}, \bibinfo{person}{Federica Bogo}, \bibinfo{person}{Jingjing Shen}, {and} \bibinfo{person}{Julien Valentin}.} \bibinfo{year}{2022}\natexlab{a}.
\newblock \showarticletitle{Learning to fit morphable models}. In \bibinfo{booktitle}{\emph{ECCV}}. \bibinfo{pages}{160--179}.
\newblock


\bibitem[Choutas et~al\mbox{.}(2022b)]%
        {Shapy:2022}
\bibfield{author}{\bibinfo{person}{Vasileios Choutas}, \bibinfo{person}{Lea M{\"u}ller}, \bibinfo{person}{Chun-Hao~P. Huang}, \bibinfo{person}{Siyu Tang}, \bibinfo{person}{Dimitrios Tzionas}, {and} \bibinfo{person}{Michael~J. Black}.} \bibinfo{year}{2022}\natexlab{b}.
\newblock \showarticletitle{Accurate {3D} Body Shape Regression using Metric and Semantic Attributes}. In \bibinfo{booktitle}{\emph{CVPR}}. \bibinfo{pages}{2718--2728}.
\newblock


\bibitem[{CLO Virtual Fashion Inc.}(2024)]%
        {mervelous}
\bibfield{author}{\bibinfo{person}{{CLO Virtual Fashion Inc.}}} \bibinfo{year}{2024}\natexlab{}.
\newblock \bibinfo{title}{MarvelousDesigner}.
\newblock \bibinfo{howpublished}{\url{https://marvelousdesigner.com/}}.
\newblock


\bibitem[Debevec(2006)]%
        {debevec2006image}
\bibfield{author}{\bibinfo{person}{Paul Debevec}.} \bibinfo{year}{2006}\natexlab{}.
\newblock \showarticletitle{Image-based lighting}.
\newblock In \bibinfo{booktitle}{\emph{ACM SIGGRAPH 2006 Courses}}. \bibinfo{pages}{4--es}.
\newblock


\bibitem[Deng et~al\mbox{.}(2009)]%
        {deng2009imagenet}
\bibfield{author}{\bibinfo{person}{Jia Deng}, \bibinfo{person}{Wei Dong}, \bibinfo{person}{Richard Socher}, \bibinfo{person}{Li-Jia Li}, \bibinfo{person}{Kai Li}, {and} \bibinfo{person}{Li Fei-Fei}.} \bibinfo{year}{2009}\natexlab{}.
\newblock \showarticletitle{{ImageNet}: A large-scale hierarchical image database}. In \bibinfo{booktitle}{\emph{CVPR}}. \bibinfo{pages}{248--255}.
\newblock


\bibitem[Dinh et~al\mbox{.}(2017)]%
        {dinh2016density}
\bibfield{author}{\bibinfo{person}{Laurent Dinh}, \bibinfo{person}{Jascha Sohl-Dickstein}, {and} \bibinfo{person}{Samy Bengio}.} \bibinfo{year}{2017}\natexlab{}.
\newblock \showarticletitle{Density Estimation Using Real {NVP}}. In \bibinfo{booktitle}{\emph{ICLR}}.
\newblock


\bibitem[{Evermotion}(2024)]%
        {evermotion}
\bibfield{author}{\bibinfo{person}{{Evermotion}}.} \bibinfo{year}{2024}\natexlab{}.
\newblock \bibinfo{title}{Evermotion}.
\newblock \bibinfo{howpublished}{\url{https://evermotion.org/}}.
\newblock


\bibitem[{Faceware Technologies}(2024)]%
        {faceware}
\bibfield{author}{\bibinfo{person}{{Faceware Technologies}}.} \bibinfo{year}{2024}\natexlab{}.
\newblock \bibinfo{title}{Faceware}.
\newblock \bibinfo{howpublished}{\url{https://facewaretech.com/}}.
\newblock


\bibitem[Feng et~al\mbox{.}(2021)]%
        {feng2021collaborative}
\bibfield{author}{\bibinfo{person}{Yao Feng}, \bibinfo{person}{Vasileios Choutas}, \bibinfo{person}{Timo Bolkart}, \bibinfo{person}{Dimitrios Tzionas}, {and} \bibinfo{person}{Michael~J Black}.} \bibinfo{year}{2021}\natexlab{}.
\newblock \showarticletitle{Collaborative Regression of Expressive Bodies using Moderation}. In \bibinfo{booktitle}{\emph{3DV}}.
\newblock


\bibitem[Fischler and Bolles(1981)]%
        {fischler1981random}
\bibfield{author}{\bibinfo{person}{Martin~A Fischler} {and} \bibinfo{person}{Robert~C Bolles}.} \bibinfo{year}{1981}\natexlab{}.
\newblock \showarticletitle{Random sample consensus: a paradigm for model fitting with applications to image analysis and automated cartography}.
\newblock \bibinfo{journal}{\emph{Commun. ACM}} \bibinfo{volume}{24}, \bibinfo{number}{6} (\bibinfo{year}{1981}), \bibinfo{pages}{381--395}.
\newblock


\bibitem[Goel et~al\mbox{.}(2023)]%
        {goel2023humans}
\bibfield{author}{\bibinfo{person}{Shubham Goel}, \bibinfo{person}{Georgios Pavlakos}, \bibinfo{person}{Jathushan Rajasegaran}, \bibinfo{person}{Angjoo Kanazawa}, {and} \bibinfo{person}{Jitendra Malik}.} \bibinfo{year}{2023}\natexlab{}.
\newblock \showarticletitle{Humans in 4D: Reconstructing and tracking humans with transformers}. In \bibinfo{booktitle}{\emph{CVPR}}. \bibinfo{pages}{14783--14794}.
\newblock


\bibitem[He et~al\mbox{.}(2016)]%
        {he2016deep}
\bibfield{author}{\bibinfo{person}{Kaiming He}, \bibinfo{person}{Xiangyu Zhang}, \bibinfo{person}{Shaoqing Ren}, {and} \bibinfo{person}{Jian Sun}.} \bibinfo{year}{2016}\natexlab{}.
\newblock \showarticletitle{Deep Residual Learning for Image Recognition}. In \bibinfo{booktitle}{\emph{CVPR}}. \bibinfo{pages}{770--778}.
\newblock


\bibitem[Ionescu et~al\mbox{.}(2014)]%
        {ionescu2014human3}
\bibfield{author}{\bibinfo{person}{Catalin Ionescu}, \bibinfo{person}{Dragos Papava}, \bibinfo{person}{Vlad Olaru}, {and} \bibinfo{person}{Cristian Sminchisescu}.} \bibinfo{year}{2014}\natexlab{}.
\newblock \showarticletitle{{Human3.6M}: Large scale datasets and predictive methods for {3D} human sensing in natural environments}.
\newblock \bibinfo{journal}{\emph{TPAMI}} \bibinfo{volume}{36}, \bibinfo{number}{7} (\bibinfo{year}{2014}), \bibinfo{pages}{1325--1339}.
\newblock


\bibitem[Iskakov et~al\mbox{.}(2019)]%
        {iskakov2019learnable}
\bibfield{author}{\bibinfo{person}{Karim Iskakov}, \bibinfo{person}{Egor Burkov}, \bibinfo{person}{Victor Lempitsky}, {and} \bibinfo{person}{Yury Malkov}.} \bibinfo{year}{2019}\natexlab{}.
\newblock \showarticletitle{Learnable Triangulation of Human Pose}. In \bibinfo{booktitle}{\emph{ICCV}}.
\newblock


\bibitem[Joo et~al\mbox{.}(2018)]%
        {joo2018total}
\bibfield{author}{\bibinfo{person}{Hanbyul Joo}, \bibinfo{person}{Tomas Simon}, {and} \bibinfo{person}{Yaser Sheikh}.} \bibinfo{year}{2018}\natexlab{}.
\newblock \showarticletitle{Total Capture: A {3D} Deformation Model for Tracking Faces, Hands, and Bodies}. In \bibinfo{booktitle}{\emph{CVPR}}. \bibinfo{pages}{8320--8329}.
\newblock


\bibitem[Kanazawa et~al\mbox{.}(2018)]%
        {Kanazawa2018_hmr}
\bibfield{author}{\bibinfo{person}{Angjoo Kanazawa}, \bibinfo{person}{Michael~J. Black}, \bibinfo{person}{David~W. Jacobs}, {and} \bibinfo{person}{Jitendra Malik}.} \bibinfo{year}{2018}\natexlab{}.
\newblock \showarticletitle{End-to-end Recovery of Human Shape and Pose}. In \bibinfo{booktitle}{\emph{CVPR}}. \bibinfo{pages}{7122--7131}.
\newblock


\bibitem[Khamis et~al\mbox{.}(2015)]%
        {khamis2015learning}
\bibfield{author}{\bibinfo{person}{Sameh Khamis}, \bibinfo{person}{Jonathan Taylor}, \bibinfo{person}{Jamie Shotton}, \bibinfo{person}{Cem Keskin}, \bibinfo{person}{Shahram Izadi}, {and} \bibinfo{person}{Andrew Fitzgibbon}.} \bibinfo{year}{2015}\natexlab{}.
\newblock \showarticletitle{Learning an Efficient Model of Hand Shape Variation from Depth Images}. In \bibinfo{booktitle}{\emph{CVPR}}. \bibinfo{pages}{2540--2548}.
\newblock


\bibitem[Kingma and Welling(2014)]%
        {kingma2013auto}
\bibfield{author}{\bibinfo{person}{Diederik~P Kingma} {and} \bibinfo{person}{Max Welling}.} \bibinfo{year}{2014}\natexlab{}.
\newblock \showarticletitle{Auto-Encoding Variational Bayes}. In \bibinfo{booktitle}{\emph{ICLR}}.
\newblock


\bibitem[Kolotouros et~al\mbox{.}(2019)]%
        {Kolotouros2019_spin}
\bibfield{author}{\bibinfo{person}{Nikos Kolotouros}, \bibinfo{person}{Georgios Pavlakos}, \bibinfo{person}{Michael~J. Black}, {and} \bibinfo{person}{Kostas Daniilidis}.} \bibinfo{year}{2019}\natexlab{}.
\newblock \showarticletitle{Learning to Reconstruct {3D} Human Pose and Shape via Model-Fitting in the Loop}. In \bibinfo{booktitle}{\emph{ICCV}}. \bibinfo{pages}{2252--2261}.
\newblock


\bibitem[Kolotouros et~al\mbox{.}(2021)]%
        {kolotouros2021probabilistic}
\bibfield{author}{\bibinfo{person}{Nikos Kolotouros}, \bibinfo{person}{Georgios Pavlakos}, \bibinfo{person}{Dinesh Jayaraman}, {and} \bibinfo{person}{Kostas Daniilidis}.} \bibinfo{year}{2021}\natexlab{}.
\newblock \showarticletitle{Probabilistic modeling for human mesh recovery}. In \bibinfo{booktitle}{\emph{ICCV}}. \bibinfo{pages}{11605--11614}.
\newblock


\bibitem[Li et~al\mbox{.}(2023)]%
        {li2023hybrik}
\bibfield{author}{\bibinfo{person}{Jiefeng Li}, \bibinfo{person}{Siyuan Bian}, \bibinfo{person}{Chao Xu}, \bibinfo{person}{Zhicun Chen}, \bibinfo{person}{Lixin Yang}, {and} \bibinfo{person}{Cewu Lu}.} \bibinfo{year}{2023}\natexlab{}.
\newblock \showarticletitle{HybrIK-X: Hybrid Analytical-Neural Inverse Kinematics for Whole-body Mesh Recovery}.
\newblock \bibinfo{journal}{\emph{arXiv preprint arXiv:2304.05690}} (\bibinfo{year}{2023}).
\newblock


\bibitem[Li et~al\mbox{.}(2021)]%
        {li2021hybrik}
\bibfield{author}{\bibinfo{person}{Jiefeng Li}, \bibinfo{person}{Chao Xu}, \bibinfo{person}{Zhicun Chen}, \bibinfo{person}{Siyuan Bian}, \bibinfo{person}{Lixin Yang}, {and} \bibinfo{person}{Cewu Lu}.} \bibinfo{year}{2021}\natexlab{}.
\newblock \showarticletitle{{HybrIK}: A Hybrid Analytical-Neural Inverse Kinematics Solution for {3D} Human Pose and Shape Estimation}. In \bibinfo{booktitle}{\emph{CVPR}}. \bibinfo{pages}{3383--3393}.
\newblock


\bibitem[Li et~al\mbox{.}(2017)]%
        {li2017learning}
\bibfield{author}{\bibinfo{person}{Tianye Li}, \bibinfo{person}{Timo Bolkart}, \bibinfo{person}{Michael~J Black}, \bibinfo{person}{Hao Li}, {and} \bibinfo{person}{Javier Romero}.} \bibinfo{year}{2017}\natexlab{}.
\newblock \showarticletitle{Learning a model of facial shape and expression from {4D} scans}.
\newblock \bibinfo{journal}{\emph{ACM TOG}} \bibinfo{volume}{36}, \bibinfo{number}{6} (\bibinfo{year}{2017}), \bibinfo{pages}{194--1}.
\newblock


\bibitem[Lin et~al\mbox{.}(2023)]%
        {lin2023one}
\bibfield{author}{\bibinfo{person}{Jing Lin}, \bibinfo{person}{Ailing Zeng}, \bibinfo{person}{Haoqian Wang}, \bibinfo{person}{Lei Zhang}, {and} \bibinfo{person}{Yu Li}.} \bibinfo{year}{2023}\natexlab{}.
\newblock \showarticletitle{One-Stage 3D Whole-Body Mesh Recovery with Component Aware Transformer}. In \bibinfo{booktitle}{\emph{CVPR}}. \bibinfo{pages}{21159--21168}.
\newblock


\bibitem[Lin et~al\mbox{.}(2021a)]%
        {lin2021end}
\bibfield{author}{\bibinfo{person}{Kevin Lin}, \bibinfo{person}{Lijuan Wang}, {and} \bibinfo{person}{Zicheng Liu}.} \bibinfo{year}{2021}\natexlab{a}.
\newblock \showarticletitle{End-to-end human pose and mesh reconstruction with transformers}. In \bibinfo{booktitle}{\emph{CVPR}}. \bibinfo{pages}{1954--1963}.
\newblock


\bibitem[Lin et~al\mbox{.}(2021b)]%
        {lin2021mesh}
\bibfield{author}{\bibinfo{person}{Kevin Lin}, \bibinfo{person}{Lijuan Wang}, {and} \bibinfo{person}{Zicheng Liu}.} \bibinfo{year}{2021}\natexlab{b}.
\newblock \showarticletitle{Mesh Graphormer}. In \bibinfo{booktitle}{\emph{ICCV}}.
\newblock


\bibitem[Ling et~al\mbox{.}(2022)]%
        {ling2022vectoradam}
\bibfield{author}{\bibinfo{person}{Selena~Zihan Ling}, \bibinfo{person}{Nicholas Sharp}, {and} \bibinfo{person}{Alec Jacobson}.} \bibinfo{year}{2022}\natexlab{}.
\newblock \showarticletitle{VectorAdam for Rotation Equivariant Geometry Optimization}.
\newblock \bibinfo{journal}{\emph{NeurIPS}}  \bibinfo{volume}{35} (\bibinfo{year}{2022}), \bibinfo{pages}{4111--4122}.
\newblock


\bibitem[Liu and Nocedal(1989)]%
        {liu1989limited}
\bibfield{author}{\bibinfo{person}{Dong~C Liu} {and} \bibinfo{person}{Jorge Nocedal}.} \bibinfo{year}{1989}\natexlab{}.
\newblock \showarticletitle{On the limited memory BFGS method for large scale optimization}.
\newblock \bibinfo{journal}{\emph{Mathematical programming}} \bibinfo{volume}{45}, \bibinfo{number}{1-3} (\bibinfo{year}{1989}), \bibinfo{pages}{503--528}.
\newblock


\bibitem[Loper et~al\mbox{.}(2014)]%
        {loper2014mosh}
\bibfield{author}{\bibinfo{person}{Matthew Loper}, \bibinfo{person}{Naureen Mahmood}, {and} \bibinfo{person}{Michael~J Black}.} \bibinfo{year}{2014}\natexlab{}.
\newblock \showarticletitle{{MoSh}: Motion and Shape Capture from Sparse Markers}.
\newblock \bibinfo{journal}{\emph{ACM TOG}} \bibinfo{volume}{33}, \bibinfo{number}{6} (\bibinfo{year}{2014}), \bibinfo{pages}{1--13}.
\newblock


\bibitem[Loper et~al\mbox{.}(2015)]%
        {loper2015smpl}
\bibfield{author}{\bibinfo{person}{Matthew Loper}, \bibinfo{person}{Naureen Mahmood}, \bibinfo{person}{Javier Romero}, \bibinfo{person}{Gerard Pons-Moll}, {and} \bibinfo{person}{Michael~J Black}.} \bibinfo{year}{2015}\natexlab{}.
\newblock \showarticletitle{{SMPL}: A Skinned Multi-Person Linear Model}.
\newblock \bibinfo{journal}{\emph{ACM TOG}} \bibinfo{volume}{34}, \bibinfo{number}{6} (\bibinfo{year}{2015}).
\newblock


\bibitem[Loshchilov and Hutter(2017)]%
        {loshchilov2017decoupled}
\bibfield{author}{\bibinfo{person}{Ilya Loshchilov} {and} \bibinfo{person}{Frank Hutter}.} \bibinfo{year}{2017}\natexlab{}.
\newblock \showarticletitle{Decoupled weight decay regularization}.
\newblock \bibinfo{journal}{\emph{arXiv preprint arXiv:1711.05101}} (\bibinfo{year}{2017}).
\newblock


\bibitem[Ma et~al\mbox{.}(2020)]%
        {ma2020learning}
\bibfield{author}{\bibinfo{person}{Qianli Ma}, \bibinfo{person}{Jinlong Yang}, \bibinfo{person}{Anurag Ranjan}, \bibinfo{person}{Sergi Pujades}, \bibinfo{person}{Gerard Pons-Moll}, \bibinfo{person}{Siyu Tang}, {and} \bibinfo{person}{Michael~J Black}.} \bibinfo{year}{2020}\natexlab{}.
\newblock \showarticletitle{Learning to dress 3d people in generative clothing}. In \bibinfo{booktitle}{\emph{CVPR}}. \bibinfo{pages}{6469--6478}.
\newblock


\bibitem[Mahmood et~al\mbox{.}(2019)]%
        {mahmood2019amass}
\bibfield{author}{\bibinfo{person}{Naureen Mahmood}, \bibinfo{person}{Nima Ghorbani}, \bibinfo{person}{Nikolaus~F Troje}, \bibinfo{person}{Gerard Pons-Moll}, {and} \bibinfo{person}{Michael~J Black}.} \bibinfo{year}{2019}\natexlab{}.
\newblock \showarticletitle{{AMASS}: Archive of Motion Capture as Surface Shapes}. In \bibinfo{booktitle}{\emph{ICCV}}.
\newblock


\bibitem[{Manus-Meta}(2024)]%
        {manus}
\bibfield{author}{\bibinfo{person}{{Manus-Meta}}.} \bibinfo{year}{2024}\natexlab{}.
\newblock \bibinfo{title}{Manus}.
\newblock \bibinfo{howpublished}{\url{https://www.manus-meta.com/}}.
\newblock


\bibitem[Moon et~al\mbox{.}(2022)]%
        {moon2022Hand4Whole}
\bibfield{author}{\bibinfo{person}{Gyeongsik Moon}, \bibinfo{person}{Hongsuk Choi}, {and} \bibinfo{person}{Kyoung~Mu Lee}.} \bibinfo{year}{2022}\natexlab{}.
\newblock \showarticletitle{Accurate 3D hand pose estimation for whole-body 3D human mesh estimation}. In \bibinfo{booktitle}{\emph{CVPR}}. \bibinfo{pages}{2308--2317}.
\newblock


\bibitem[Moon and Lee(2020)]%
        {moon2020i2l}
\bibfield{author}{\bibinfo{person}{Gyeongsik Moon} {and} \bibinfo{person}{Kyoung~Mu Lee}.} \bibinfo{year}{2020}\natexlab{}.
\newblock \showarticletitle{{I2L-MeshNet}: Image-to-Lixel Prediction Network for Accurate {3D} Human Pose and Mesh Estimation from a Single {RGB} Image}. In \bibinfo{booktitle}{\emph{ECCV}}. \bibinfo{pages}{752--768}.
\newblock


\bibitem[Moon et~al\mbox{.}(2023)]%
        {moon2023reinterhand}
\bibfield{author}{\bibinfo{person}{Gyeongsik Moon}, \bibinfo{person}{Shunsuke Saito}, \bibinfo{person}{Weipeng Xu}, \bibinfo{person}{Rohan Joshi}, \bibinfo{person}{Julia Buffalini}, \bibinfo{person}{Harley Bellan}, \bibinfo{person}{Nicholas Rosen}, \bibinfo{person}{Jesse Richardson}, \bibinfo{person}{Mize Mallorie}, \bibinfo{person}{Philippe Bree}, \bibinfo{person}{Tomas Simon}, \bibinfo{person}{Bo Peng}, \bibinfo{person}{Shubham Garg}, \bibinfo{person}{Kevyn McPhail}, {and} \bibinfo{person}{Takaaki Shiratori}.} \bibinfo{year}{2023}\natexlab{}.
\newblock \showarticletitle{A Dataset of Relighted {3D} Interacting Hands}. In \bibinfo{booktitle}{\emph{NeurIPS Track on Datasets and Benchmarks}}.
\newblock


\bibitem[Moon et~al\mbox{.}(2020)]%
        {Moon_2020_ECCV_InterHand2.6M}
\bibfield{author}{\bibinfo{person}{Gyeongsik Moon}, \bibinfo{person}{Shoou-I Yu}, \bibinfo{person}{He Wen}, \bibinfo{person}{Takaaki Shiratori}, {and} \bibinfo{person}{Kyoung~Mu Lee}.} \bibinfo{year}{2020}\natexlab{}.
\newblock \showarticletitle{InterHand2.6M: A Dataset and Baseline for 3D Interacting Hand Pose Estimation from a Single RGB Image}. In \bibinfo{booktitle}{\emph{ECCV}}.
\newblock


\bibitem[{Move AI}(2024)]%
        {moveai}
\bibfield{author}{\bibinfo{person}{{Move AI}}.} \bibinfo{year}{2024}\natexlab{}.
\newblock \bibinfo{title}{Move AI}.
\newblock \bibinfo{howpublished}{\url{https://www.move.ai/}}.
\newblock


\bibitem[{Movella}(2024)]%
        {xsens}
\bibfield{author}{\bibinfo{person}{{Movella}}.} \bibinfo{year}{2024}\natexlab{}.
\newblock \bibinfo{title}{XSens}.
\newblock \bibinfo{howpublished}{\url{https://www.movella.com/products/motion-capture}}.
\newblock


\bibitem[{Natural Point}(2024)]%
        {optitrack}
\bibfield{author}{\bibinfo{person}{{Natural Point}}.} \bibinfo{year}{2024}\natexlab{}.
\newblock \bibinfo{title}{OptiTrack}.
\newblock \bibinfo{howpublished}{\url{https://www.optitrack.com/}}.
\newblock


\bibitem[Nicolet et~al\mbox{.}(2021)]%
        {Nicolet2021Large}
\bibfield{author}{\bibinfo{person}{Baptiste Nicolet}, \bibinfo{person}{Alec Jacobson}, {and} \bibinfo{person}{Wenzel Jakob}.} \bibinfo{year}{2021}\natexlab{}.
\newblock \showarticletitle{Large Steps in Inverse Rendering of Geometry}.
\newblock \bibinfo{journal}{\emph{SIGGRAPH Asia}} \bibinfo{volume}{40}, \bibinfo{number}{6} (\bibinfo{date}{Dec.} \bibinfo{year}{2021}).
\newblock


\bibitem[Pavlakos et~al\mbox{.}(2019)]%
        {pavlakos2019expressive}
\bibfield{author}{\bibinfo{person}{Georgios Pavlakos}, \bibinfo{person}{Vasileios Choutas}, \bibinfo{person}{Nima Ghorbani}, \bibinfo{person}{Timo Bolkart}, \bibinfo{person}{Ahmed~AA Osman}, \bibinfo{person}{Dimitrios Tzionas}, {and} \bibinfo{person}{Michael~J Black}.} \bibinfo{year}{2019}\natexlab{}.
\newblock \showarticletitle{Expressive body capture: {3D} hands, face, and body from a single image}. In \bibinfo{booktitle}{\emph{CVPR}}. \bibinfo{pages}{10975--10985}.
\newblock


\bibitem[Pavlakos et~al\mbox{.}(2017)]%
        {pavlakos2017harvesting}
\bibfield{author}{\bibinfo{person}{Georgios Pavlakos}, \bibinfo{person}{Xiaowei Zhou}, \bibinfo{person}{Konstantinos~G Derpanis}, {and} \bibinfo{person}{Kostas Daniilidis}.} \bibinfo{year}{2017}\natexlab{}.
\newblock \showarticletitle{Harvesting multiple views for marker-less 3d human pose annotations}. In \bibinfo{booktitle}{\emph{CVPR}}. \bibinfo{pages}{6988--6997}.
\newblock


\bibitem[{Poly Haven}(2024)]%
        {polyhaven}
\bibfield{author}{\bibinfo{person}{{Poly Haven}}.} \bibinfo{year}{2024}\natexlab{}.
\newblock \bibinfo{title}{Poly Haven}.
\newblock \bibinfo{howpublished}{\url{https://polyhaven.com/hdris}}.
\newblock


\bibitem[Rai et~al\mbox{.}(2024)]%
        {rai2023towards}
\bibfield{author}{\bibinfo{person}{Aashish Rai}, \bibinfo{person}{Hiresh Gupta}, \bibinfo{person}{Ayush Pandey}, \bibinfo{person}{Francisco~Vicente Carrasco}, \bibinfo{person}{Shingo~Jason Takagi}, \bibinfo{person}{Amaury Aubel}, \bibinfo{person}{Daeil Kim}, \bibinfo{person}{Aayush Prakash}, {and} \bibinfo{person}{Fernando De~la Torre}.} \bibinfo{year}{2024}\natexlab{}.
\newblock \showarticletitle{Towards realistic generative 3d face models}. In \bibinfo{booktitle}{\emph{WACV}}. \bibinfo{pages}{3738--3748}.
\newblock


\bibitem[Rempe et~al\mbox{.}(2021)]%
        {rempe2021humor}
\bibfield{author}{\bibinfo{person}{Davis Rempe}, \bibinfo{person}{Tolga Birdal}, \bibinfo{person}{Aaron Hertzmann}, \bibinfo{person}{Jimei Yang}, \bibinfo{person}{Srinath Sridhar}, {and} \bibinfo{person}{Leonidas~J. Guibas}.} \bibinfo{year}{2021}\natexlab{}.
\newblock \showarticletitle{{HuMoR}: {3D} Human Motion Model for Robust Pose Estimation}. In \bibinfo{booktitle}{\emph{ICCV}}.
\newblock


\bibitem[{Renderpeople}(2024)]%
        {renderpeople}
\bibfield{author}{\bibinfo{person}{{Renderpeople}}.} \bibinfo{year}{2024}\natexlab{}.
\newblock \bibinfo{title}{Renderpeople}.
\newblock \bibinfo{howpublished}{\url{https://renderpeople.com/}}.
\newblock


\bibitem[Rezende and Mohamed(2015)]%
        {rezende2015variational}
\bibfield{author}{\bibinfo{person}{Danilo Rezende} {and} \bibinfo{person}{Shakir Mohamed}.} \bibinfo{year}{2015}\natexlab{}.
\newblock \showarticletitle{Variational Inference with Normalizing Flows}. In \bibinfo{booktitle}{\emph{ICML}}. \bibinfo{pages}{1530--1538}.
\newblock


\bibitem[Romero et~al\mbox{.}(2017)]%
        {romero2017embodied}
\bibfield{author}{\bibinfo{person}{Javier Romero}, \bibinfo{person}{Dimitrios Tzionas}, {and} \bibinfo{person}{Michael~J Black}.} \bibinfo{year}{2017}\natexlab{}.
\newblock \showarticletitle{Embodied Hands: Modeling and Capturing Hands and Bodies Together}.
\newblock \bibinfo{journal}{\emph{ACM TOG}} \bibinfo{volume}{36}, \bibinfo{number}{6} (\bibinfo{year}{2017}), \bibinfo{pages}{1--17}.
\newblock


\bibitem[Rong et~al\mbox{.}(2021)]%
        {rong2020frankmocap}
\bibfield{author}{\bibinfo{person}{Yu Rong}, \bibinfo{person}{Takaaki Shiratori}, {and} \bibinfo{person}{Hanbyul Joo}.} \bibinfo{year}{2021}\natexlab{}.
\newblock \showarticletitle{{FrankMocap}: A Monocular {3D} Whole-Body Pose Estimation System via Regression and Integration}. In \bibinfo{booktitle}{\emph{ICCVW}}.
\newblock


\bibitem[{Russian3DScanner}(2021)]%
        {Wrap3}
\bibfield{author}{\bibinfo{person}{{Russian3DScanner}}.} \bibinfo{year}{2021}\natexlab{}.
\newblock \bibinfo{title}{Wrap3}.
\newblock \bibinfo{howpublished}{\url{https://www.russian3dscanner.com/}}.
\newblock


\bibitem[Sanyal et~al\mbox{.}(2019)]%
        {Sanyal2019_ringnet}
\bibfield{author}{\bibinfo{person}{Soubhik Sanyal}, \bibinfo{person}{Timo Bolkart}, \bibinfo{person}{Haiwen Feng}, {and} \bibinfo{person}{Michael~J. Black}.} \bibinfo{year}{2019}\natexlab{}.
\newblock \showarticletitle{Learning to Regress {3D} Face Shape and Expression From an Image Without {3D} Supervision}. In \bibinfo{booktitle}{\emph{CVPR}}. \bibinfo{pages}{7763--7772}.
\newblock


\bibitem[Sengupta et~al\mbox{.}(2020)]%
        {sengupta2020straps}
\bibfield{author}{\bibinfo{person}{Akash Sengupta}, \bibinfo{person}{Ignas Budvytis}, {and} \bibinfo{person}{Roberto Cipolla}.} \bibinfo{year}{2020}\natexlab{}.
\newblock \showarticletitle{Synthetic Training for Accurate {3D} Human Pose and Shape Estimation in the Wild}. In \bibinfo{booktitle}{\emph{BMVC}}.
\newblock


\bibitem[Sengupta et~al\mbox{.}(2021a)]%
        {sengupta2021hierarchical}
\bibfield{author}{\bibinfo{person}{Akash Sengupta}, \bibinfo{person}{Ignas Budvytis}, {and} \bibinfo{person}{Roberto Cipolla}.} \bibinfo{year}{2021}\natexlab{a}.
\newblock \showarticletitle{Hierarchical Kinematic Probability Distributions for {3D} Human Shape and Pose Estimation from Images in the Wild}. In \bibinfo{booktitle}{\emph{ICCV}}. \bibinfo{pages}{11219--11229}.
\newblock


\bibitem[Sengupta et~al\mbox{.}(2021b)]%
        {sengupta2021probabilistic}
\bibfield{author}{\bibinfo{person}{Akash Sengupta}, \bibinfo{person}{Ignas Budvytis}, {and} \bibinfo{person}{Roberto Cipolla}.} \bibinfo{year}{2021}\natexlab{b}.
\newblock \showarticletitle{Probabilistic {3D} Human Shape and Pose Estimation from Multiple Unconstrained Images in the Wild}. In \bibinfo{booktitle}{\emph{CVPR}}. \bibinfo{pages}{16094--16104}.
\newblock


\bibitem[Shin et~al\mbox{.}(2024)]%
        {shin2024wham}
\bibfield{author}{\bibinfo{person}{Soyong Shin}, \bibinfo{person}{Juyong Kim}, \bibinfo{person}{Eni Halilaj}, {and} \bibinfo{person}{Michael~J Black}.} \bibinfo{year}{2024}\natexlab{}.
\newblock \showarticletitle{Wham: Reconstructing world-grounded humans with accurate 3d motion}. In \bibinfo{booktitle}{\emph{CVPR}}. \bibinfo{pages}{2070--2080}.
\newblock


\bibitem[{StretchSense}(2024)]%
        {stretchsense}
\bibfield{author}{\bibinfo{person}{{StretchSense}}.} \bibinfo{year}{2024}\natexlab{}.
\newblock \bibinfo{title}{StretchSense}.
\newblock \bibinfo{howpublished}{\url{https://stretchsense.com/}}.
\newblock


\bibitem[Sun et~al\mbox{.}(2019)]%
        {sun2019deep}
\bibfield{author}{\bibinfo{person}{Ke Sun}, \bibinfo{person}{Bin Xiao}, \bibinfo{person}{Dong Liu}, {and} \bibinfo{person}{Jingdong Wang}.} \bibinfo{year}{2019}\natexlab{}.
\newblock \showarticletitle{Deep High-Resolution Representation Learning for Human Pose Estimation}.
\newblock \bibinfo{journal}{\emph{CVPR}} (\bibinfo{year}{2019}).
\newblock


\bibitem[Szymanowicz et~al\mbox{.}(2022)]%
        {szymanowicz2022photo}
\bibfield{author}{\bibinfo{person}{Stanislaw Szymanowicz}, \bibinfo{person}{Virginia Estellers}, \bibinfo{person}{Tadas Baltru{\v{s}}aitis}, {and} \bibinfo{person}{Matthew Johnson}.} \bibinfo{year}{2022}\natexlab{}.
\newblock \showarticletitle{Photo-Realistic 360$^{\circ}$ Head Avatars in the Wild}. In \bibinfo{booktitle}{\emph{ECCV}}. \bibinfo{pages}{660--667}.
\newblock


\bibitem[Taheri et~al\mbox{.}(2022)]%
        {Taheri_2022_CVPR}
\bibfield{author}{\bibinfo{person}{Omid Taheri}, \bibinfo{person}{Vasileios Choutas}, \bibinfo{person}{Michael~J. Black}, {and} \bibinfo{person}{Dimitrios Tzionas}.} \bibinfo{year}{2022}\natexlab{}.
\newblock \showarticletitle{GOAL: Generating 4D Whole-Body Motion for Hand-Object Grasping}. In \bibinfo{booktitle}{\emph{CVPR}}. \bibinfo{pages}{13263--13273}.
\newblock


\bibitem[Taheri et~al\mbox{.}(2020)]%
        {taheri2020grab}
\bibfield{author}{\bibinfo{person}{Omid Taheri}, \bibinfo{person}{Nima Ghorbani}, \bibinfo{person}{Michael~J Black}, {and} \bibinfo{person}{Dimitrios Tzionas}.} \bibinfo{year}{2020}\natexlab{}.
\newblock \showarticletitle{GRAB: A dataset of whole-body human grasping of objects}. In \bibinfo{booktitle}{\emph{ECCV}}. \bibinfo{pages}{581--600}.
\newblock


\bibitem[Tan et~al\mbox{.}(2016)]%
        {tan2016fits}
\bibfield{author}{\bibinfo{person}{David~Joseph Tan}, \bibinfo{person}{Thomas Cashman}, \bibinfo{person}{Jonathan Taylor}, \bibinfo{person}{Andrew Fitzgibbon}, \bibinfo{person}{Daniel Tarlow}, \bibinfo{person}{Sameh Khamis}, \bibinfo{person}{Shahram Izadi}, {and} \bibinfo{person}{Jamie Shotton}.} \bibinfo{year}{2016}\natexlab{}.
\newblock \showarticletitle{Fits like a glove: Rapid and reliable hand shape personalization}. In \bibinfo{booktitle}{\emph{CVPR}}. \bibinfo{pages}{5610--5619}.
\newblock


\bibitem[{Ten24 Media}(2024)]%
        {3dscanstore}
\bibfield{author}{\bibinfo{person}{{Ten24 Media}}.} \bibinfo{year}{2024}\natexlab{}.
\newblock \bibinfo{title}{3D Scane Store}.
\newblock \bibinfo{howpublished}{\url{https://www.3dscanstore.com/}}.
\newblock


\bibitem[{Vicon Motion Systems}(2024)]%
        {vicon}
\bibfield{author}{\bibinfo{person}{{Vicon Motion Systems}}.} \bibinfo{year}{2024}\natexlab{}.
\newblock \bibinfo{title}{Vicon}.
\newblock \bibinfo{howpublished}{\url{https://www.vicon.com/}}.
\newblock


\bibitem[Wightman(2019)]%
        {rw2019timm}
\bibfield{author}{\bibinfo{person}{Ross Wightman}.} \bibinfo{year}{2019}\natexlab{}.
\newblock \bibinfo{title}{PyTorch Image Models}.
\newblock \bibinfo{howpublished}{\url{https://github.com/rwightman/pytorch-image-models}}.
\newblock


\bibitem[Wood et~al\mbox{.}(2021)]%
        {wood2021fake}
\bibfield{author}{\bibinfo{person}{Erroll Wood}, \bibinfo{person}{Tadas Baltru{\v{s}}aitis}, \bibinfo{person}{Charlie Hewitt}, \bibinfo{person}{Sebastian Dziadzio}, \bibinfo{person}{Thomas~J Cashman}, {and} \bibinfo{person}{Jamie Shotton}.} \bibinfo{year}{2021}\natexlab{}.
\newblock \showarticletitle{Fake it till you make it: face analysis in the wild using synthetic data alone}. In \bibinfo{booktitle}{\emph{ICCV}}. \bibinfo{pages}{3681--3691}.
\newblock


\bibitem[Wood et~al\mbox{.}(2022)]%
        {wood20223d}
\bibfield{author}{\bibinfo{person}{Erroll Wood}, \bibinfo{person}{Tadas Baltru{\v{s}}aitis}, \bibinfo{person}{Charlie Hewitt}, \bibinfo{person}{Matthew Johnson}, \bibinfo{person}{Jingjing Shen}, \bibinfo{person}{Nikola Milosavljevi{\'c}}, \bibinfo{person}{Daniel Wilde}, \bibinfo{person}{Stephan Garbin}, \bibinfo{person}{Toby Sharp}, \bibinfo{person}{Ivan Stojiljkovi{\'c}}, {et~al\mbox{.}}} \bibinfo{year}{2022}\natexlab{}.
\newblock \showarticletitle{3d face reconstruction with dense landmarks}. In \bibinfo{booktitle}{\emph{ECCV}}. \bibinfo{pages}{160--177}.
\newblock


\bibitem[Yang et~al\mbox{.}(2014)]%
        {yang2014semantic}
\bibfield{author}{\bibinfo{person}{Yipin Yang}, \bibinfo{person}{Yao Yu}, \bibinfo{person}{Yu Zhou}, \bibinfo{person}{Sidan Du}, \bibinfo{person}{James Davis}, {and} \bibinfo{person}{Ruigang Yang}.} \bibinfo{year}{2014}\natexlab{}.
\newblock \showarticletitle{Semantic parametric reshaping of human body models}. In \bibinfo{booktitle}{\emph{3DV}}, Vol.~\bibinfo{volume}{2}. \bibinfo{pages}{41--48}.
\newblock


\bibitem[Yang et~al\mbox{.}(2023)]%
        {yang2023synbody}
\bibfield{author}{\bibinfo{person}{Zhitao Yang}, \bibinfo{person}{Zhongang Cai}, \bibinfo{person}{Haiyi Mei}, \bibinfo{person}{Shuai Liu}, \bibinfo{person}{Zhaoxi Chen}, \bibinfo{person}{Weiye Xiao}, \bibinfo{person}{Yukun Wei}, \bibinfo{person}{Zhongfei Qing}, \bibinfo{person}{Chen Wei}, \bibinfo{person}{Bo Dai}, {et~al\mbox{.}}} \bibinfo{year}{2023}\natexlab{}.
\newblock \showarticletitle{Synbody: Synthetic dataset with layered human models for 3d human perception and modeling}. In \bibinfo{booktitle}{\emph{ICCV}}.
\newblock


\bibitem[Zhang et~al\mbox{.}(2023b)]%
        {zhang2023pymaf}
\bibfield{author}{\bibinfo{person}{Hongwen Zhang}, \bibinfo{person}{Yating Tian}, \bibinfo{person}{Yuxiang Zhang}, \bibinfo{person}{Mengcheng Li}, \bibinfo{person}{Liang An}, \bibinfo{person}{Zhenan Sun}, {and} \bibinfo{person}{Yebin Liu}.} \bibinfo{year}{2023}\natexlab{b}.
\newblock \showarticletitle{Pymaf-x: Towards well-aligned full-body model regression from monocular images}.
\newblock \bibinfo{journal}{\emph{PAMI}} (\bibinfo{year}{2023}).
\newblock


\bibitem[Zhang et~al\mbox{.}(2023a)]%
        {zhang2023accurate}
\bibfield{author}{\bibinfo{person}{Tianke Zhang}, \bibinfo{person}{Xuangeng Chu}, \bibinfo{person}{Yunfei Liu}, \bibinfo{person}{Lijian Lin}, \bibinfo{person}{Zhendong Yang}, \bibinfo{person}{Zhengzhuo Xu}, \bibinfo{person}{Chengkun Cao}, \bibinfo{person}{Fei Yu}, \bibinfo{person}{Changyin Zhou}, \bibinfo{person}{Chun Yuan}, {et~al\mbox{.}}} \bibinfo{year}{2023}\natexlab{a}.
\newblock \showarticletitle{Accurate 3D Face Reconstruction with Facial Component Tokens}. In \bibinfo{booktitle}{\emph{ICCV}}. \bibinfo{pages}{9033--9042}.
\newblock


\bibitem[Zhang et~al\mbox{.}(2021)]%
        {zhang2021we}
\bibfield{author}{\bibinfo{person}{Yan Zhang}, \bibinfo{person}{Michael~J Black}, {and} \bibinfo{person}{Siyu Tang}.} \bibinfo{year}{2021}\natexlab{}.
\newblock \showarticletitle{We are more than our joints: Predicting how 3d bodies move}. In \bibinfo{booktitle}{\emph{CVPR}}. \bibinfo{pages}{3372--3382}.
\newblock


\bibitem[Zhou et~al\mbox{.}(2019)]%
        {zhou2019continuity}
\bibfield{author}{\bibinfo{person}{Yi Zhou}, \bibinfo{person}{Connelly Barnes}, \bibinfo{person}{Jingwan Lu}, \bibinfo{person}{Jimei Yang}, {and} \bibinfo{person}{Hao Li}.} \bibinfo{year}{2019}\natexlab{}.
\newblock \showarticletitle{On the Continuity of Rotation Representations in Neural Networks}. In \bibinfo{booktitle}{\emph{CVPR}}. \bibinfo{pages}{5745--5753}.
\newblock


\bibitem[Zielonka et~al\mbox{.}(2022)]%
        {zielonka2022towards}
\bibfield{author}{\bibinfo{person}{Wojciech Zielonka}, \bibinfo{person}{Timo Bolkart}, {and} \bibinfo{person}{Justus Thies}.} \bibinfo{year}{2022}\natexlab{}.
\newblock \showarticletitle{Towards metrical reconstruction of human faces}. In \bibinfo{booktitle}{\emph{ECCV}}. \bibinfo{pages}{250--269}.
\newblock


\bibitem[Zimmermann et~al\mbox{.}(2019)]%
        {zimmermann2019freihand}
\bibfield{author}{\bibinfo{person}{Christian Zimmermann}, \bibinfo{person}{Duygu Ceylan}, \bibinfo{person}{Jimei Yang}, \bibinfo{person}{Bryan Russell}, \bibinfo{person}{Max Argus}, {and} \bibinfo{person}{Thomas Brox}.} \bibinfo{year}{2019}\natexlab{}.
\newblock \showarticletitle{{FreiHAND}: A Dataset for Markerless Capture of Hand Pose and Shape from Single {RGB} Images}. In \bibinfo{booktitle}{\emph{ICCV}}. \bibinfo{pages}{813--822}.
\newblock


\end{thebibliography}

\end{document}